\documentclass[final]{cvpr}

\usepackage{times}
\usepackage{epsfig}
\usepackage{graphicx}
\usepackage{amsmath}
\usepackage{amssymb}

\usepackage{dsfont}
\usepackage{bm}  
\usepackage{booktabs}
\usepackage{xspace}
\usepackage{multirow}
\usepackage{subfigure}
\usepackage{xcolor}
\usepackage{colortbl}
\usepackage{enumitem}
\usepackage{placeins}
\usepackage{footnote}
\usepackage[title]{appendix}

\usepackage[pagebackref=true,breaklinks=true,colorlinks,bookmarks=false]{hyperref}

\newcommand{\hatm}{\hat{m}}
\newcommand{\hatc}{\hat{c}}

\newcommand{\pqobjpos}{{\cal O}_{\rm PQ}^{\rm pos}}
\newcommand{\pqlosspos}{{\cal L}_{\rm PQ}^{\rm pos}}
\newcommand{\pqlossneg}{{\cal L}_{\rm PQ}^{\rm neg}}
\newcommand{\pqloss}{{\cal L}_{\rm PQ}}

\newcommand{\insloss}{{\cal L}^{\rm InstDis}}

\newcommand{\hz}{\hat{z}}

\newcommand{\noobject}{\varnothing}

\newcommand{\hy}{\hat{y}}

\newcommand{\hp}{\hat{p}}

\renewcommand{\Sigma}{\mathfrak{S}}

\def\1{\bm{1}}

\DeclareMathAlphabet{\mathsfit}{\encodingdefault}{\sfdefault}{m}{sl}
\SetMathAlphabet{\mathsfit}{bold}{\encodingdefault}{\sfdefault}{bx}{n}

\def\sC{{\mathbb{C}}}

\DeclareMathOperator*{\argmax}{arg\,max}

\definecolor{Gray}{gray}{0.9}

\newcommand{\eqnspace}{
\setlength{\abovedisplayskip}{5pt plus 1.0pt minus 1.0pt}
\setlength{\belowdisplayskip}{5pt plus 1.0pt minus 1.0pt}
\setlength{\abovedisplayshortskip}{0.0pt plus 3.0pt}
\setlength{\belowdisplayshortskip}{6.0pt plus 3.0pt minus 3.0pt}}

\newcommand{\eqnsm}[2]{\begin{equation}\label{eq:#1}#2\end{equation}}
\newcommand{\eqnalism}[2]{\begin{equation}\begin{aligned}\label{eq:#1}#2\end{aligned}\end{equation}}
\newcommand{\eqngatsm}[2]{\begin{equation}\begin{gathered}\label{eq:#1}#2\end{gathered}\end{equation}}

\newlength\secmargin
\newlength\paramargin
\newlength\abovetabcapmargin
\newlength\belowtabcapmargin
\newlength\abovefigcapmargin
\newlength\belowfigcapmargin

\makeatletter\renewcommand\paragraph{\@startsection{paragraph}{4}{\z@}
  {.5em \@plus1ex \@minus.2ex}{-.5em}{\normalfont\normalsize\bfseries}}\makeatother

\usepackage{pifont}
\newcommand{\cmark}{\ding{51}}
\newcommand{\xmark}{\ding{55}}

\def\cvprPaperID{458} 
\def\confYear{CVPR 2021}

\definecolor{vis_red}{HTML}{EA4335}
\definecolor{vis_green}{HTML}{34A853}
\definecolor{vis_blue}{HTML}{4285F4}
\definecolor{vis_yellow}{HTML}{FBBC04}

\newcommand{\red}[1]{\textcolor{vis_red}{\textbf{#1}}}
\newcommand{\green}[1]{\textcolor{vis_green}{\textbf{#1}}}
\newcommand{\blue}[1]{\textcolor{vis_blue}{\textbf{#1}}}
\newcommand{\yellow}[1]{\textcolor{vis_yellow}{\textbf{#1}}}

\begin{document}

\title{MaX-DeepLab: End-to-End Panoptic Segmentation with Mask Transformers}

\author{
Huiyu Wang\textsuperscript{1}\thanks{Work done while an intern at Google.}~~~~~~Yukun Zhu\textsuperscript{2}~~~~~~Hartwig Adam\textsuperscript{2}~~~~~~Alan Yuille\textsuperscript{1}~~~~~~Liang-Chieh Chen\textsuperscript{2}\\\textsuperscript{1}Johns Hopkins University~~~~~~\textsuperscript{2}Google Research\\
\normalsize Code: \url{https://github.com/google-research/deeplab2}}

\maketitle

\begin{abstract}
We present MaX-DeepLab, the first end-to-end model for panoptic segmentation. Our approach simplifies the current pipeline that depends heavily on surrogate sub-tasks and hand-designed components, such as box detection, non-maximum suppression, thing-stuff merging, \etc. Although these sub-tasks are tackled by area experts, they fail to comprehensively solve the target task. By contrast, our MaX-DeepLab directly predicts class-labeled masks with a mask transformer, and is trained with a panoptic quality inspired loss via bipartite matching. Our mask transformer employs a dual-path architecture that introduces a global memory path in addition to a CNN path, allowing direct communication with any CNN layers. As a result, MaX-DeepLab shows a significant 7.1\% PQ gain in the box-free regime on the challenging COCO dataset, closing the gap between box-based and box-free methods for the first time. A small variant of MaX-DeepLab improves 3.0\% PQ over DETR with similar parameters and M-Adds. Furthermore, MaX-DeepLab, without test time augmentation, achieves new state-of-the-art 51.3\% PQ on COCO test-dev set.
\end{abstract}
\eqnspace
\FloatBarrier
\section{Introduction}

\begin{figure}[t]
    \centering
    \includegraphics[width=\linewidth]{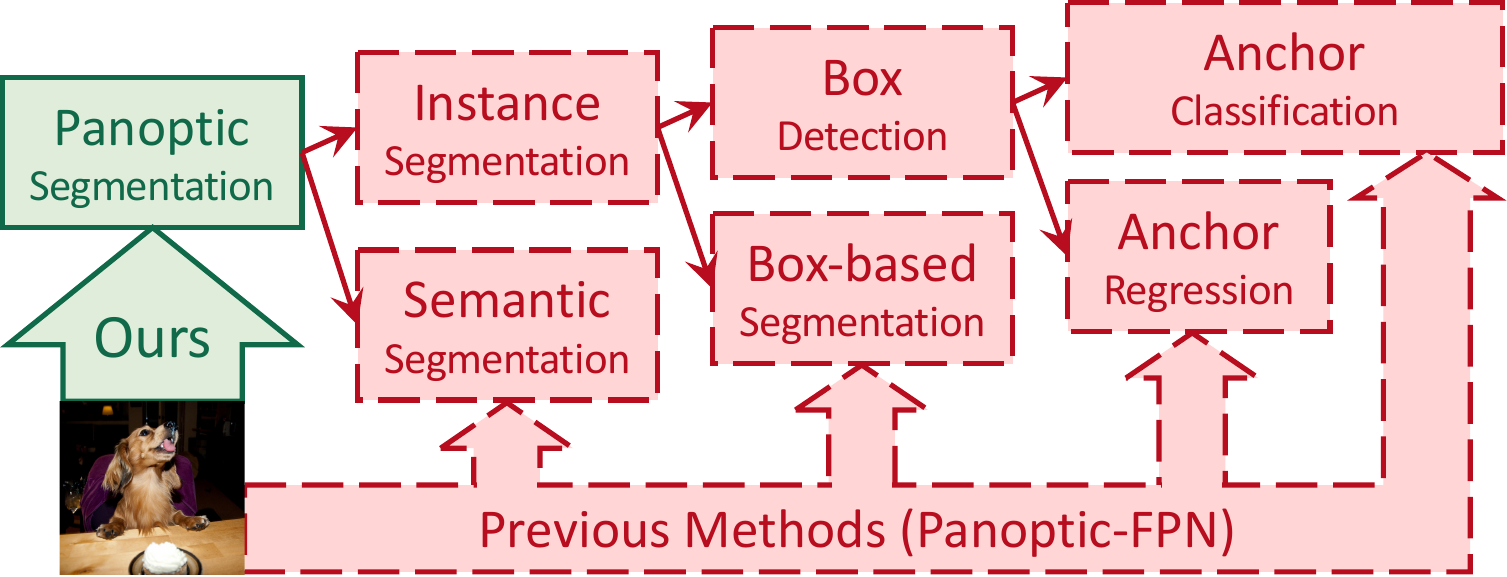}
    \vskip \abovefigcapmargin
    \caption{Our method predicts panoptic segmentation masks {\bf directly from images}, while previous methods (Panoptic-FPN as an example) rely on {\bf a tree of surrogate sub-tasks}.
    Panoptic segmentation masks are obtained by merging semantic and instance segmentation results. Instance segmentation is further decomposed into box detection and box-based segmentation, while box detection is achieved by anchor regression and anchor classification.
    }
    \label{fig:surrogate}
    \vspace{\belowfigcapmargin}
\end{figure}

\begin{figure}[t]
    \centering
    \setlength{\tabcolsep}{0.0em}
    \subfigure[Our MaX-DeepLab \newline \phantom{.....} 51.1 PQ (box-free)]{
        \begin{tabular}{c}
        \includegraphics[width=0.32\linewidth]{figures/original_image.jpg} \\
        \includegraphics[width=0.32\linewidth]{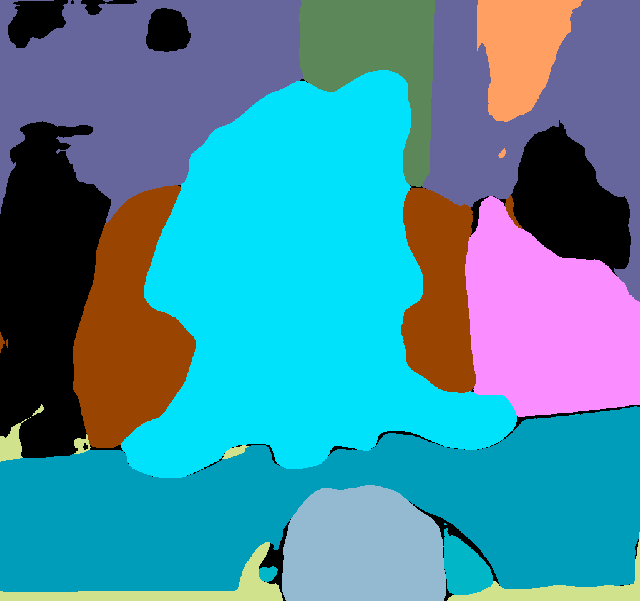}
        \end{tabular}
        \label{fig:max}}
        \hspace{-1.7ex}
    \subfigure[Axial-DeepLab~\cite{wang2020axial} \newline \phantom{.....} 43.4 PQ (box-free)]{
        \begin{tabular}{c}
        \includegraphics[width=0.32\linewidth]{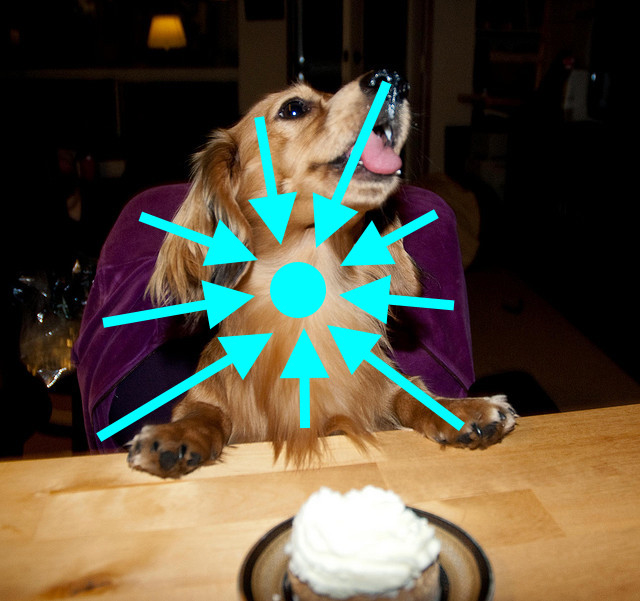} \\
        \includegraphics[width=0.32\linewidth]{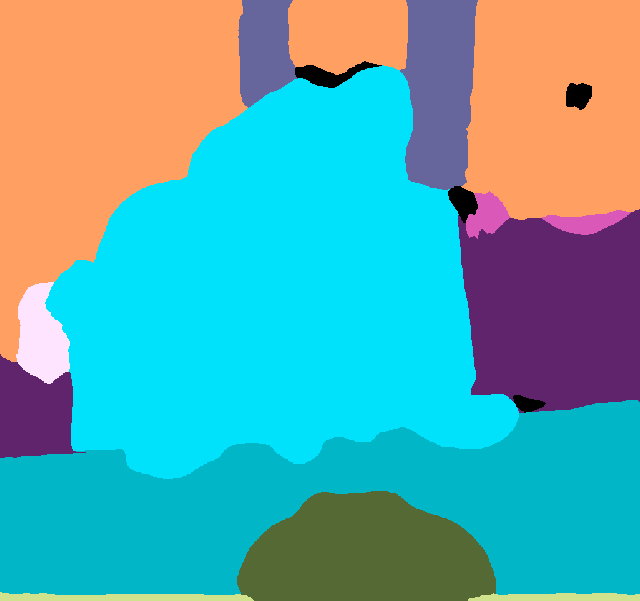}
        \end{tabular}
        \label{fig:axial}}
        \hspace{-1.7ex}
    \subfigure[\phantom{..} DetectoRS~\cite{qiao2020detectors} \newline \phantom{...} 48.6 PQ (box-based)]{
        \begin{tabular}{c}
        \includegraphics[width=0.32\linewidth]{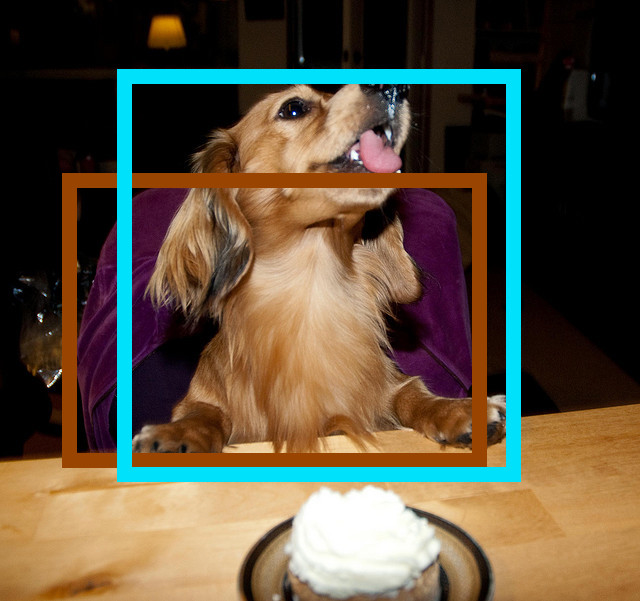} \\
        \includegraphics[width=0.32\linewidth]{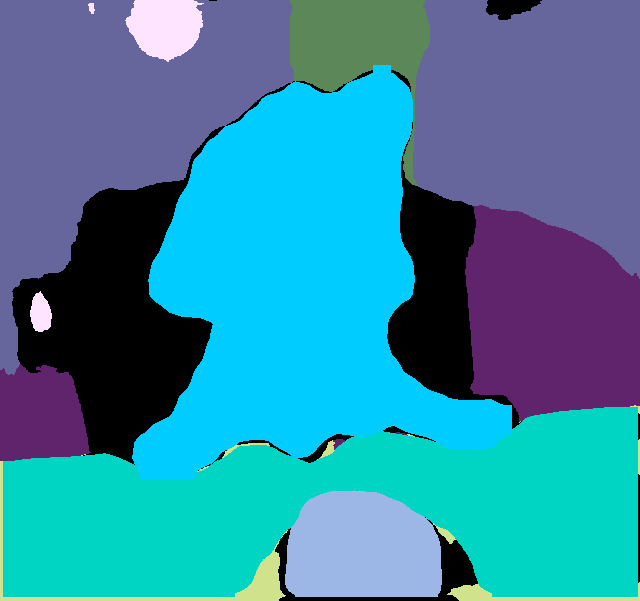}
        \end{tabular}
        \label{fig:detectors}}
    \vskip \abovefigcapmargin
    \caption{A case study for our method and state-of-the-art {\it box-free} and {\it box-based} methods. \subref{fig:max}~Our end-to-end MaX-DeepLab correctly segments a dog sitting on a chair. \subref{fig:axial}~Axial-DeepLab~\cite{wang2020axial} relies on a surrogate sub-task of regressing object center offsets~\cite{cheng2019panoptic}. It fails because the centers of the dog and the chair are close to each other. \subref{fig:detectors}~DetectoRS~\cite{qiao2020detectors} classifies object bounding boxes, instead of masks, as a surrogate sub-task. It filters out the chair mask because the chair bounding box has a low confidence. } 
    \label{fig:teaser}
    \vspace{\belowfigcapmargin}
\end{figure}

The goal of panoptic segmentation~\cite{kirillov2018panoptic} is to predict a set of non-overlapping masks along with their corresponding class labels.
Modern panoptic segmentation methods address this mask prediction problem by approximating the target task with multiple surrogate sub-tasks.
For example, Panoptic-FPN~\cite{kirillov2019panoptic} adopts a `box-based pipeline' with three levels of surrogate sub-tasks, as demonstrated in a tree structure in \figref{fig:surrogate}.
Each level of this proxy tree involves manually-designed modules, such as anchors~\cite{ren2015faster}, box assignment rules~\cite{zhang2020bridging}, non-maximum suppression (NMS)~\cite{bodla2017soft}, thing-stuff merging~\cite{xiong2019upsnet}, \etc.
Although there are good solutions~\cite{ren2015faster,deeplabv12015,he2017mask} to individual surrogate sub-tasks and modules, undesired artifacts are introduced when these sub-tasks fit into a pipeline for panoptic segmentation, especially in the challenging conditions (\figref{fig:teaser}).

\begin{table}[t]
    \label{tab:intro}
    \vspace{-0.0em}
    \setlength{\tabcolsep}{0.4em}
        \begin{center}
        \resizebox{\linewidth}{!}{
        \begin{tabular}{l|ccccc}
        \toprule[0.2em]
            \multirow{2}{*}{Method} & Anchor & Center & NMS & Merge & Box \\
            & -Free & -Free & -Free & -Free & -Free \\
            \toprule[0.2em]
            Panoptic-FPN~\cite{kirillov2019panoptic} & \xmark & \cmark & \xmark & \xmark & \xmark \\
            UPSNet~\cite{xiong2019upsnet} & \xmark & \cmark & \xmark & \cmark & \xmark \\
            DETR~\cite{carion2020end} & \cmark & \cmark & \cmark & \cmark & \xmark \\
            \midrule \midrule
            Axial-DeepLab~\cite{wang2020axial} & \cmark & \xmark & \xmark & \xmark & \cmark \\
            MaX-DeepLab & \cmark & \cmark & \cmark & \cmark & \cmark \\
            \bottomrule[0.1em]
        \end{tabular}
        }
        \end{center}
        \vspace{\abovetabcapmargin}
    \caption{Our end-to-end MaX-DeepLab dispenses with these common hand-designed components necessary for existing methods.}
    \vspace{\belowtabcapmargin}
    \end{table}
    
Recent work on panoptic segmentation attempted to simplify this box-based pipeline. For example, UPSNet~\cite{xiong2019upsnet} proproses a parameter-free panoptic head, permitting back-propagation to both semantic and instance segmentation modules. Recently, DETR~\cite{carion2020end} presents an end-to-end approach for box detection, which is used to replace detectors in panoptic segmentation, but the whole training process of DETR still relies heavily on the box detection task.

Another line of work made efforts to completely remove boxes from the pipeline, which aligns better with the mask-based definition of panoptic segmentation. 
The state-of-the-art method in this regime, Axial-DeepLab~\cite{wang2020axial}, along with other box-free methods~\cite{yang2019deeperlab,cheng2019panoptic,chen2020naive}, predicts pixel-wise offsets to pre-defined instance centers.
But this center-based surrogate sub-task makes it challenging to deal with highly deformable objects, or near-by objects with close centers.
As a result, box-free methods do not perform as well as box-based methods on the challenging COCO dataset~\cite{lin2014microsoft}.

In this paper, we streamline the panoptic segmentation pipeline with an end-to-end approach.
Inspired by DETR~\cite{carion2020end}, our model {\it directly} predicts a set of non-overlapping masks and their corresponding semantic labels with a mask transformer.
The output masks and classes are optimized with a panoptic quality (PQ) style objective.
Specifically, inspired by the definition of PQ~\cite{kirillov2018panoptic}, we define a similarity metric between two class-labeled masks as the multiplication of their mask similarity and their class similarity.
Our model is trained by maximizing this similarity between ground truth masks and predicted masks via one-to-one bipartite matching~\cite{Kuhn1955NAVAL,stewart2016end,carion2020end}.
This direct modeling of panoptic segmentation enables end-to-end training and inference, 
removing those hand-coded priors that are necessary in existing box-based and box-free methods~(\tabref{tab:intro}).
Our method is dubbed MaX-DeepLab for extending Axial-DeepLab with a {\bf Ma}sk {\bf X}former.

In companion with direct training and inference, we equip our mask transformer with a novel architecture.
Instead of stacking a traditional transformer~\cite{vaswani2017attention,carion2020end} on top of a Convolutional Neural Network (CNN)~\cite{lecun1998gradient}, we propose a dual-path framework for combining CNNs with transformers.
Specifically, we enable any CNN layer to read and write a global memory, using our dual-path transformer block. This block supports all types of attention between the CNN-path and the memory-path, including memory-path self-attention ({\it M2M}), pixel-path axial self-attention ({\it P2P}), memory-to-pixel attention ({\it M2P}), and finally pixel-to-memory attention ({\it P2M}). The transformer block can be inserted anywhere in a CNN, enabling communication with the global memory at any layer. Besides this communication module, our MaX-DeepLab employs a stacked-hourglass-style decoder~\cite{ronneberger2015u,newell2016stacked,cheng2019spgnet}. 
The decoder aggregates multi-scale features into a high resolution output, which 
is then multiplied with the global memory feature, to form our mask set prediction. The classes for the masks are predicted with another branch of the mask transformer.

We evaluate MaX-DeepLab on one of the most challenging panoptic segmentation datasets, COCO~\cite{lin2014microsoft}, against the state-of-the-art box-free method, Axial-DeepLab~\cite{wang2020axial}, and state-of-the-art box-based method, DetectoRS~\cite{wu2019detectron2} (\figref{fig:teaser}). Our MaX-DeepLab, {\it without} test time augmentation (TTA), achieves the state-of-the-art result of 51.3\% PQ on the test-dev set. This result surpasses Axial-DeepLab (with TTA) by 7.1\% PQ in the box-free regime, and outperforms DetectoRS (with TTA) by 1.7\% PQ, bridging the gap between box-based and box-free methods for the first time. For a fair comparison with DETR~\cite{carion2020end}, we also evaluate a lightweight model, MaX-DeepLab-S, that matches the number of parameters and M-Adds of DETR. We observe that MaX-DeepLab-S outperforms DETR by 3.3\% PQ on the val set and 3.0\% PQ on the test-dev set. In addition, we perform extensive ablation studies and analyses on our end-to-end formulation, model scaling, dual-path architectures, and our loss functions. We also notice that the extra-long training schedule of DETR~\cite{carion2020end} is not necessary for MaX-DeepLab.

To summarize, our contributions are four-fold:
\begin{itemize}[nosep]
  \item MaX-DeepLab is the first end-to-end model for panoptic segmentation, inferring masks and classes directly without hand-coded priors like object centers or boxes.
  \item We propose a training objective that optimizes a PQ-style loss function via a PQ-style bipartite matching between predicted masks and ground truth masks.
  \item Our dual-path transformer enables CNNs to read and write a global memory at any layer, providing a new way of combining transformers with CNNs.
  \item MaX-DeepLab closes the gap between box-based and box-free methods and sets a new state-of-the-art on COCO, even without using test time augmentation.
\end{itemize}

\section{Related Work}
\label{sec:relatedwork}

\vspace{0.3\paramargin}\paragraph{Transformers.}
Transformers~\cite{vaswani2017attention}, first introduced for neural machine translation, have advanced the state-of-the-art in many natural language processing tasks~\cite{devlin2019bert,shaw2018self,dai2019transformer}.
Attention~\cite{bahdanau2014neural}, as the core component of Transformers, was developed to capture both correspondence of tokens across modalities~\cite{bahdanau2014neural} and long-range interactions in a single context (self-attention)~\cite{cheng2016long,vaswani2017attention}.
Later, the complexity of transformer attention has been reduced~\cite{kitaev2020reformer,wang2020linformer}, by introducing local~\cite{luong2015effective} or sparse attention~\cite{child2019generating}, together with a global memory~\cite{beltagy2020longformer,zaheer2020big,gupta2020gmat,ainslie2020etc}. The global memory, which inspires our dual-path transformer, recovers long-range context by propagating information globally.

Transformer and attention have been applied to computer vision as well, by combining non-local modules~\cite{wang2018non,buades2005non} with CNNs or by applying self-attention only~\cite{parmar2019stand,hu2019local,wang2020axial}. Both classes of methods have boosted various vision tasks such as image classification~\cite{chen20182,bello2019attention, parmar2019stand,hu2019local,li2020neural,wang2020axial,dosovitskiy2020image}, object detection~\cite{wang2018non,shen2021efficient,parmar2019stand,hu2018relation,carion2020end,zhu2020deformable}, semantic segmentation~\cite{chen2016attention,zhao2018psanet,huang2019ccnet,fu2019dual,zhu2019asymmetric,zhu2019empirical}, video recognition~\cite{wang2018non,chen20182}, image generation~\cite{parmar2018image,ho2019axial}, and panoptic segmentation~\cite{wang2020axial}. It is worth mentioning that DETR~\cite{carion2020end} stacked a transformer on top of a CNN for end-to-end object detection.

\vspace{\paramargin}\paragraph{Box-based panoptic segmentation.} 
Most panoptic segmentation models, such as Panoptic FPN~\cite{kirillov2019panoptic}, follow a box-based approach that detects object bounding boxes and predicts a mask for each box, usually with a Mask R-CNN~\cite{he2017mask} and FPN~\cite{lin2017feature}. Then, the instance segments (`thing') and semantic segments (`stuff')~\cite{chen2018deeplabv2} are fused by merging modules~\cite{li2018learning,li2018attention,porzi2019seamless,liu2019e2e,yang2020sognet} to generate panoptic segmentation. For example, UPSNet \cite{xiong2019upsnet} developed a parameter-free panoptic head, which facilitates unified training and inference~\cite{li2020unifying}. Recently, DETR~\cite{carion2020end} extended box-based methods with its transformer-based end-to-end detector. And DetectoRS~\cite{qiao2020detectors} advanced the state-of-the-art
with recursive feature pyramid and switchable atrous convolution.

\vspace{\paramargin}\paragraph{Box-free panoptic segmentation.}
Contrary to box-based approaches, box-free methods typically start with semantic segments~\cite{deeplabv12015,chen2017deeplabv3,deeplabv3plus2018}. Then, instance segments are obtained by grouping `thing' pixels with various methods, such as instance center regression~\cite{kendall2018multi,uhrig2018box2pix,neven2019instance,yang2019deeperlab,cheng2019panopticworkshop}, Watershed transform \cite{vincent1991watersheds,bai2017deep,bonde2020towards}, Hough-voting \cite{ballard1981generalizing,leibe2004combined,bonde2020towards}, or pixel affinity \cite{keuper2015efficient,liu2018affinity,sofiiuk2019adaptis,gao2019ssap,bonde2020towards}. Recently, Axial-DeepLab~\cite{wang2020axial} advanced the state-of-the-art by equipping Panoptic-DeepLab~\cite{cheng2019panoptic} with a fully axial-attention~\cite{ho2019axial} backbone. In this work, we extend Axial-DeepLab with a mask transformer for end-to-end panoptic segmentation.

\section{Method}
In this section, we describe how MaX-DeepLab directly predicts class-labeled masks for panoptic segmentation, followed by the PQ-style loss used to train the model. Then, we introduce our dual-path transformer architecture as well as the auxiliary losses that are helpful in training.

\subsection{MaX-DeepLab formulation}
\label{sec:model}
The goal of panoptic segmentation is to segment the image $I \in \mathbb{R}^{H \times W \times 3}$ into a set of class-labeled masks:
\eqnsm{yi}{\{y_i\}_{i=1}^K = \{(m_i, c_i)\}_{i=1}^K \,.}
The $K$ ground truth masks $m_i \in {\{0,1\}}^{H \times W}$ do not overlap with each other, \ie, $\sum_{i=1}^{K} m_i \leq 1^{H \times W}$, and $c_i$ denotes the ground truth class label of mask $m_i$.

Our MaX-DeepLab directly predicts outputs in the exact same form as the ground truth. MaX-DeepLab segments the image $I$ into a fixed-size set of class-labeled masks:
\eqnsm{hyi}{\{\hy_i\}_{i=1}^N = \{(\hatm_i, \hp_i(c))\}_{i=1}^N \,.}
The constant size $N$ of the set is much larger than the typical number of masks in an image~\cite{carion2020end}. The predicted masks $ \hatm_i \in {[0,1]}^{H \times W}$ are softly exclusive to each other, \ie, $\sum_{i=1}^{N} \hatm_i = 1^{H \times W}$, and $\hp_i(c)$ denotes the probability of assigning class $c$ to mask $\hatm_i$. Possible classes $\sC \ni c$ include thing classes, stuff classes, and a $\noobject$ class (no object). In this way, MaX-DeepLab deals with thing and stuff classes in a unified manner, removing the need for merging operators.

\vspace{\paramargin}\paragraph{Simple inference.}
End-to-end inference of MaX-DeepLab is enabled by adopting the same formulation for both ground truth definition and model prediction.
As a result, the final panoptic segmentation prediction is obtained by simply performing argmax twice.
Specifically, the first argmax predicts a class label for each mask:
\setlength{\abovedisplayskip}{3.2pt plus 1.0pt minus 1.0pt}
\setlength{\belowdisplayskip}{3.2pt plus 1.0pt minus 1.0pt}
\eqnsm{classargmax}{\hatc_i = \argmax_c \hp_i(c) \,.}
And the other argmax assigns a mask-ID $\hz_{h,w}$ to each pixel:
\eqngatsm{pixelargmax}{\hz_{h,w} = \argmax_i \hatm_{i,h,w} \,, \\
\forall h \in \{1, 2, \dots, H\},\quad\forall w \in \{1, 2, \dots, W\} \,.}
In practice, we filter each argmax with a confidence threshold -- Masks or pixels with a low confidence are removed as described in \secref{sec:exp}. In this way, MaX-DeepLab infers panoptic segmentation directly, dispensing with common manually-designed post-processing, \eg, NMS and thing-stuff merging in almost all previous methods~\cite{kirillov2019panoptic,xiong2019upsnet}. Besides, MaX-DeepLab does not rely on hand-crafted priors such as anchors, object boxes, or instance mass centers, \etc.

\subsection{PQ-style loss}
\label{sec:pqloss}
In addition to simple inference, MaX-DeepLab enables end-to-end training as well. In this section, we introduce how we train MaX-DeepLab with our PQ-style loss, which draws inspiration from the definition of \textit{panoptic quality} (PQ)~\cite{kirillov2018panoptic}. This evaluation metric of panoptic segmentation, PQ, is defined as the multiplication of a \textit{recognition quality} (RQ) term and a \textit{segmentation quality} (SQ) term:
\setlength{\abovedisplayskip}{2pt plus 1.0pt minus 1.0pt}
\setlength{\belowdisplayskip}{2pt plus 1.0pt minus 1.0pt}
\eqnsm{PQ}{
\small{\text{PQ}} = \text{RQ} \times \text{SQ} \,.}
\setlength{\abovedisplayskip}{5pt plus 1.0pt minus 1.0pt}
\setlength{\belowdisplayskip}{5pt plus 1.0pt minus 1.0pt}
Based on this decomposition of PQ, we design our objective in the same manner: First, we define a PQ-style similarity metric between a class-labeled ground truth mask and a predicted mask. Next, we show how we match a predicted mask to each ground truth mask with this metric, and finally how to optimize our model with the same metric.

\vspace{\paramargin}\paragraph{Mask similarity metric.}
Our mask similarity metric ${\rm sim} (\cdot, \cdot)$ between a class-labeled ground truth mask $y_i = (m_i, c_i)$ and a prediction $\hy_j = (\hatm_j, \hp_j(c))$ is defined as
\eqnsm{similarity} {
{\rm sim} (y_i, \hy_j) = \underbrace{\hp_j(c_i)}_{\approx~\text{RQ}} \times \underbrace{{\rm Dice}(m_i, \hatm_j)}_{\approx~\text{SQ}} \,,
}
where $\hp_j(c_i) \in [0, 1]$ is the probability of predicting the correct class (recognition quality) and ${\rm Dice}(m_i, \hatm_j) \in [0, 1]$ is the Dice coefficient between a predicted mask $\hatm_j$ and a ground truth $m_i$ (segmentation quality). The two terms are multiplied together, analogous to the decomposition of PQ.

This mask similarity metric has a lower bound of 0, which means either the class prediction is incorrect, OR the two masks do not overlap with each other. The upper bound, 1, however, is only achieved when the class prediction is correct AND the mask is perfect. The AND gating enables this metric to serve as a good optimization objective for both model training and mask matching.

\vspace{\paramargin}\paragraph{Mask matching.}
In order to assign a predicted mask to each ground truth, we solve a one-to-one bipartite matching problem between the prediction set $\{\hy_i\}_{i=1}^N$ and the ground truth set $\{y_i\}_{i=1}^K$. Formally, we search for a permutation of $N$ elements $\sigma \in \Sigma_N$ that best assigns the predictions to achieve the maximum total similarity to the ground truth:
\eqnsm{matching}{\hat{\sigma} = \argmax_{\sigma\in\Sigma_N} \sum_{i=1}^{K} {\rm sim}(y_i, \hy_{\sigma(i)}) \,.}
The optimal assignment is computed efficiently with the Hungarian algorithm~\cite{Kuhn1955NAVAL}, following prior work~\cite{carion2020end,stewart2016end}. We refer to the $K$ matched predictions as positive masks which will be optimized to predict the corresponding ground truth masks and classes. The $(N-K)$ masks left are negatives, which should predict the $\noobject$ class (no object).

Our one-to-one matching is similar to DETR~\cite{carion2020end}, but with a different purpose: DETR allows only one positive match in order to remove duplicated boxes in the absence of NMS, while in our case, duplicated or overlapping masks are precluded by design. But in our case, assigning multiple predicted masks to one ground truth mask is problematic too, because multiple masks cannot possibly be optimized to fit a single ground truth mask at the same time. In addition, our one-to-one matching is consistent with the PQ metric, where only one predicted mask can theoretically match (\ie, have an IoU over 0.5) with each ground truth mask.

\vspace{\paramargin}\paragraph{PQ-style loss.} Given our mask similarity metric and the mask matching process based on this metric, it is straight forward to optimize model parameters $\theta$ by maximizing this same similarity metric over matched (\ie, positive) masks:
\eqnsm{shortobj}{\max_{\theta} \sum_{i=1}^K {\rm sim}(y_i, \hy_{\hat{\sigma}(i)}) 	\Leftrightarrow \max_{\theta, \sigma\in\Sigma_N} \sum_{i=1}^K {\rm sim}(y_i, \hy_{\sigma(i)}) \,.}
Substituting the similarity metric (\equref{eq:similarity}) gives our PQ-style objective $\pqobjpos$ to be maximized for positive masks:
\eqnsm{obj} {
\max_{\theta} \pqobjpos = \sum_{i=1}^K \underbrace{\hp_{\hat{\sigma}(i)}(c_i)}_{\approx~\text{RQ}} \times \underbrace{{\rm Dice}(m_i, \hatm_{\hat{\sigma}(i)})}_{\approx~\text{SQ}}\,.}
In practice, we rewrite $\pqobjpos$ into two common loss terms by applying the product rule of gradient and then changing a probability $\hp$ to a log probability $\bm{\log{}}\hp$. The change from $\hp$ to $\bm{\log{}}\hp$ aligns with the common cross-entropy loss and scales gradients better in practice for optimization:
\eqnalism{pqlosspos} {
\pqlosspos =&\sum_{i=1}^K \underbrace{\hp_{\hat{\sigma}(i)}(c_i)}_{\text{weight}} \cdot \underbrace{\big[-{\rm Dice}(m_i, \hatm_{\hat{\sigma}(i)})\big]}_{\text{Dice loss}} \\
+&\sum_{i=1}^K \underbrace{{\rm Dice}(m_i, \hatm_{\hat{\sigma}(i)})}_{\text{weight}} \cdot \underbrace{\big[- \bm{\log{}} \hp_{\hat{\sigma}(i)}(c_i)\big]}_{\text{Cross-entropy loss}} \,,}
where the loss weights are constants (\ie, no gradient is passed to them). This reformulation provides insights by bridging our objective with common loss functions: Our PQ-style loss is equivalent to optimizing a dice loss weighted by the class correctness and optimizing a cross-entropy loss weighted by the mask correctness. The logic behind this loss is intuitive: we want {\it both} of the mask and class to be correct at the same time. For example, if a mask is far off the target, it is a false negative anyway, so we disregard its class. This intuition aligns with the down-weighting of class losses for wrong masks, and vice versa. 

Apart from the $\pqlosspos$ for positive masks, we define a cross-entropy term $\pqlossneg$ for negative (unmatched) masks:
\eqnsm{pqlossneg} {
\pqlossneg = \sum_{i=K+1}^N \big[-\log\hp_{\hat{\sigma}(i)}(\noobject)\big] \,.}
This term trains the model to predict $\noobject$ for negative masks. We balance the two terms by $\alpha$, as a common practice to weight positive and negative samples~\cite{lin2017focal}: 
\eqnsm{pqloss} {
\pqloss = \alpha \pqlosspos + (1 - \alpha) \pqlossneg
\,,}
where $\pqloss$ denotes our final PQ-style loss.
\subsection{MaX-DeepLab Architecture}
\begin{figure}[!t]
    \centering
    \includegraphics[width=\linewidth]{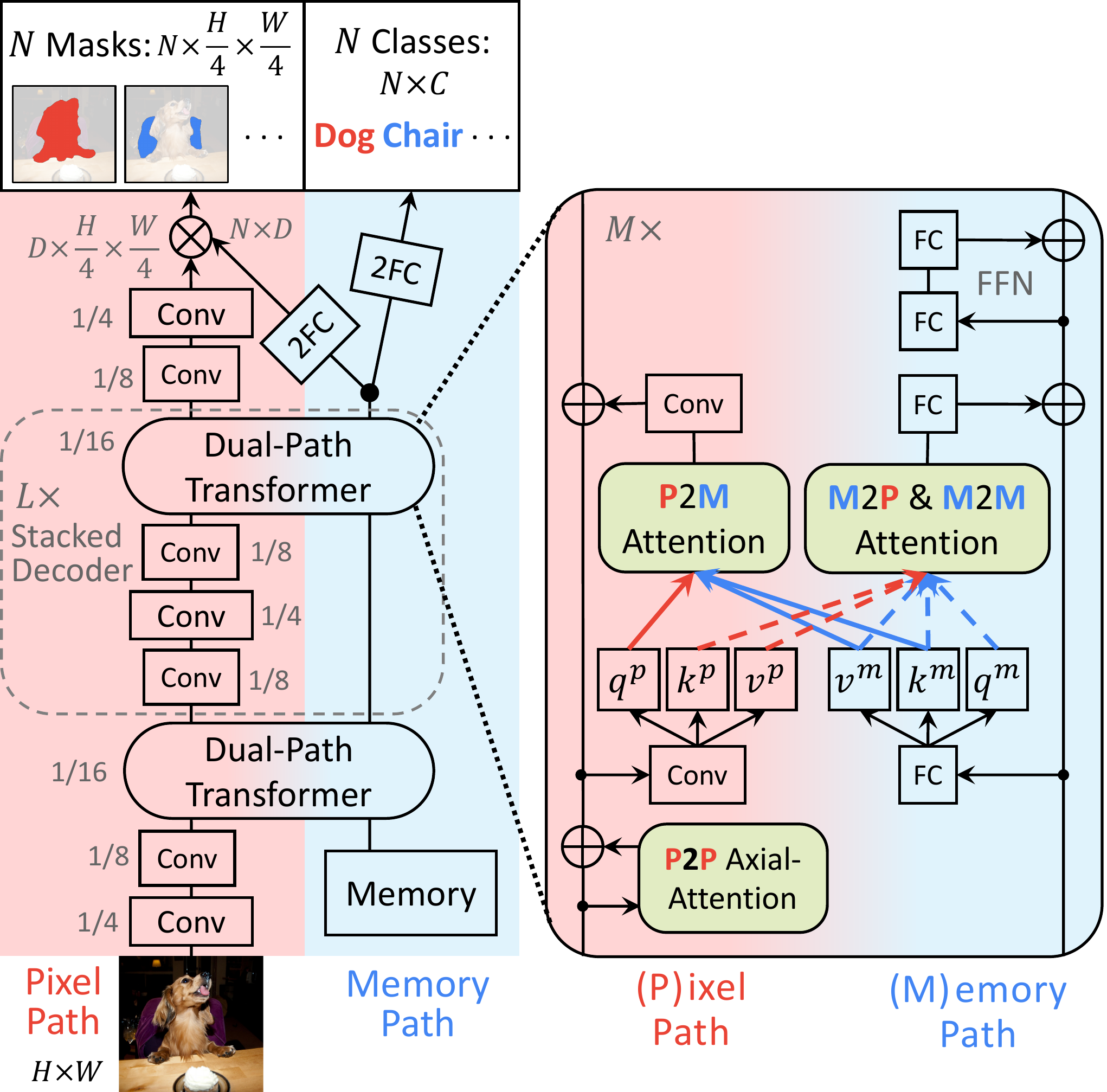}
    \vskip -13pt
    \subfigure[Overview of MaX-DeepLab]{\includegraphics[width=0.49\linewidth,height=0.0\linewidth]{figures/overview.pdf}\label{fig:overview}}
    \subfigure[Dual-path transformer block]{\includegraphics[width=0.49\linewidth,height=0.0\linewidth]{figures/overview.pdf}\label{fig:block}}
    \vskip \abovefigcapmargin
    \caption{\subref{fig:overview} An image and a global memory are fed into a dual-path transformer, which directly predicts a set of masks and classes (residual connections omitted). \subref{fig:block} A dual-path transformer block is equipped with all 4 types of attention between the two paths.}
    \label{fig:architecture}
    \vspace{\belowfigcapmargin}
\end{figure}
As shown in \figref{fig:architecture}, MaX-DeepLab architecture includes a dual-path transformer, a stacked decoder, and output heads that predict the masks and classes.
\vspace{\paramargin}\paragraph{Dual-path transformer.}
Instead of stacking a transformer on top of a CNN~\cite{carion2020end}, we integrate the transformer and the CNN in a dual-path fashion, with bidirectional communication between the two paths. Specifically, we augment a 2D {\it pixel}-based CNN with a 1D global {\it memory}
of size $N$ (\ie, the total number of predictions)
and propose a transformer block as a drop-in replacement for any CNN block or an add-on for a pretrained CNN block.
Our transformer block enables all four possible types of communication between the 2D pixel-path CNN and the 1D memory-path: (1)~the traditional memory-to-pixel (\textit{M2P}) attention, (2)~memory-to-memory (\textit{M2M}) self-attention, (3)~pixel-to-memory (\textit{P2M}) feedback attention that allows pixels to read from the memory, and (4)~pixel-to-pixel (\textit{P2P}) self-attention, implemented as axial-attention blocks~\cite{huang2019ccnet,ho2019axial,wang2020axial}. We select axial-attention~\cite{wang2020axial} rather than global 2D attention~\cite{carion2020end,wang2018non,bello2019attention} for efficiency on high resolution feature maps. One could optionally approximate the pixel-to-pixel self-attention with a convolutional block that only allows local communication. This transformer design with a memory path besides the main CNN path is termed dual-path transformer. Unlike previous work~\cite{carion2020end}, it allows transformer blocks to be inserted anywhere in the backbone at any resolution. In addition, the \textit{P2M} feedback attention enables the pixel-path CNN to refine its feature given the memory-path features that encode mask information.

Formally, given a 2D input feature $x^p \in \mathbb{R}^{\hat{H} \times \hat{W} \times d_{in}}$ with height $\hat{H}$, width $\hat{W}$, channels $d_{in}$, and a 1D global memory feature $x^m \in \mathbb{R}^{N \times d_{in}}$ with length $N$ (\ie, the size of the prediction set). We compute pixel-path queries $q^p$, keys $k^p$, and values $v^p$, by learnable linear projections of the pixel-path feature map $x^p$ at each pixel. Similarly, $q^m$, $k^m$, $v^m$ are computed from $x^m$ with another set of projection matrices. The query (key) and value channels are $d_q$ and $d_v$, for both paths. Then, the output of feedback attention (\textit{P2M}), $y_a^p \in \mathbb{R}^{d_{out}}$, at pixel position $a$, is computed as
\eqnsm{pixel}{y_a^p = \sum_{n=1}^{N} \text{softmax}_{n}\left(q_a^p \cdot k_n^m\right)v_n^m \,,}
where the $\text{softmax}_{n}$ denotes a softmax function applied to the whole memory of length $N$. Similarly, the output of memory-to-pixel (\textit{M2P}) and memory-to-memory (\textit{M2M}) attention $y_b^m \in \mathbb{R}^{d_{out}}$, at memory position $b$, is
\eqngatsm{memory}{y_b^m = \sum_{n=1}^{\hat{H}\hat{W}+N} \text{softmax}_{n}\left(q_b^m \cdot k_n^{pm}\right)v_n^{pm}\,, \\
k^{pm}=\begin{bmatrix}k^p\\k^m\end{bmatrix}\,, \quad v^{pm}=\begin{bmatrix}v^p\\v^m\end{bmatrix}\,,}
where a single softmax is performed over the concatenated dimension of size $(\hat{H}\hat{W}, +N)$, inspired by ETC~\cite{ainslie2020etc}.

\vspace{\paramargin}\paragraph{Stacked decoder.}
Unlike previous work~\cite{cheng2019panoptic,wang2020axial} that uses a light-weight decoder, we explore stronger hourglass-style stacked decoders~\cite{ronneberger2015u,newell2016stacked,cheng2019spgnet}. As shown in \figref{fig:architecture}, our decoder is stacked $L$ times, traversing output strides (4, 8, and 16~\cite{deeplabv3plus2018,liu2019auto}) multiple times. At each decoding resolution, features are fused by simple summation after bilinear resizing. Then, convolutional blocks or transformer blocks are applied, before the decoder feature is sent to the next resolution. This stacked decoder is similar to feature pyramid networks~\cite{lin2017feature,liu2018path,tan2020efficientdet,qiao2020detectors}
designed for pyramidal anchor predictions~\cite{liu2016ssd}, but our purpose here is only to aggregate multi-scale features, \ie, intermediate pyramidal features are not directly used for prediction.
\vspace{\paramargin}\paragraph{Output heads.}
From the memory feature of length $N$, we
predict mask classes $\hp(c) \in \mathbb{R}^{N \times |\mathbb{C}|}$ with two fully-connected layers (2FC) and a softmax.
Another 2FC head predicts mask feature $f \in \mathbb{R}^{N \times D}$. Similarly, we employ two convolutions (2Conv) to produce a normalized feature $g \in \mathbb{R}^{D \times \frac{H}{4} \times \frac{W}{4}}$ from the decoder output at stride 4. Then, our mask prediction $\hatm$ is simply the multiplication of transformer feature $f$ and decoder feature $g$:
\eqnsm{mask_pred}{\hatm = \text{softmax}_{N} \left(f \cdot g\right) \in \mathbb{R}^{N \times \frac{H}{4} \times \frac{W}{4}} \,.}
In practice, we use batch norm~\cite{ioffe2015batch} on $f$ and ($f \cdot g$) to avoid deliberate initialization, and we bilinear upsample the mask prediction $\hatm$ to the original image resolution. Finally, the combination $\{(\hatm_i, \hp_i(c))\}_{i=1}^N$ is our mask transformer output to generate panoptic results as introduced in \secref{sec:model}.

Our mask prediction head is inspired by CondInst~\cite{tian2020conditional} and SOLOv2~\cite{wang2020solov2}, which extend dynamic convolution~\cite{jia2016dynamic,yang2019condconv} to instance segmentation. However, unlike our end-to-end method, these methods require hand-designed object centers and assignment rules for instance segmentation, and a thing-stuff merging module for panoptic segmentation.

\subsection{Auxiliary losses}
In addition to the PQ-style loss (\secref{sec:pqloss}), we find it beneficial to incorporate auxiliary losses in training. Specifically, we propose a pixel-wise instance discrimination loss that helps cluster decoder features into instances. We also use a per-pixel mask-ID cross-entropy loss that classifies each pixel into $N$ masks, and a semantic segmentation loss. Our total loss function thus consists of the PQ-style loss $\pqloss$ and these three auxiliary losses.

\vspace{0.5\paramargin}\paragraph{Instance discrimination.}
We use a per-pixel instance discrimination loss to help the learning of the feature map $g \in \mathbb{R}^{D \times \frac{H}{4} \times \frac{W}{4}}$. Given a downsampled ground truth mask $m_i \in {\{0,1\}}^{\frac{H}{4} \times \frac{W}{4}}$, we first compute a normalized feature embedding $t_{i,:} \in \mathbb{R}^{D}$ for each annotated mask by averaging the feature vectors $g_{:,h,w}$ inside the mask $m_i$:
\eqnsm{mask_embedding}{t_{i,:} = \frac{\sum_{h,w} m_{i,h,w} \cdot g_{:,h,w}}{||\sum_{h,w} m_{i,h,w} \cdot g_{:,h,w}||}\,, \; i=1,2,\dots, K\,.}
This gives us $K$ instance embeddings $\{t_{i,:}\}_{i=1}^K$ representing $K$ ground truth masks.
Then, we let each pixel feature $g_{:,h,w}$ perform an instance discrimination task, \ie, each pixel should correctly identify which mask embedding (out of $K$) it belongs to, as annotated by the ground truth masks. The contrastive loss at a pixel $(h, w)$ is written as:
\eqnsm{instdis}{\insloss_{h,w}=-\log \frac{\sum_{i=1}^K m_{i,h,w} \exp{\left(t_{i,:} \cdot g_{:,h,w} / \tau\right)}}{\sum_{i=1}^K \exp{\left(t_{i,:} \cdot g_{:,h,w} / \tau\right)}}\,,}
where $\tau$ denotes the temperature, and note that $m_{i,h,w}$ is non-zero only when pixel $(h,w)$ belongs to the ground truth mask $m_i$. In practice, this per-pixel loss is applied to all instance pixels in an image, encouraging features from the same instance to be similar and features from different instances to be distinct, in a contrastive fashion, which is exactly the property required for instance segmentation.

Our instance discrimination loss is inspired by previous works~\cite{wu2018unsupervised,wu2018improving,hwang2019segsort,chen2020simple,he2020momentum,khosla2020supervised}. However, they discriminate instances either unsupervisedly or with image classes~\cite{khosla2020supervised}, whereas we perform a pixel-wise instance discrimination task, as annotated by panoptic segmentation ground truth.

\vspace{\paramargin}\paragraph{Mask-ID cross-entropy.}
In \equref{eq:pixelargmax}, we describe how we infer the mask-ID map given our mask prediction. In fact, we can train this per-pixel classification task by applying a cross-entropy loss on it. This is consistent with the literature~\cite{isensee2018nnu,taghanaki2019combo,carion2020end} that uses a cross-entropy loss together with a dice loss~\cite{milletari2016v} to learn better segmentation masks.

\vspace{\paramargin}\paragraph{Semantic segmentation.}
We also use an auxiliary semantic segmentation loss to help capture per pixel semantic feature. Specifically, we apply a semantic head~\cite{cheng2019panoptic} on top of the backbone if no stacked decoder is used (\ie, $L=0$). Otherwise, we connect the semantic head to the first decoder output at stride 4, because we find it helpful to separate the final mask feature $g$ with semantic segmentation.

\section{Experiments}
\label{sec:exp}
We report our main results on COCO, comparing with state-of-the-art methods. Then, we provide a detailed ablation study on the architecture variants and losses. Finally, we analyze how MaX-DeepLab works with visualizations.

\vspace{\paramargin}\paragraph{Technical details.}
Most of our default settings follow Axial-DeepLab~\cite{wang2020axial}.
Specifically, we train our models with 32 TPU cores for 100k (400k for main results) iterations (54 epochs),
a batch size of 64, 
Radam \cite{liu2019variance} Lookahead \cite{zhang2019lookahead}, 
a `poly' schedule learning rate of $10^{-3}$ ($3\times10^{-4}$ for MaX-DeepLab-L), a backbone learning rate multiplier of 0.1, a weight decay of $10^{-4}$, and a drop path rate~\cite{huang2016deep} of 0.2.
We resize and pad images to 641 $\times$ 641~\cite{cheng2019panoptic,wang2020axial} (1025 $\times$ 1025 for main results) for inference and M-Adds calculation. 
During inference, we set masks with class confidence below 0.7 to void and filter pixels with mask-ID confidence below 0.4. Finally, following previous work~\cite{xiong2019upsnet,cheng2019panoptic,wang2020axial}, we filter stuff masks with an area limit of 4096 pixels, and instance masks with a limit of 256 pixels.
In training, we set our PQ-style loss weight (\equref{eq:pqloss}, normalized by $N$) to 3.0, with $\alpha=0.75$. Our instance discrimination uses $\tau=0.3$, and a weight of 1.0. We set the mask-ID cross-entropy weight to 0.3, and semantic segmentation weight to 1.0. We use an output size $N=128$ and $D=128$ channels. We fill the initial memory with learnable weights~\cite{carion2020end} (more details and architectures in \secref{sec:appendix_details}).

\subsection{Main results}
We present our main results on COCO {\it val} set and {\it test-dev} set~\cite{lin2014microsoft}, with a small model, MaX-DeepLab-S, and a large model, MaX-DeepLab-L.

MaX-DeepLab-S augments ResNet-50~\cite{he2016deep} with axial-attention blocks~\cite{wang2020axial} in the last two stages. After pretaining, we replace the last stage with dual-path transformer blocks and use an $L=0$ (not stacked) decoder.
We match parameters and M-Adds to DETR-R101~\cite{carion2020end}, for fair comparison.

MaX-DeepLab-L stacks an $L=2$ decoder on top of Wide-ResNet-41~\cite{wrn2016wide,wu2019wider,chen2020naive}. And we replace all stride 16 residual blocks by our dual-path transformer blocks with wide axial-attention blocks~\cite{wang2020axial}.
This large variant is meant to be compared with state-of-the-art results.

\vspace{\paramargin}\paragraph{Val set.}
In \tabref{tab:coco_val}, we report our validation set results and compare with both box-based and box-free panoptic segmentation methods. As shown in the table, our {\it single-scale} MaX-DeepLab-S already outperforms all other {\it box-free} methods by a large margin of more than 4.5 \% PQ, no matter whether other methods use test time augmentation (TTA, usually flipping and multi-scale) or not. Specifically, it surpasses {\it single-scale} Panoptic-DeepLab by 8.7\% PQ, and {\it single-scale} Axial-DeepLab by 5.0\% PQ with similar M-Adds. We also compare MaX-DeepLab-S with DETR~\cite{carion2020end}, which is based on an end-to-end detector, in a controlled environment of similar number of parameters and M-Adds. Our MaX-DeepLab-S outperforms DETR~\cite{carion2020end} by 3.3\% PQ in this fair comparison. Next, we scale up MaX-DeepLab to a wider variant with stacked decoder, MaX-DeepLab-L. This scaling further improves the {\it single-scale} performance to 51.1\% PQ, outperforming {\it multi-scale} Axial-DeepLab \cite{wang2020axial} by 7.2\% PQ with similar inference M-Adds.

\vspace{\paramargin}\paragraph{Test-dev set.}
Our improvements on the {\it val} set transfers well to the test-dev set, as shown in \tabref{tab:coco_test}. On the test-dev set, we are able to compare with more competitive methods and stronger backbones equipped with group convolution~\cite{krizhevsky2012imagenet,xie2017aggregated}, deformable convolution~\cite{dai2017deformable}, or recursive backbone~\cite{liu2020cbnet,qiao2020detectors}, while we do not use these improvements in our model. In the regime of no TTA, our MaX-DeepLab-S outperforms Axial-DeepLab~\cite{wang2020axial} by 5.4\% PQ, and DETR~\cite{carion2020end} by 3.0\% PQ. Our MaX-DeepLab-L without TTA further attains 51.3\% PQ, surpassing Axial-DeepLab with TTA by 7.1\% PQ. This result also outperforms the best box-based method DetectoRS~\cite{qiao2020detectors} with TTA by 1.7\% PQ, closing the large gap between box-based and box-free methods on COCO for the first time. Our MaX-DeepLab sets a new state-of-the-art on COCO, even without using TTA.

\begin{savenotes}
    \begin{table}[!t]
    \setlength{\tabcolsep}{0.2em}
        \begin{center}
        \resizebox{\linewidth}{!}{
        \begin{tabular}{l|c|c|cc|ccc}
            \toprule[0.2em]
            Method & Backbone & TTA & Params & M-Adds & PQ & PQ\textsuperscript{Th} & PQ\textsuperscript{St} \\
            \toprule[0.2em]
            \multicolumn{8}{c}{Box-based panoptic segmentation methods}\\
            \midrule
            Panoptic-FPN~\cite{kirillov2019panoptic} & RN-101 & & & & 40.3 & 47.5 & 29.5 \\
            UPSNet~\cite{xiong2019upsnet} & RN-50 & & & & 42.5 & 48.5 & 33.4 \\
            Detectron2~\cite{wu2019detectron2} & RN-101 & & & & 43.0 & - & - \\
            UPSNet~\cite{xiong2019upsnet} & RN-50 & \ding{51} & & & 43.2 & 49.1 & 34.1 \\
            DETR~\cite{carion2020end} & RN-101 & & 61.8M & 314B\footnote{https://github.com/facebookresearch/detr} & 45.1 & 50.5 & 37.0 \\
            \midrule
            \multicolumn{8}{c}{Box-free panoptic segmentation methods}\\
            \midrule
            Panoptic-DeepLab~\cite{cheng2019panoptic} & X-71~\cite{chollet2016xception} & & 46.7M & 274B & 39.7 & 43.9 & 33.2 \\
            Panoptic-DeepLab~\cite{cheng2019panoptic} & X-71~\cite{chollet2016xception} & \ding{51} & 46.7M & 3081B & 41.2 & 44.9 & 35.7 \\
            Axial-DeepLab-L~\cite{wang2020axial} & AX-L~\cite{wang2020axial} & & 44.9M & 344B & 43.4 & 48.5 & 35.6 \\
            Axial-DeepLab-L~\cite{wang2020axial} & AX-L~\cite{wang2020axial} & \ding{51} & 44.9M & 3868B & 43.9 & 48.6 & 36.8 \\
            \midrule \midrule
            MaX-DeepLab-S & MaX-S & & 61.9M & 324B & 48.4 & 53.0 & 41.5 \\
            MaX-DeepLab-L & MaX-L & & 451M & 3692B & 51.1 & 57.0 & 42.2 \\
            \bottomrule[0.1em]
        \end{tabular}
        }
        \end{center}
        \vspace{\abovetabcapmargin}
        \caption{COCO val set. {\bf TTA:} Test-time augmentation
        }
        \label{tab:coco_val}
        \vspace{\belowtabcapmargin}
    \end{table}
    \end{savenotes}

    \begin{table}[!t]
    \setlength{\tabcolsep}{0.3em}
        \begin{center}
        \resizebox{\linewidth}{!}{
        \begin{tabular}{l|c|c|ccc}
            \toprule[0.2em]
            Method & Backbone & TTA & PQ & PQ\textsuperscript{Th} & PQ\textsuperscript{St} \\
            \toprule[0.2em]
            \multicolumn{6}{c}{Box-based panoptic segmentation methods} \\
            \midrule
            Panoptic-FPN~\cite{kirillov2019panoptic} & RN-101 & & 40.9 & 48.3 & 29.7  \\
            DETR~\cite{carion2020end} & RN-101 & & 46.0 & - & - \\
            UPSNet~\cite{xiong2019upsnet} & DCN-101 \cite{dai2017deformable} & \cmark & 46.6 & 53.2 & 36.7 \\
            DetectoRS~\cite{qiao2020detectors} & RX-101~\cite{xie2017aggregated} & \cmark & 49.6 & 57.8 & 37.1 \\
            \midrule
            \multicolumn{6}{c}{Box-free panoptic segmentation methods} \\
            \midrule
            Panoptic-DeepLab~\cite{cheng2019panoptic} & X-71~\cite{chollet2016xception,dai2017coco} & \cmark & 41.4 & 45.1 & 35.9 \\
            Axial-DeepLab-L~\cite{wang2020axial} & AX-L~\cite{wang2020axial} & & 43.6 & 48.9 & 35.6 \\
            Axial-DeepLab-L~\cite{wang2020axial} & AX-L~\cite{wang2020axial} & \cmark & 44.2 & 49.2 & 36.8 \\
            \midrule \midrule
            MaX-DeepLab-S & MaX-S & & 49.0 & 54.0 & 41.6 \\
            MaX-DeepLab-L & MaX-L & & 51.3 & 57.2 & 42.4 \\
            \bottomrule[0.1em]
        \end{tabular}
        }
        \end{center}
        \vspace{\abovetabcapmargin}
        \caption{COCO test-dev set. {\bf TTA:} Test-time augmentation}
        \label{tab:coco_test}
        \vspace{\belowtabcapmargin}
    \end{table}
\subsection{Ablation study}
\label{sec:ablation}
In this subsection, we provide more insights by teasing apart the effects of MaX-DeepLab components on the {\it val} set. We first define a default baseline setting and then vary each component of it: We augment Wide-ResNet-41~\cite{wrn2016wide,wu2019wider,chen2020naive} by applying dual-path transformer to all blocks at stride 16, enabling all four types of attention. For faster wall-clock training, we use an $L=0$ (not stacked) decoder and approximate {\it P2P} attention with convolutional blocks.

\vspace{\paramargin}\paragraph{Scaling.}
We first study the scaling of MaX-DeepLab in \tabref{tab:scaling}. We notice that replacing convolutional blocks with axial-attention blocks gives the most improvement. Further changing the input resolution to $1025\times1025$ improves the performance to 49.4\% PQ, with a short 100k schedule (54 epochs).
Stacking the decoder $L=1$ time improves 1.4\% PQ, but further scaling to $L=2$ starts to saturate. Training with more iterations helps convergence, but we find it not as necessary as DETR which is trained for 500 epochs.

\begin{table}[t]
    \setlength{\tabcolsep}{0.4em}
        \begin{center}
        \resizebox{\linewidth}{!}{
        \begin{tabular}{cccc|cc|ccc}
        \toprule[0.2em]
            Res & Axial & $L$ & Iter & Params & M-Adds & PQ & PQ\textsuperscript{Th} & PQ\textsuperscript{St} \\
            \toprule[0.2em]
            \rowcolor{Gray}
            641 & \xmark & 0 & 100k & 196M & 746B & 45.7 & 49.8 & 39.4 \\
            641 & \cmark & 0 & 100k & 277M & 881B & 47.8 & 51.9 & 41.5 \\
            \it 1025 & \xmark & 0 & 100k & 196M & 1885B & 46.1 & 50.7 & 39.1 \\
            \it 1025 & \cmark & 0 & 100k & 277M & 2235B & \bf 49.4 & \bf 54.5 & \bf 41.8 \\
            \midrule
            641 & \xmark & \it 1 & 100k & 271M & 1085B & 47.1 & 51.6 & 40.3 \\
            641 & \xmark & \it 2 & 100k & 347M & 1425B & 47.5 & 52.3 & 40.2 \\
            \midrule
            641 & \xmark & 0 & \it 200k & 196M & 746B & 46.9 & 51.5 & 40.0 \\
            641 & \xmark & 0 & \it 400k & 196M & 746B & 47.7 & 52.5 & 40.4 \\
            \bottomrule[0.1em]
        \end{tabular}
        }
        \end{center}
        \vspace{\abovetabcapmargin}
        \caption{Scaling MaX-DeepLab by using a larger input {\bf Res}olution, replacing convolutional blocks with {\bf Axial}-attention blocks, stacking decoder $\bm{L}$ times, and training with more {\bf Iter}ations.}
        \label{tab:scaling}
        \vspace{\belowtabcapmargin}
    \end{table}
    
\begin{table}[t]
    \setlength{\tabcolsep}{0.3em}
        \begin{center}
        \resizebox{\linewidth}{!}{
        \begin{tabular}{ccl|cc|ccc}
        \toprule[0.2em]
            P2M & M2M & Stride & Params & M-Adds & PQ & PQ\textsuperscript{Th} & PQ\textsuperscript{St} \\
            \toprule[0.2em]
            \rowcolor{Gray}
            \cmark & \cmark & 16 & 196M & 746B & 45.7 & 49.8 & 39.4 \\ 
             & \cmark & 16 & 188M & 732B & 45.0 & 48.9 & 39.2 \\ 
             \cmark & & 16 & 196M & 746B & 45.1 & 49.3 & 38.9 \\ 
             & & 16 & 186M & 731B & 44.7 & 48.5 & 39.0 \\ 
            \midrule
            \cmark & \cmark & 16 \& 8 & 220M & 768B & \bf 46.7 & \bf 51.3 & \bf 39.7 \\
            \cmark & \cmark & 16 \& 8 \& 4 & 234M & 787B & 46.3 & 51.1 & 39.0 \\
            \bottomrule[0.1em]
        \end{tabular}
        }
        \end{center}
        \vspace{\abovetabcapmargin}
        \caption{Varying transformer {\bf P2M} feedback attention, {\bf M2M} self-attention, and the {\bf Stride} where we apply the transformer.}
        \label{tab:transformer}
        \vspace{\belowtabcapmargin}
    \end{table}
    
\begin{figure*}[t]
    \centering
    \subfigure[Validation PQ]{\includegraphics[width=0.19\textwidth]{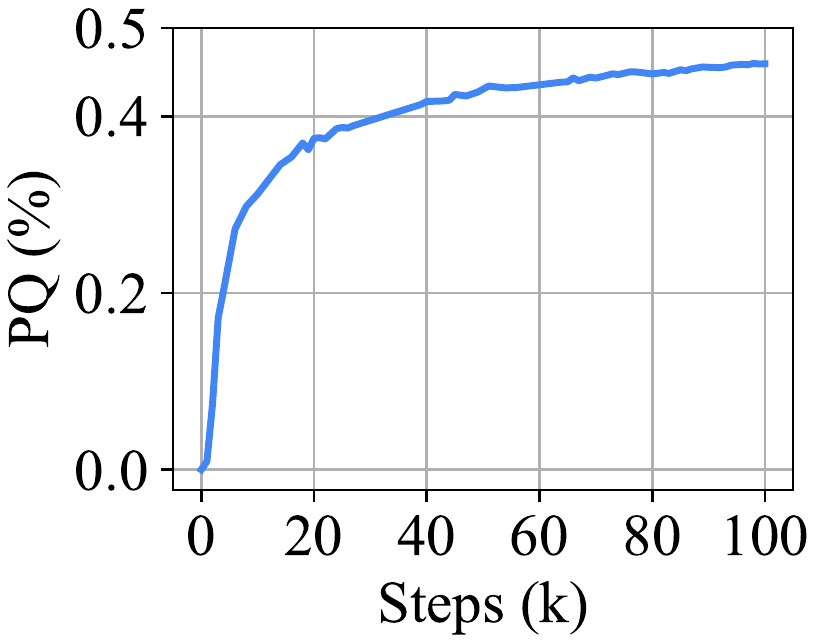}\label{fig:valpq}}
    \subfigure[Matched class confidence ]{\includegraphics[width=0.19\textwidth]{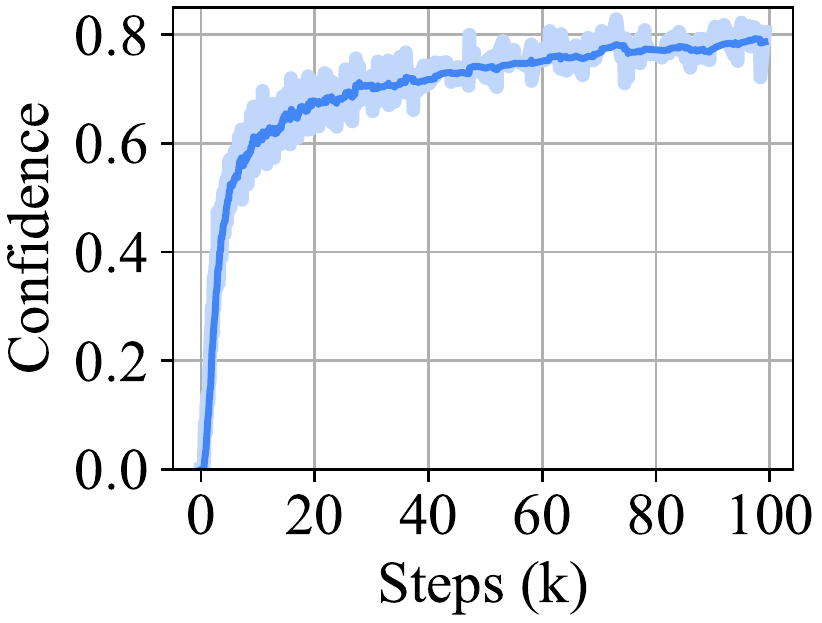}\label{fig:classsim}}
    \subfigure[Matched mask dice ]{\includegraphics[width=0.19\textwidth]{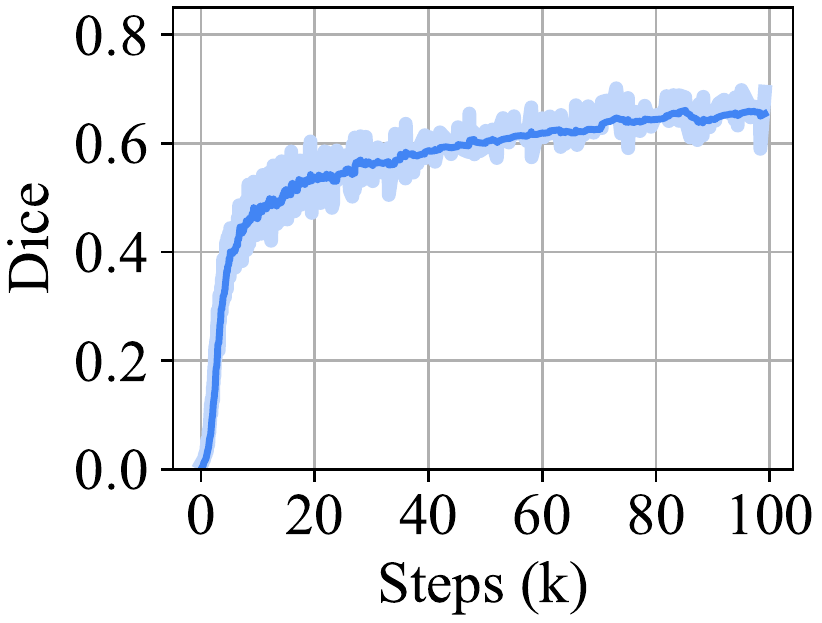}\label{fig:masksim}}
    \subfigure[Instance discrimination ]{\includegraphics[width=0.19\textwidth]{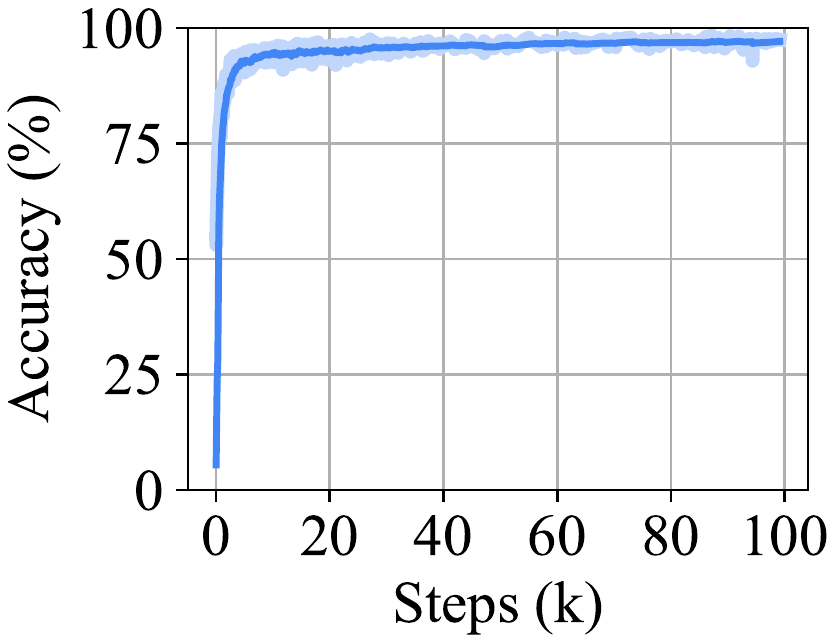}\label{fig:instdis}}
    \subfigure[Mask-ID prediction ]{\includegraphics[width=0.19\textwidth]{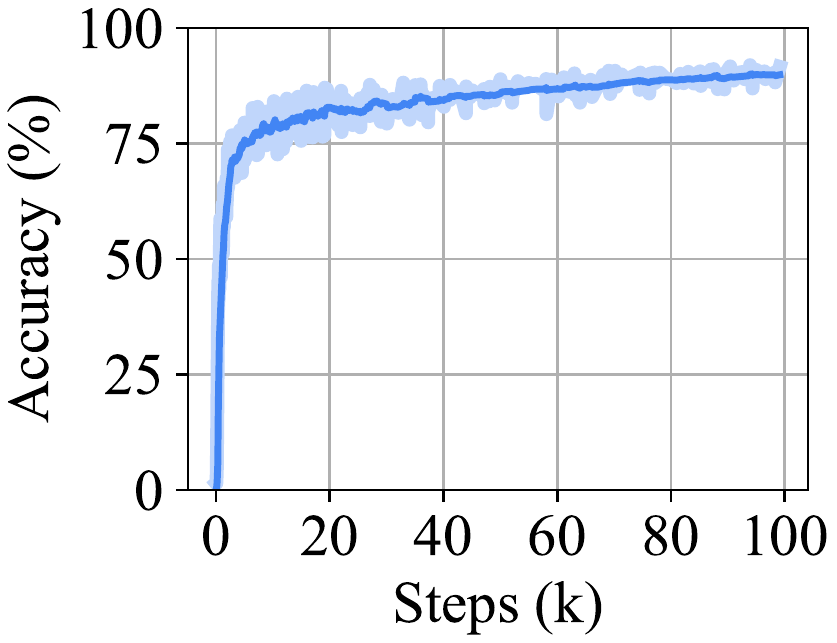}\label{fig:maskid}}
    \vskip \abovefigcapmargin
    \caption{Training curves for \subref{fig:valpq}~validation PQ, \subref{fig:classsim}~average class confidence, $\hp_{\hat{\sigma}(i)}(c_i)$, of matched masks, \subref{fig:masksim}~average mask dice, ${\rm Dice}(m_i, \hatm_{\hat{\sigma}(i)})$, of matched masks, \subref{fig:instdis}~per-pixel instance discrimination accuracy, and \subref{fig:maskid}~per-pixel mask-ID prediction accuray.}
    \label{fig:curves}
    \vspace{\belowfigcapmargin}
\end{figure*}

\vspace{\paramargin}\paragraph{Dual-path transformer.}
Next, we vary attention types of our dual-path transformer and the stages (strides) where we apply transformer blocks. Note that we always apply {\it M2P} attention that attaches the transformer to the CNN. And {\it P2P} attention is already ablated above.
As shown in \tabref{tab:transformer}, removing our {\it P2M} feedback attention causes a drop of 0.7\% PQ. On the other hand, we find MaX-DeepLab robust (-0.6\% PQ) to the removal of {\it M2M} self-attention. We attribute this robustness to our non-overlapping mask formulation. Note that DETR~\cite{carion2020end} relies on {\it M2M} self-attention to remove duplicated boxes. In addition, it is helpful (+1.0\% PQ) to apply transformer blocks to stride 8 also, which is impossible for DETR without our dual-path design. Pushing it further to stride 4 does not show more improvements.

\vspace{\paramargin}\paragraph{Loss ablation.}
Finally, we ablate our PQ-style loss and auxiliary losses in \tabref{tab:losses}. We first switch our PQ-style similarity in \equref{eq:similarity} from $\text{RQ} \times \text{SQ}$ to $\text{RQ} + \text{SQ}$, which differs in the hungarian matching (\equref{eq:matching}) and removes dynamic loss weights in \equref{eq:pqlosspos}. We observe that $\text{RQ} + \text{SQ}$ works reasonably well, but $\text{RQ} \times \text{SQ}$ improves 0.8\% PQ on top of it, confirming the effect of our PQ-style loss in practice, besides its conceptual soundness.
Next, we vary auxiliary losses applied to MaX-DeepLab, without tuning loss weights for remaining losses.
Our PQ-style loss alone achieves a reasonable performance of 39.5\% PQ.
Adding instance discrimination significantly improves PQ\textsuperscript{Th}, showing the importance of a clustered feature embedding.
Mask-ID prediction shares the same target with the Dice term in \equref{eq:pqlosspos}, but helps focus on large masks when the Dice term is overwhelmed by small objects.
Combining both of the auxiliary losses leads to a large 5.6\% PQ gain.
Further multi-tasking with semantic segmentation improves 0.6\% PQ,
because its class-level supervision helps stuff classes but not instance-level discrimination for thing classes.

\begin{table}[t]
    \setlength{\tabcolsep}{0.3em}
        \begin{center}
        \resizebox{\linewidth}{!}{
        \begin{tabular}{c|ccc|ccccc}
        \toprule[0.2em]
            $\rm sim$ & InstDis & Mask & Sem & PQ & PQ\textsuperscript{Th} & PQ\textsuperscript{St} & SQ & RQ \\
            \toprule[0.2em]
            \rowcolor{Gray}
            RQ $\times$ SQ & \cmark & \cmark & \cmark & \bf 45.7 & 49.8 & \bf 39.4 & \bf 80.9 & \bf 55.3 \\
            RQ $+$ SQ & \cmark & \cmark & \cmark & 44.9 & 48.6 & 39.3 & 80.2 & 54.5 \\
            \midrule
            RQ $\times$ SQ & \cmark & \cmark & & 45.1 & \bf 50.1 & 37.6 & 80.6 & 54.5 \\
            RQ $\times$ SQ & & \cmark & & 43.3 & 46.4 & 38.6 & 80.1 & 52.6 \\
            RQ $\times$ SQ & \cmark & & & 42.6 & 48.1 & 34.1 & 80.0 & 52.0 \\
            RQ $\times$ SQ & & & & 39.5 & 41.8 & 36.1 & 78.9 & 49.0 \\
            \bottomrule[0.1em]
        \end{tabular}
        }
        \end{center}
        \vspace{\abovetabcapmargin}
        \caption{Varying the similarity metric $\mathbf{sim}$ and whether to apply the auxiliary {\bf Inst}ance {\bf Dis}crimination loss, {\bf Mask}-ID cross-entropy loss or the {\bf Sem}antic segmentation loss.}
        \label{tab:losses}
        \vspace{\belowtabcapmargin}
    \end{table}
\subsection{Analysis}
We provide more insights of MaX-DeepLab by plotting our training curves and visualizing the mask output head.

\vspace{\paramargin}\paragraph{Training curves.}
We first report the validation PQ curve in \figref{fig:valpq}, with our default ablation model. MaX-DeepLab converges quickly to around 46\% PQ within 100k iterations (54 epochs), 1/10 of DETR~\cite{carion2020end}.
In \figref{fig:classsim} and \figref{fig:masksim}, we plot the characteristics of all matched masks in an image.
The matched masks tend to have a better class correctness than mask correctness.
Besides, we report per-pixel accuracies for instance discrimination (\figref{fig:instdis}) and mask-ID prediction (\figref{fig:maskid}). We see that most pixels learn quickly to find their own instances (out of $K$) and predict their own mask-IDs (out of $N$). Only 10\% of all pixels predict wrong mask-IDs, but they contribute to most of the PQ error.

\begin{figure}[t]
    \centering
    \subfigure[Original image ]{\includegraphics[height=0.3\linewidth]{figures/original_image.jpg}}
    \subfigure[Decoder feature $g$ ]{\includegraphics[height=0.3\linewidth]{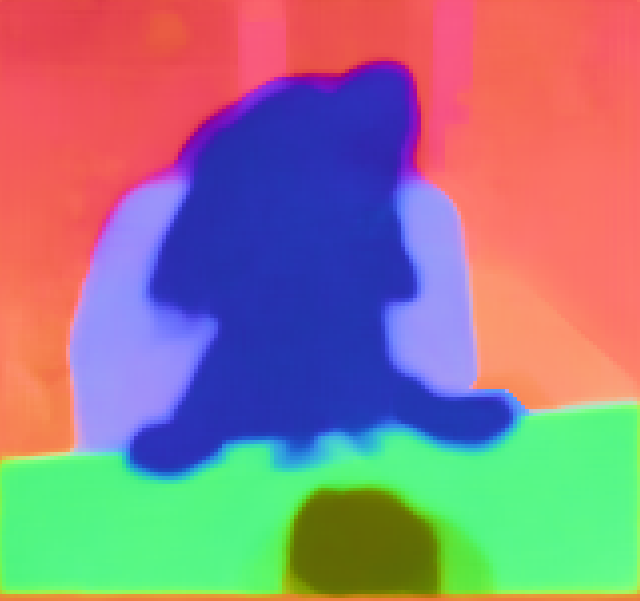}\label{fig:vis3decoder}}
    \hskip 3pt 
    \subfigure[Transformer output]{\hskip 3pt \includegraphics[height=0.3\linewidth]{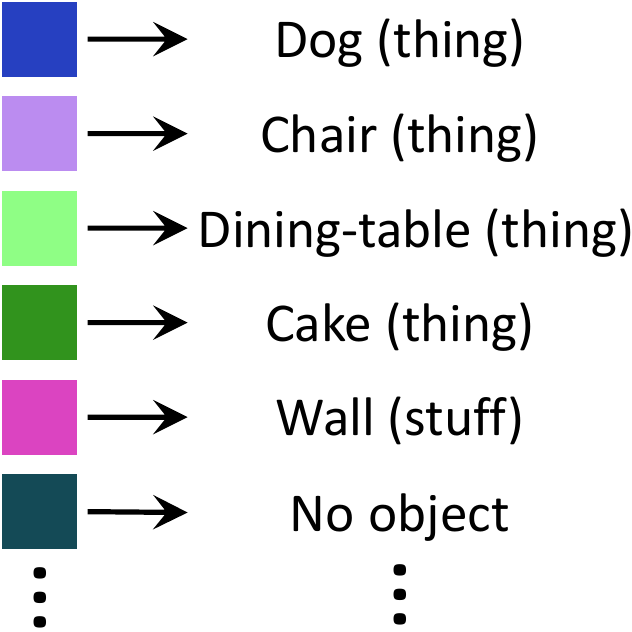}\label{fig:vis3transformer}}
    \vskip \abovefigcapmargin
    \caption{\subref{fig:vis3decoder}~Pixels of the same instance have similar colors (features), while pixels of different instances have distinct colors. \subref{fig:vis3transformer}~The transformer predicts mask colors (features) and classes.}
    \label{fig:vis3channels}
    \vspace{\belowfigcapmargin}
\end{figure}

\vspace{\paramargin}\paragraph{Visualization.}
In order to intuitively understand the normalized decoder output $g$, the transformer mask feature $f$, and how they are multiplied to generate our mask output $\hatm$, we train a MaX-DeepLab with $D=3$ and directly visualize the normalized features as RGB colors. As shown in \figref{fig:vis3channels}, the decoder feature $g$ assigns similar colors (or feature vectors) to pixels of the same mask, no matter the mask is a thing or stuff, while different masks are colored differently. Such effective instance discrimination (as colorization) facilitates our simple mask extraction with an inner product.
\FloatBarrier
\section{Conclusion}
In this work, we have shown for the first time that panoptic segmentation can be trained end-to-end. Our MaX-DeepLab directly predicts masks and classes with a mask transformer, removing the needs for many hand-designed priors such as object bounding boxes, thing-stuff merging, \etc. Equipped with a PQ-style loss and a dual-path transformer, MaX-DeepLab achieves the state-of-the-art result on the challenging COCO dataset, closing the gap between box-based and box-free methods for the first time.

\paragraph{Acknowledgments.}
We would like to thank Maxwell Collins and Sergey Ioffe for their feedbacks on the paper, Jiquan Ngiam for Hungarian Matching implementation, Siyuan Qiao for DetectoRS segmentation results, Chen Wei for instance discrimination insights, Jieneng Chen for dice loss comments and the support from Google Mobile Vision. This work is supported by Google Research Faculty Award.

\appendix
\section{Appendix}

\renewcommand\thefigure{\thesection.\arabic{figure}}
\renewcommand\thetable{\thesection.\arabic{table}}
\setcounter{figure}{0} 
\setcounter{table}{0} 

\renewcommand{\thetable}{A.\arabic{table}}

\begin{figure*}[!p]
    \centering
    \setlength\tabcolsep{2.12345pt}
    \resizebox*{0.99\linewidth}{0.974\textheight}{%
    \begin{tabular}{cccccc}
    \toprule[0.2em]
        Original Image & MaX-DeepLab-L & Axial-DeepLab~\cite{wang2020axial} & DetectoRS~\cite{qiao2020detectors} & DETR~\cite{carion2020end} & Ground Truth \\
        \midrule[0.07em]
         & Mask Transformer & Box-Free & Box-Based & Box Transformer & \\
         & 51.1\% PQ & 43.4\% PQ & 48.6\% PQ & 45.1\% PQ & \\
        \midrule[0.07em] \addlinespace[0.5em]
        \includegraphics[width=0.155\textwidth,height=0.2\textwidth]{figures_appendix/12748.jpg} &
        \includegraphics[width=0.155\textwidth,height=0.2\textwidth]{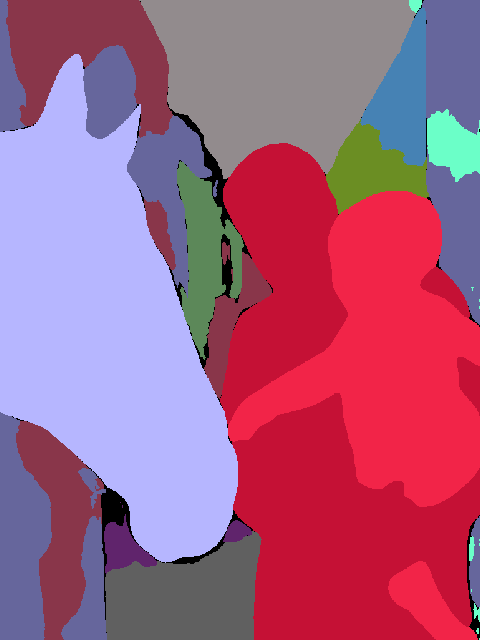} &
        \includegraphics[width=0.155\textwidth,height=0.2\textwidth]{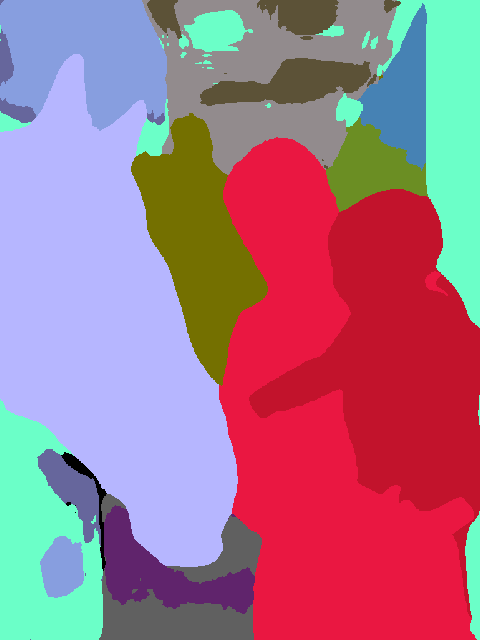} &
        \includegraphics[width=0.155\textwidth,height=0.2\textwidth]{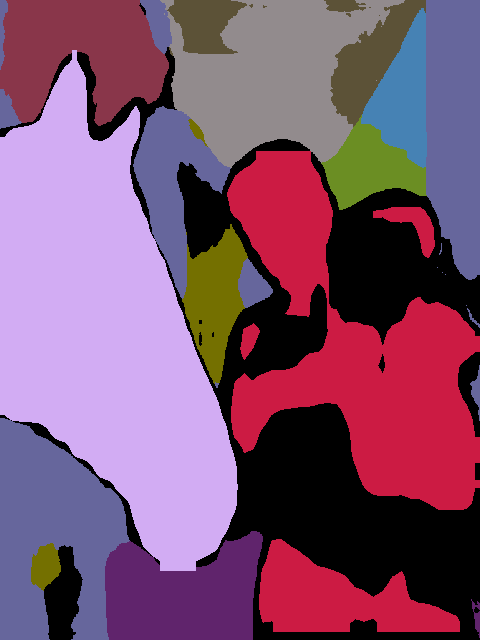} &
        \includegraphics[width=0.155\textwidth,height=0.2\textwidth]{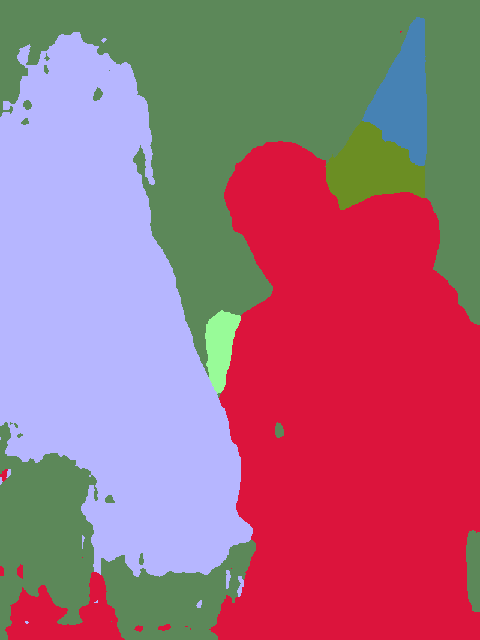} &
        \includegraphics[width=0.155\textwidth,height=0.2\textwidth]{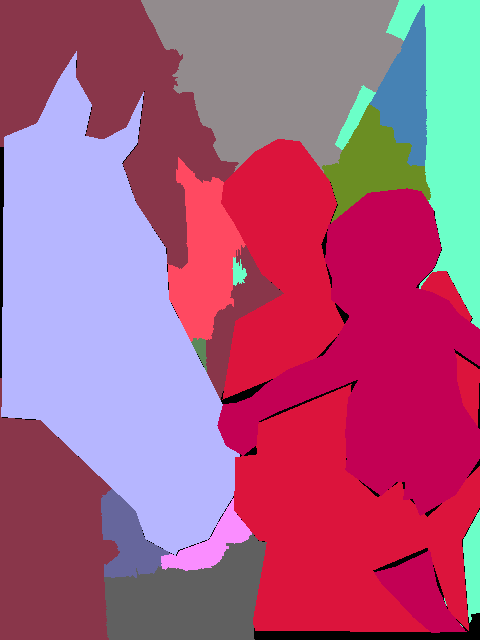} \\
        \multicolumn{6}{p{0.98\textwidth}}{\small MaX-DeepLab segments the baby with its occluded leg correctly. DetectoRS and DETR merge the two people into one instance, probably because the two people have similar bounding boxes. In addition, DETR introduces artifacts around the head of the horse.} \\
        \midrule[0.07em] \addlinespace[0.5em]
        \includegraphics[width=0.155\textwidth]{figures_appendix/127517.jpg} &
        \includegraphics[width=0.155\textwidth]{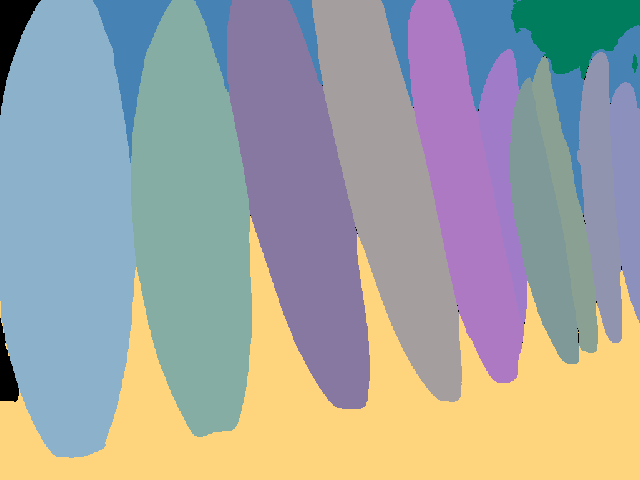} &
        \includegraphics[width=0.155\textwidth]{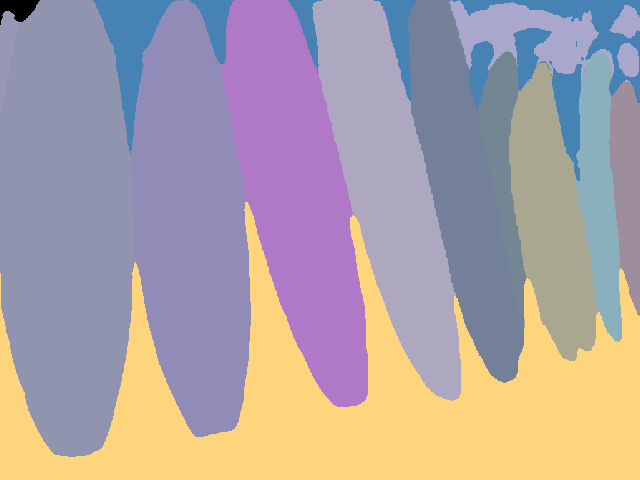} &
        \includegraphics[width=0.155\textwidth]{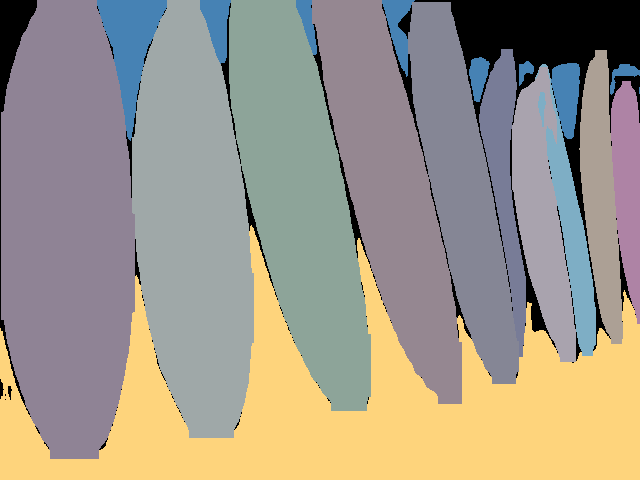} &
        \includegraphics[width=0.155\textwidth]{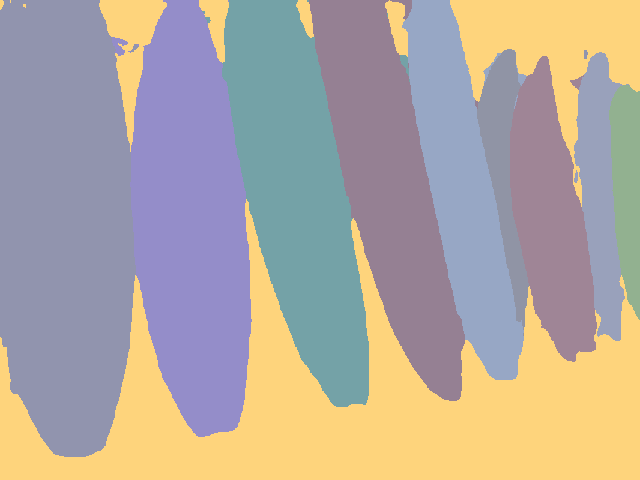} &
        \includegraphics[width=0.155\textwidth]{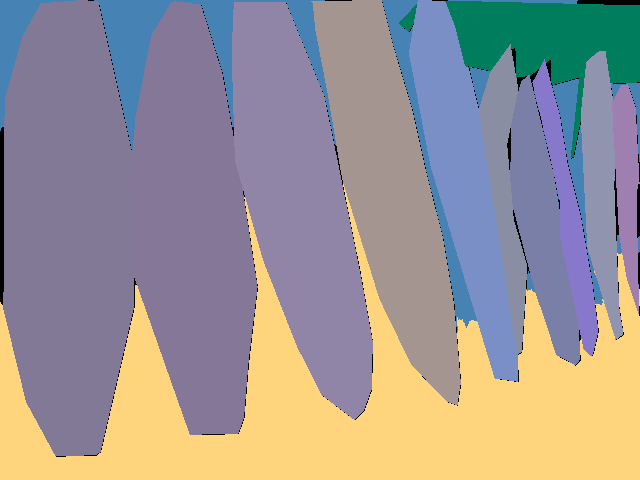} \\
        \includegraphics[width=0.155\textwidth]{figures_appendix/124975.jpg} &
        \includegraphics[width=0.155\textwidth]{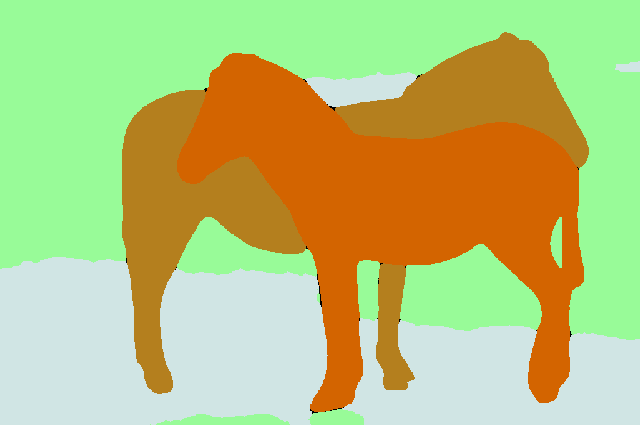} &
        \includegraphics[width=0.155\textwidth]{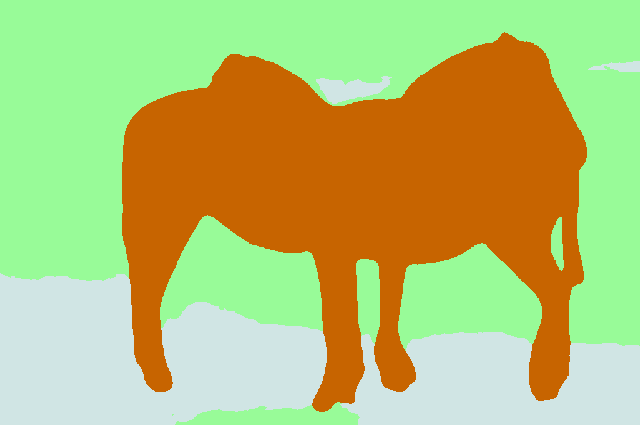} &
        \includegraphics[width=0.155\textwidth]{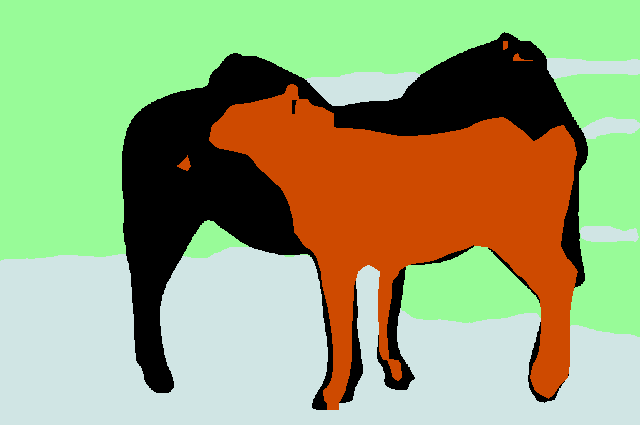} &
        \includegraphics[width=0.155\textwidth]{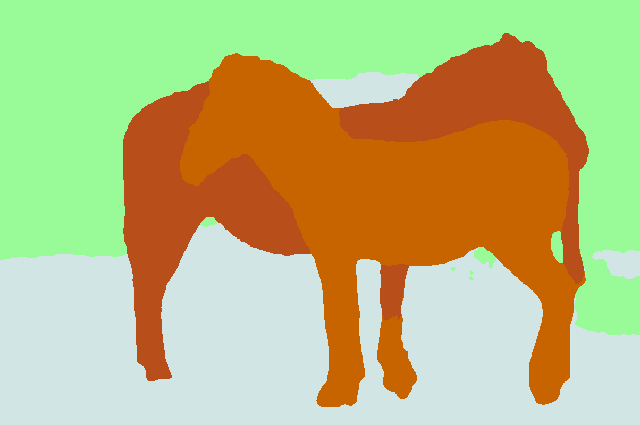} &
        \includegraphics[width=0.155\textwidth]{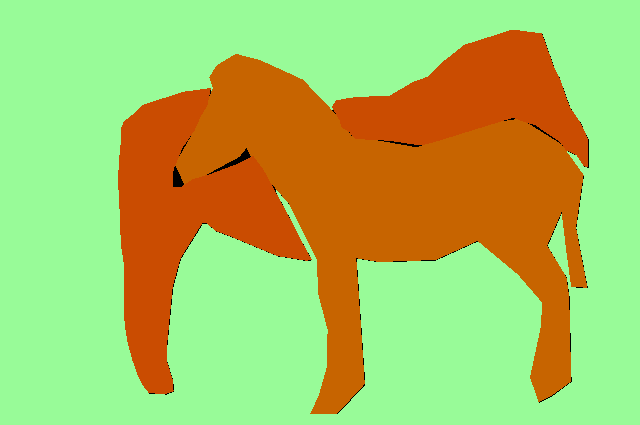} \\
        \includegraphics[width=0.155\textwidth]{figures_appendix/533493.jpg} &
        \includegraphics[width=0.155\textwidth]{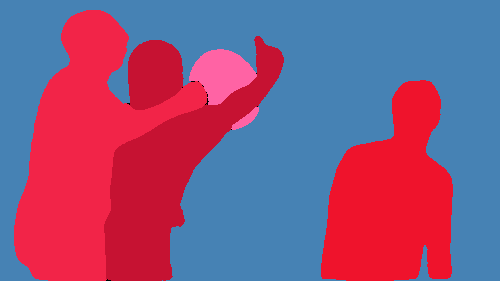} &
        \includegraphics[width=0.155\textwidth]{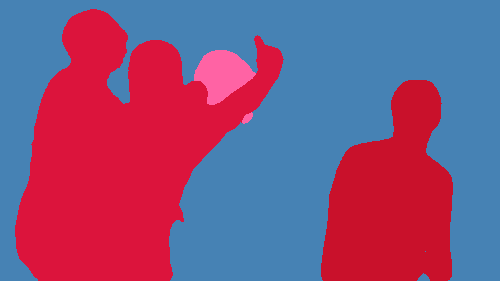} &
        \includegraphics[width=0.155\textwidth]{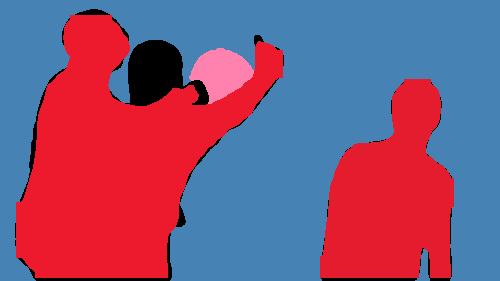} &
        \includegraphics[width=0.155\textwidth]{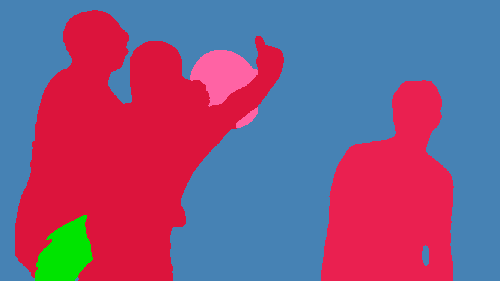} &
        \includegraphics[width=0.155\textwidth]{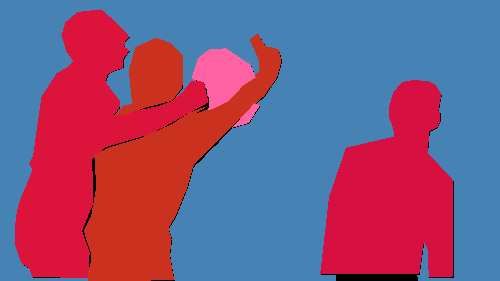} \\
        \multicolumn{6}{p{0.98\textwidth}}{\small MaX-DeepLab correctly segments all the boards, the zebras, and the people. All other methods fail in these challenging cases of similar bounding boxes and nearby object centers.} \\
        \midrule[0.07em] \addlinespace[0.5em]
        \includegraphics[width=0.155\textwidth]{figures_appendix/109055.jpg} &
        \includegraphics[width=0.155\textwidth]{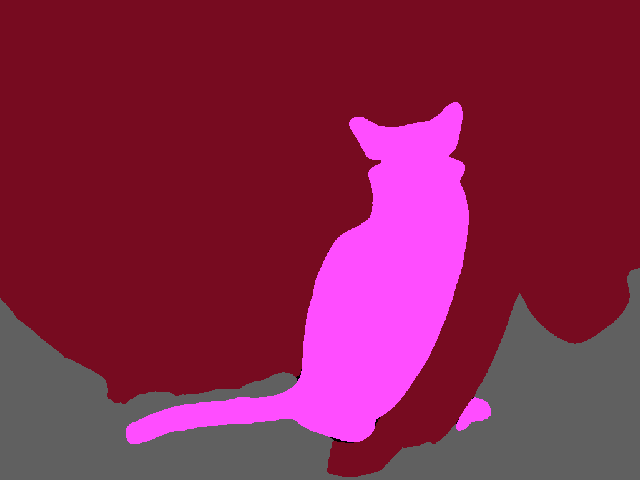} &
        \includegraphics[width=0.155\textwidth]{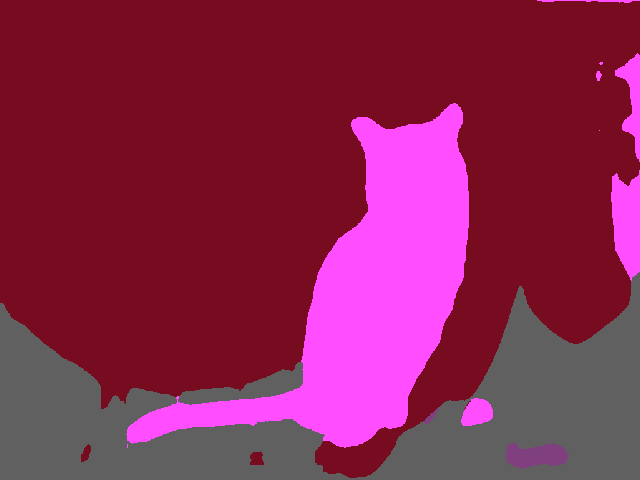} &
        \includegraphics[width=0.155\textwidth]{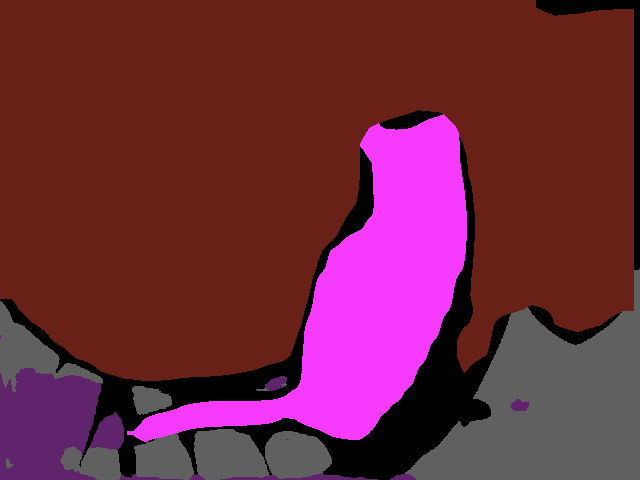} &
        \includegraphics[width=0.155\textwidth]{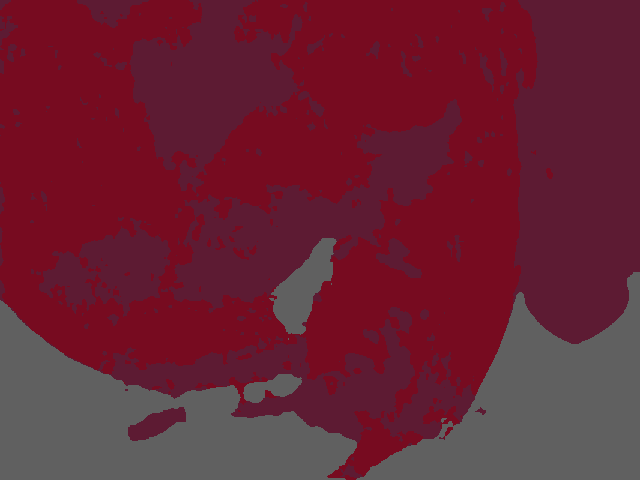} &
        \includegraphics[width=0.155\textwidth]{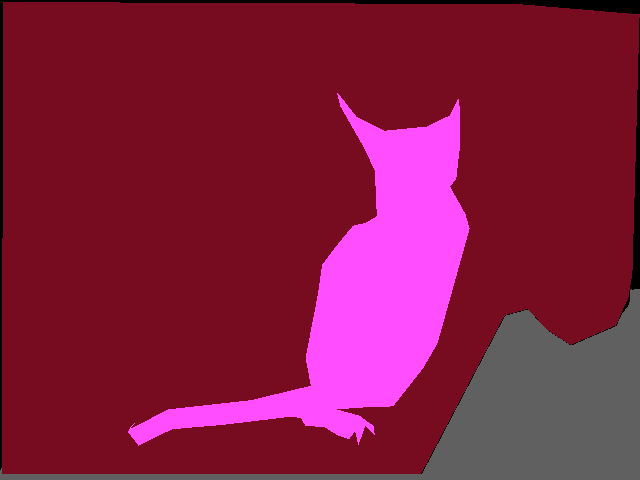} \\
        \multicolumn{6}{p{0.98\textwidth}}{\small MaX-DeepLab generates a high quality mask for the cat, arguably better than the ground truth. Axial-DeepLab predicts cat pixels on the right of the image, as the center of the cat is close to the center of the bike. And DETR misses the cat and introduces artifacts.} \\
        \midrule[0.07em] \addlinespace[0.5em]
        \includegraphics[width=0.155\textwidth]{figures_appendix/561958.jpg} &
        \includegraphics[width=0.155\textwidth]{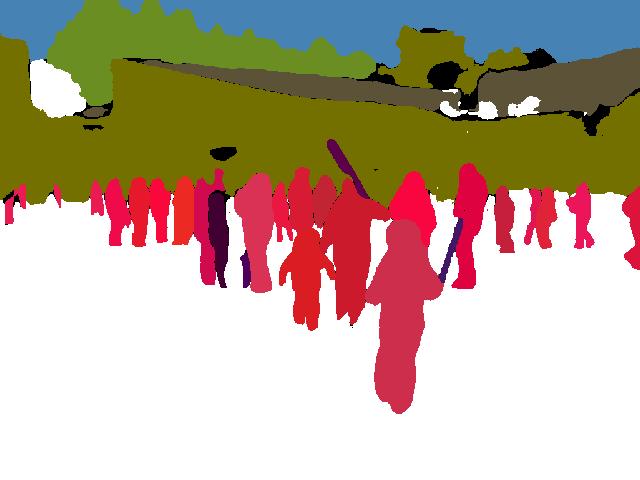} &
        \includegraphics[width=0.155\textwidth]{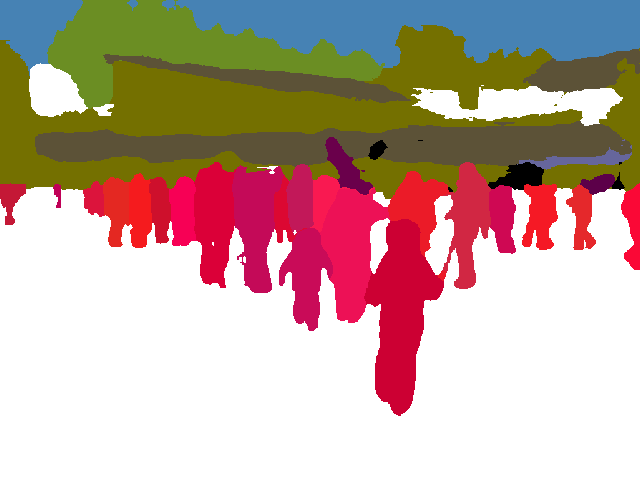} &
        \includegraphics[width=0.155\textwidth]{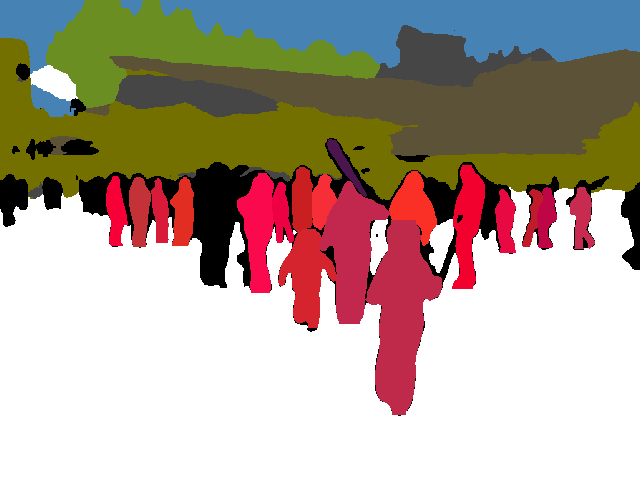} &
        \includegraphics[width=0.155\textwidth]{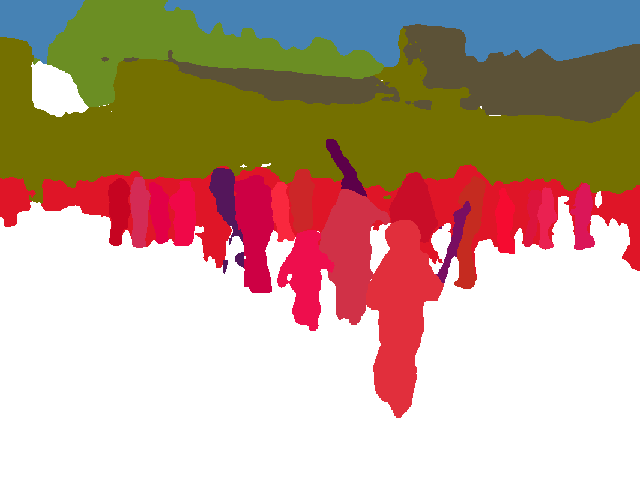} &
        \includegraphics[width=0.155\textwidth]{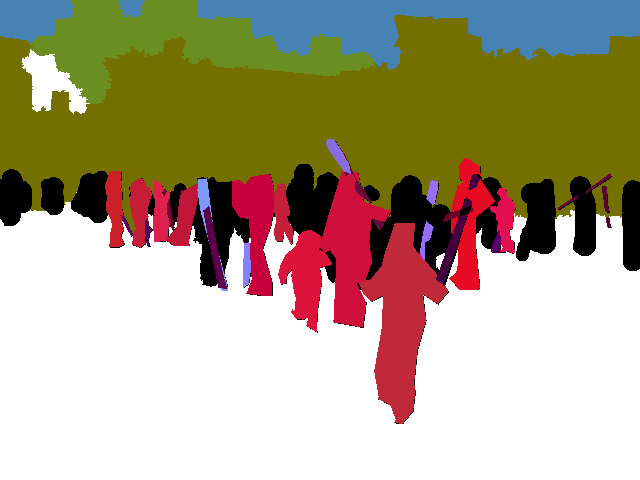} \\
        \includegraphics[width=0.155\textwidth]{figures_appendix/376322.jpg} &
        \includegraphics[width=0.155\textwidth]{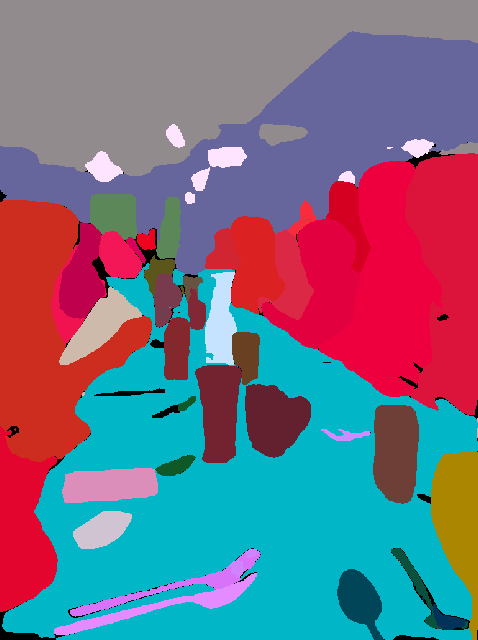} &
        \includegraphics[width=0.155\textwidth]{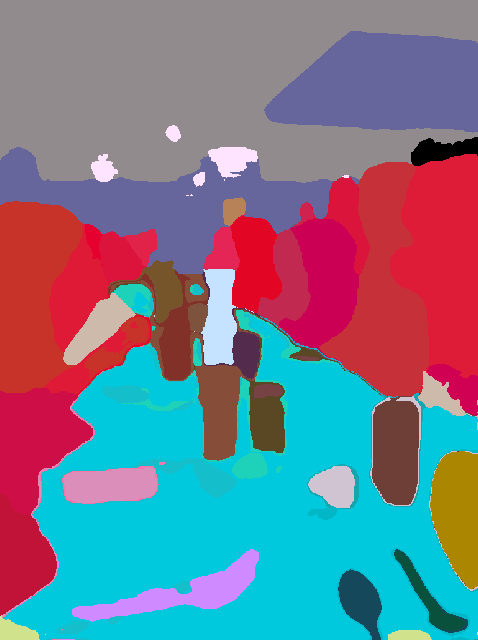} &
        \includegraphics[width=0.155\textwidth]{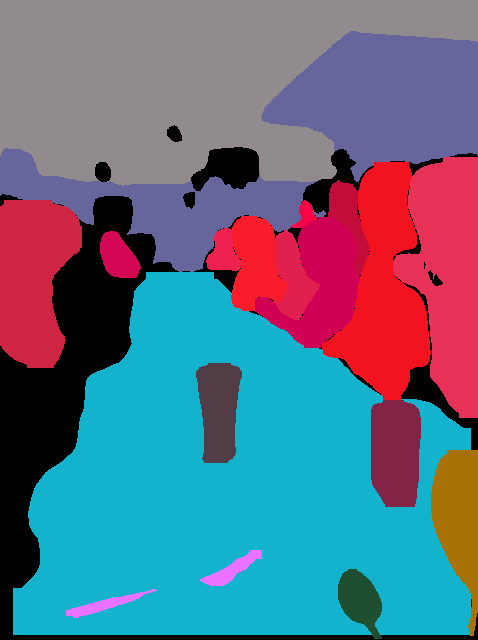} &
        \includegraphics[width=0.155\textwidth]{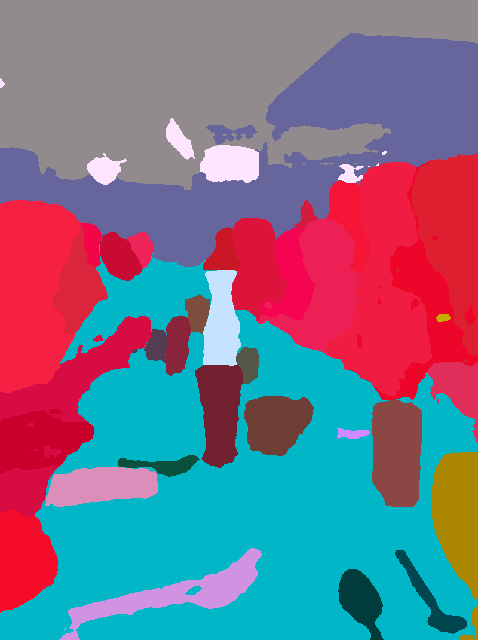} &
        \includegraphics[width=0.155\textwidth]{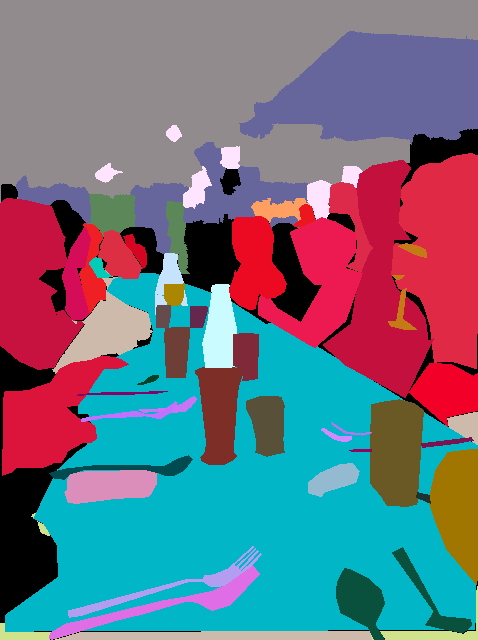} \\
        \multicolumn{6}{p{0.98\textwidth}}{\small MaX-DeepLab also performs well in the presence of many small instances.} \\
        \bottomrule[0.1em]
    \end{tabular}%
    }
    \vspace{7pt}
    \caption{Comparing  MaX-DeepLab with other representative methods on the COCO {\it val} set. (Colors modified for better visualization).}
    \label{fig:comparison}
\end{figure*}

\begin{figure*}[p]
    \centering
    \setlength\tabcolsep{2.12345pt}
    \resizebox*{0.99\linewidth}{!}{%
    \begin{tabular}{cccccc}
    \toprule[0.2em]
        Original Image & MaX-DeepLab-L & Axial-DeepLab~\cite{wang2020axial} & DetectoRS~\cite{qiao2020detectors} & DETR~\cite{carion2020end} & Ground Truth \\
        \midrule[0.07em]
         & Mask Transformer & Box-Free & Box-Based & Box Transformer & \\
         & 51.1\% PQ & 43.4\% PQ & 48.6\% PQ & 45.1\% PQ & \\
        \midrule[0.07em] \addlinespace[0.5em]
        \includegraphics[width=0.155\textwidth]{figures_appendix/24021.jpg} &
        \includegraphics[width=0.155\textwidth]{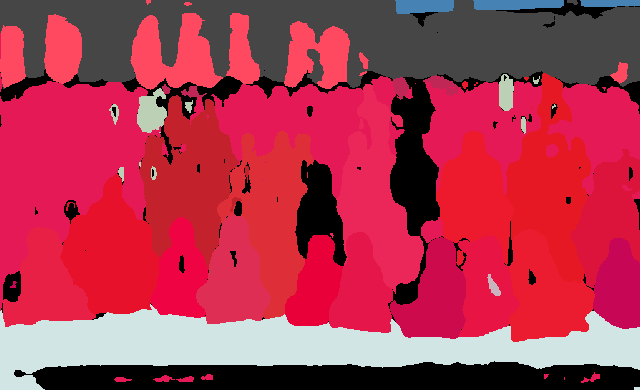} &
        \includegraphics[width=0.155\textwidth]{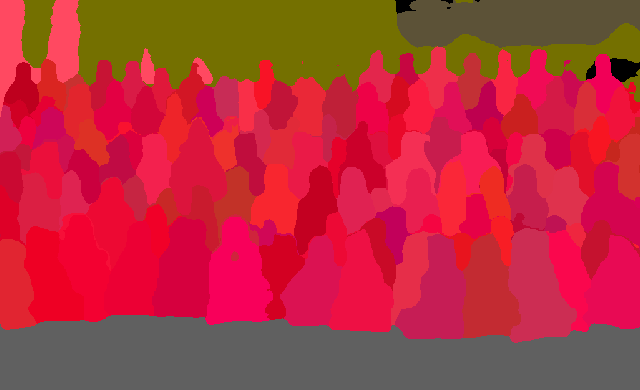} &
        \includegraphics[width=0.155\textwidth]{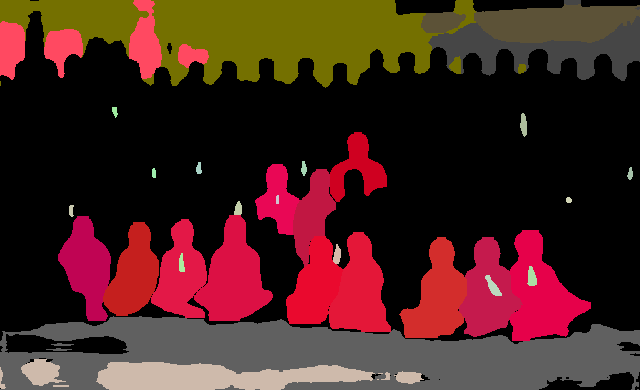} &
        \includegraphics[width=0.155\textwidth]{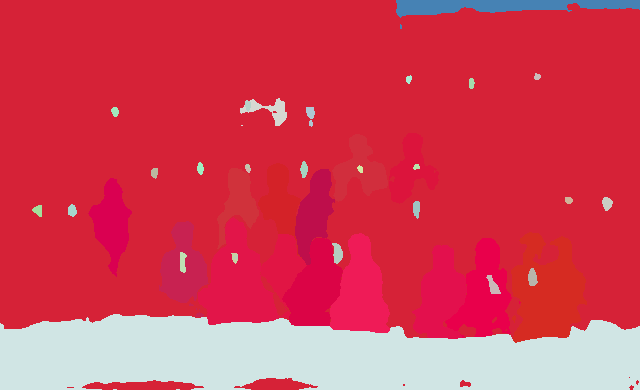} &
        \includegraphics[width=0.155\textwidth]{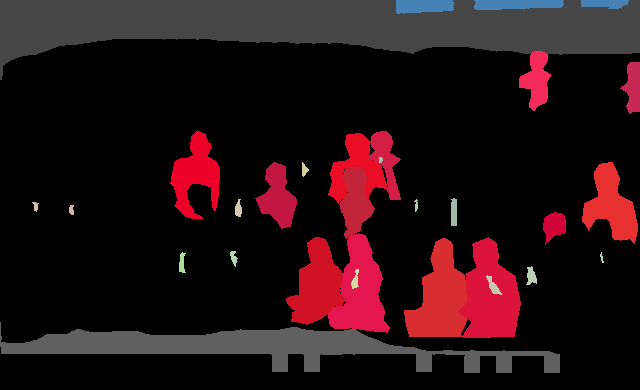} \\
        \multicolumn{6}{p{0.98\textwidth}}{\small Similar to DETR~\cite{carion2020end}, MaX-DeepLab fails typically when there are too many masks to segment in an image. This example contains more than 200 masks that should be predicted, mostly people and ties.} \\
        \midrule[0.07em] \addlinespace[0.5em]
        \includegraphics[width=0.155\textwidth]{figures_appendix/186345.jpg} &
        \includegraphics[width=0.155\textwidth]{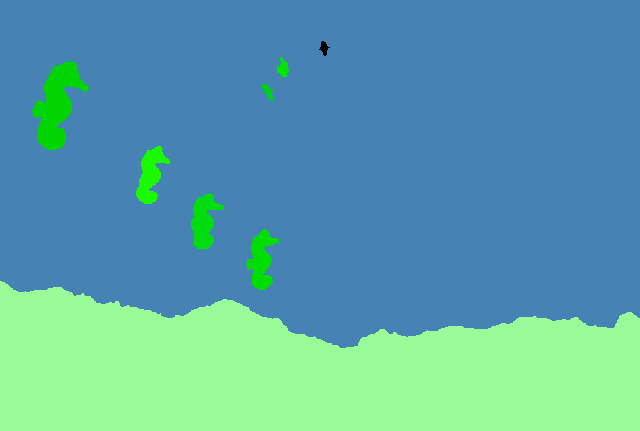} &
        \includegraphics[width=0.155\textwidth]{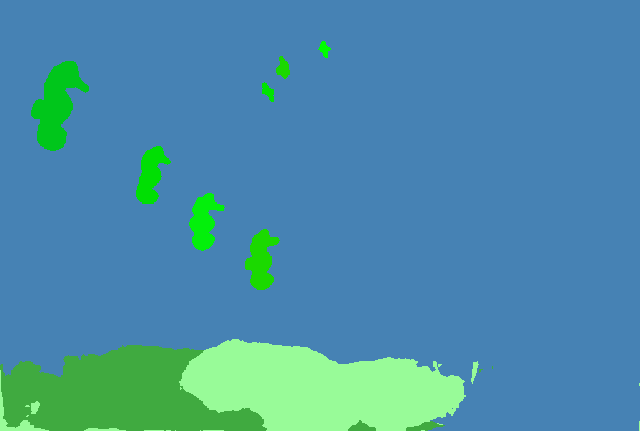} &
        \includegraphics[width=0.155\textwidth]{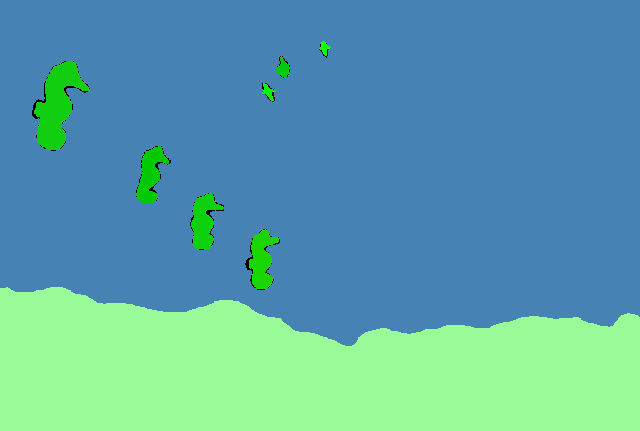} &
        \includegraphics[width=0.155\textwidth]{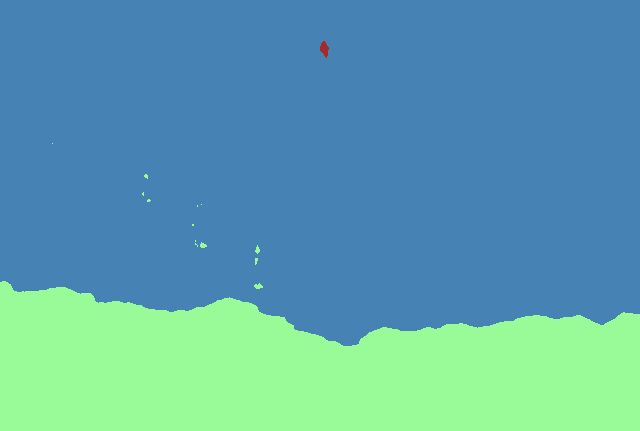} &
        \includegraphics[width=0.155\textwidth]{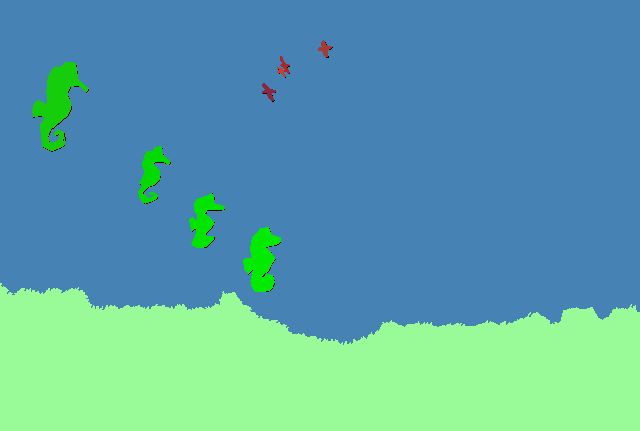} \\
        \multicolumn{6}{p{0.98\textwidth}}{\small In this failure case, MaX-DeepLab mistakes the birds for kites in the sky, probably because the birds are too small.} \\
        \bottomrule[0.1em]
    \end{tabular}%
    }
    \vspace{7pt}
    \caption{Failure cases of MaX-DeepLab on the COCO {\it val} set.}
    \label{fig:failures}
\end{figure*}

\begin{savenotes}
\begin{table*}[t]
    \setlength{\tabcolsep}{1.0em}
        \begin{center}
        \begin{tabular}{l|c|c|c|c|c}
            \toprule[0.2em]
            Method & Backbone & Input Size & Runtime (ms) & PQ [val] & PQ [test] \\
            \toprule[0.2em]
            \multicolumn{6}{c}{Fast Regime}\\
            \midrule
            Panoptic-DeepLab~\cite{cheng2019panoptic} & X-71~\cite{chollet2016xception} & 641$\times$641 & 74 & 38.9 & 38.8 \\
            MaX-DeepLab-S & MaX-S & 641$\times$641 & 67 & 46.4 & 46.7 \\
            \midrule
            \multicolumn{6}{c}{Slow Regime}\\
            \midrule
            DETR~\cite{carion2020end} & RN-101 & 1333$\times$800 & 128\footnote{https://github.com/facebookresearch/detr} & 45.1 & 46.0 \\
            Panoptic-DeepLab~\cite{cheng2019panoptic} & X-71~\cite{chollet2016xception} & 1025$\times$1025 & 132 & 39.7 & 39.6 \\
            MaX-DeepLab-S & MaX-S & 1025$\times$1025 & 131 & 48.4 & 49.0 \\
            \bottomrule[0.1em]
        \end{tabular}
        \end{center}
        \vspace{\abovetabcapmargin}
        \caption{End-to-end runtime. {\bf PQ [val]:} PQ (\%) on COCO val set. {\bf PQ [test]:} PQ (\%) on COCO test-dev set.
        }
        \label{tab:runtime}
        \vspace{\belowtabcapmargin}
    \end{table*}
    \end{savenotes}

\begin{figure*}[t]
    \centering
    \includegraphics[width=0.59\linewidth]{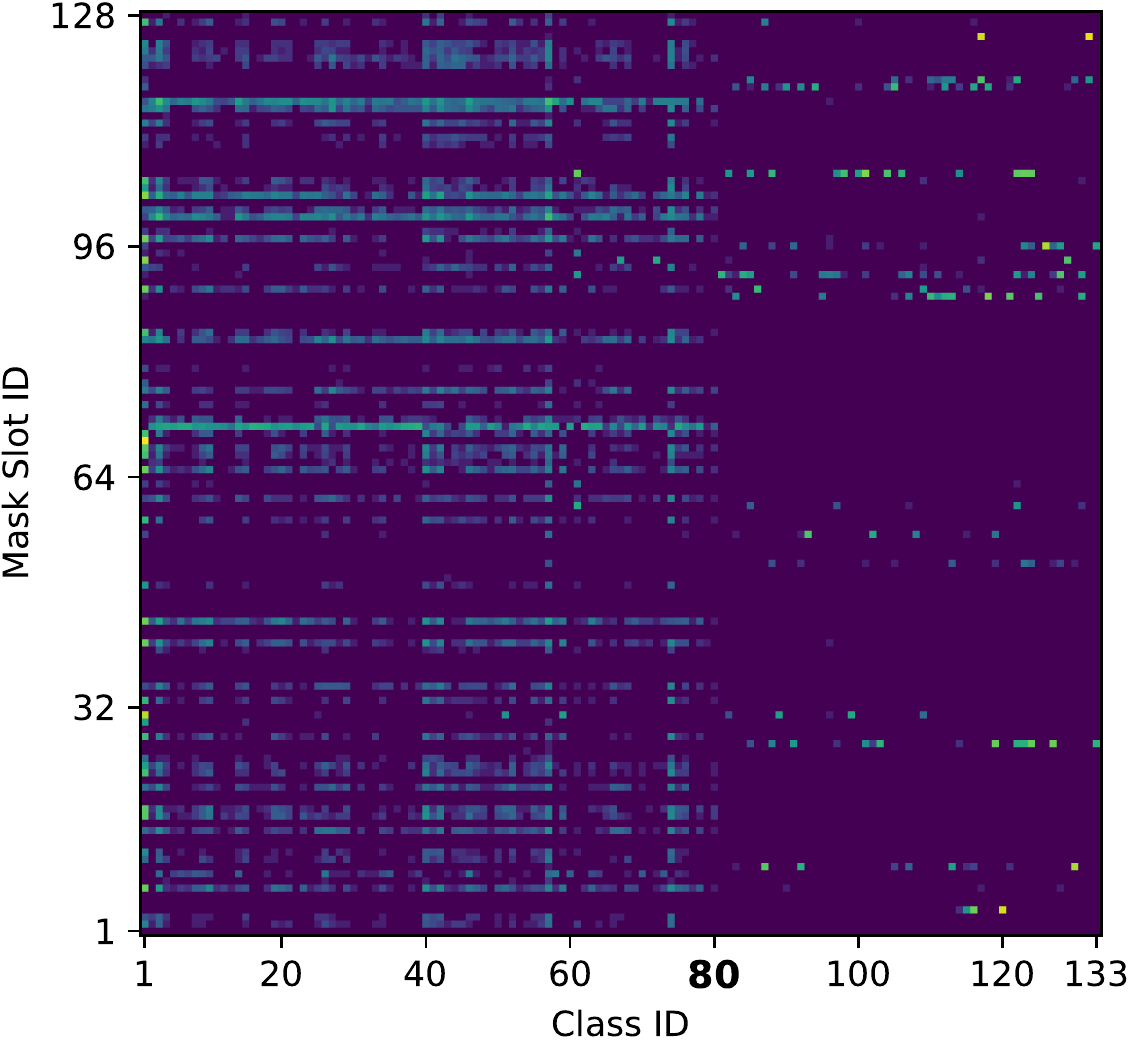}
    \caption{The joint distribution for our $N=128$ mask slots and 133 classes with 80 `thing' classes on the left and 53 `stuff' classes on the right. We observe that a few mask slots predict a lot of the masks. Some mask slots are used less frequently, probably only when there are a lot of objects in one image. Some other slots do not fire at all. In addition, we see automatic functional segregation between `thing' mask slots and `stuff' mask slots, with a few exceptions that can predict both thing and stuff masks.
    }
    \label{fig:distribution}
\end{figure*}

\begin{figure*}[t]
    \centering
    \setlength\tabcolsep{1.0pt}
    \begin{tabular}{ccccccccc}
    \midrule
    & \small Mask Slot 71 & \small Mask Slot 106 & \small Mask Slot 125 & \small Mask Slot 69 & \small Mask Slot 116 & \small Mask Slot 4 & \small Mask Slot 27 & \small Mask Slot 103\\
    \begin{tabular}{c}
    Most \\ Firings \\ (sorted)
    \end{tabular} &
    \raisebox{-.5\height}{\includegraphics[width=0.11\linewidth]{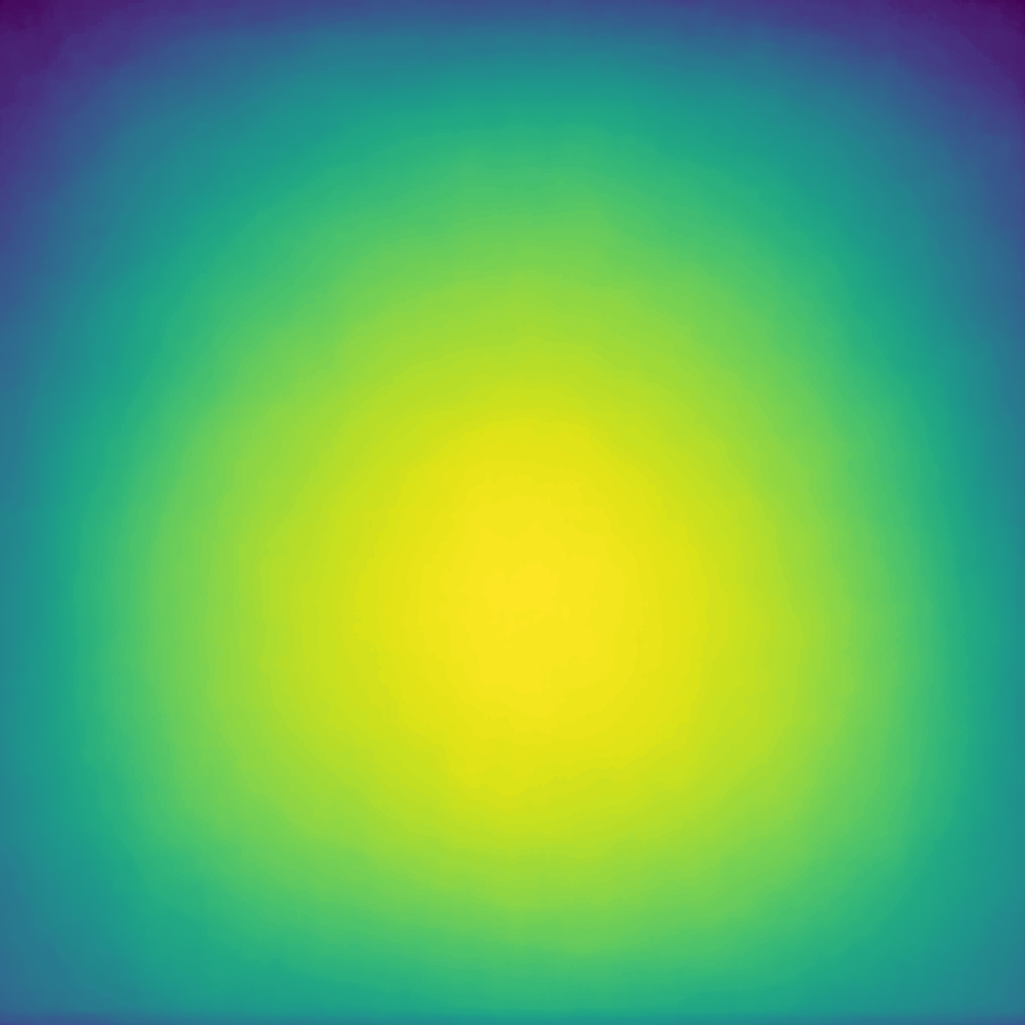}} &
    \raisebox{-.5\height}{\includegraphics[width=0.11\linewidth]{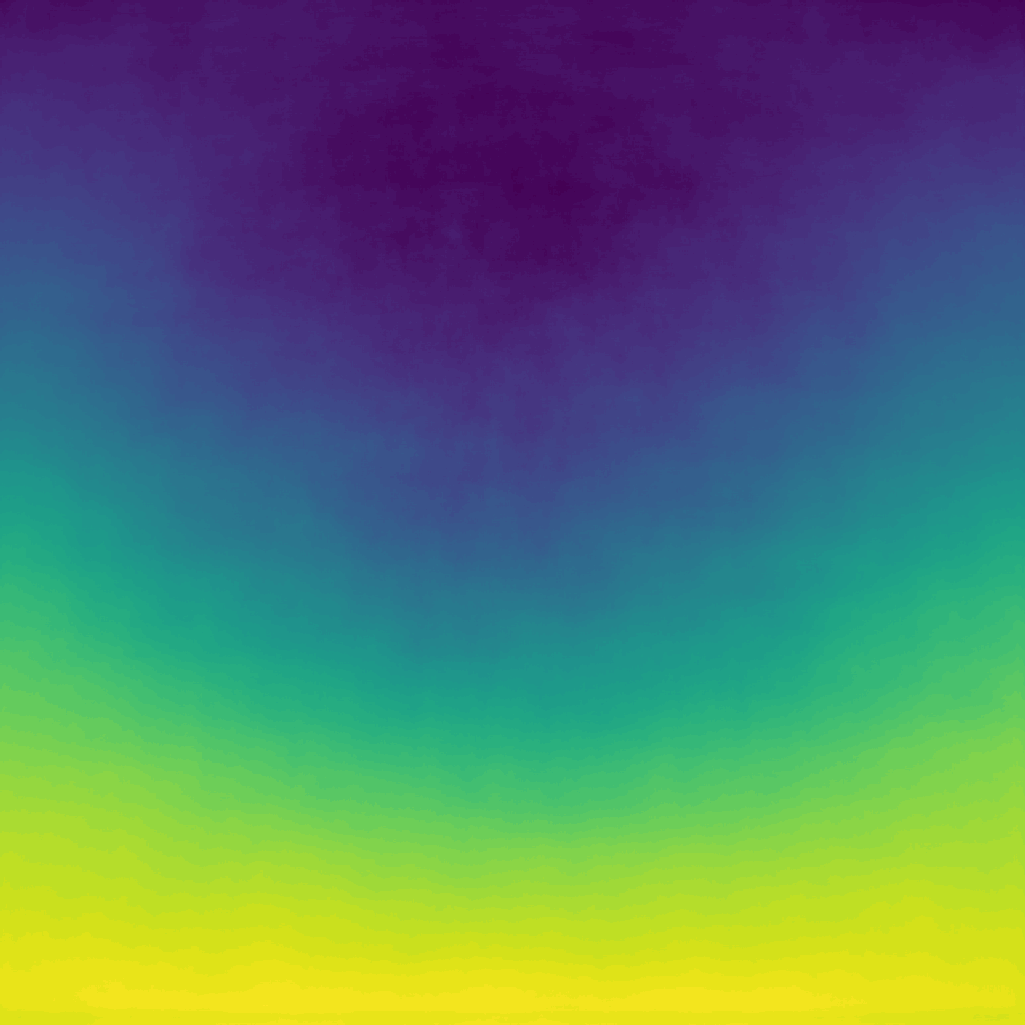}} &
    \raisebox{-.5\height}{\includegraphics[width=0.11\linewidth]{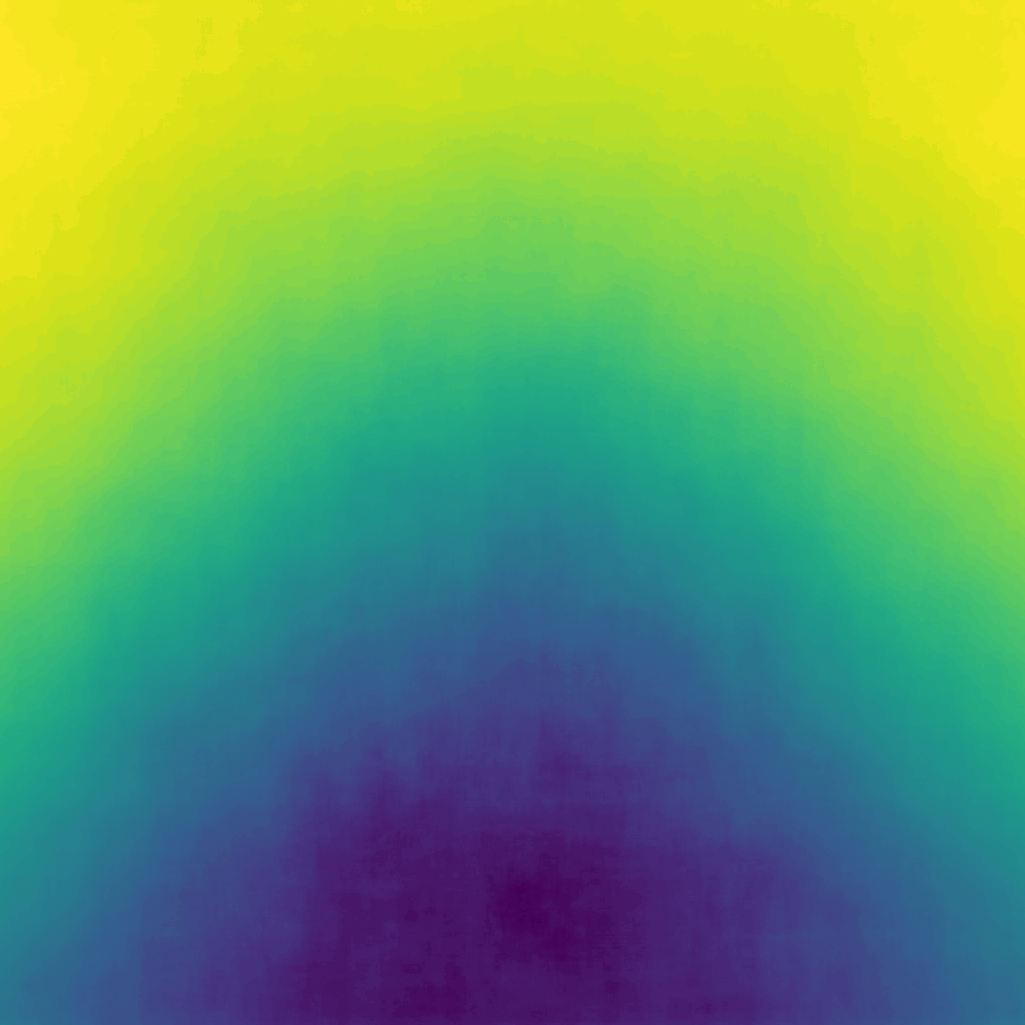}} &
    \raisebox{-.5\height}{\includegraphics[width=0.11\linewidth]{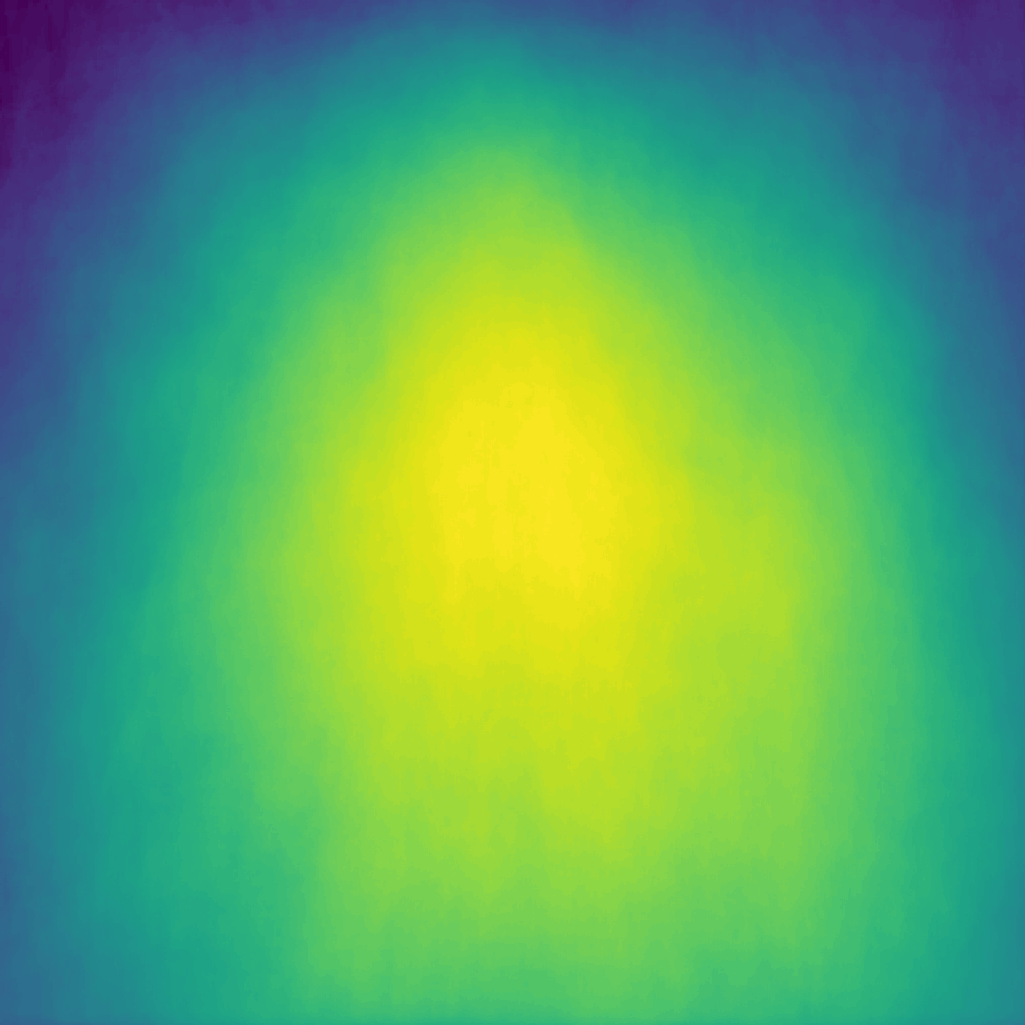}} &
    \raisebox{-.5\height}{\includegraphics[width=0.11\linewidth]{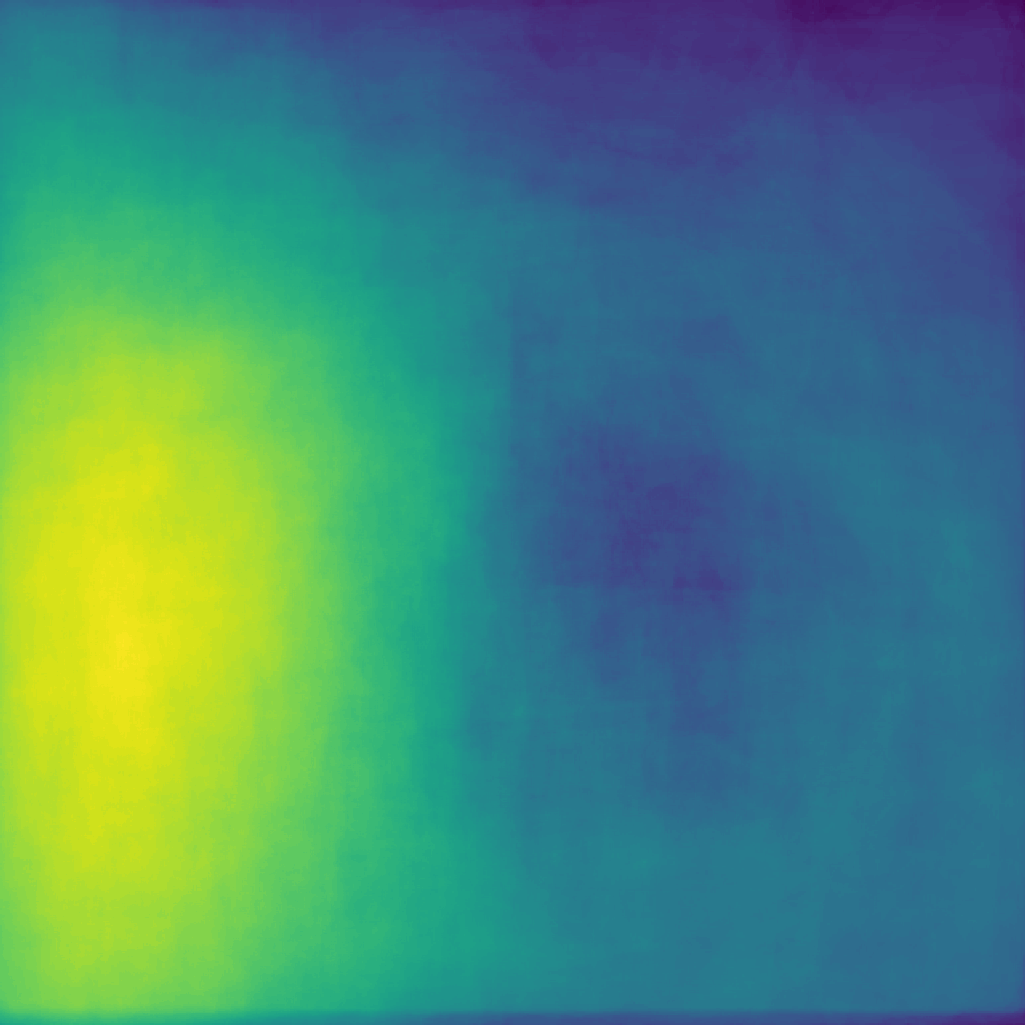}} &
    \raisebox{-.5\height}{\includegraphics[width=0.11\linewidth]{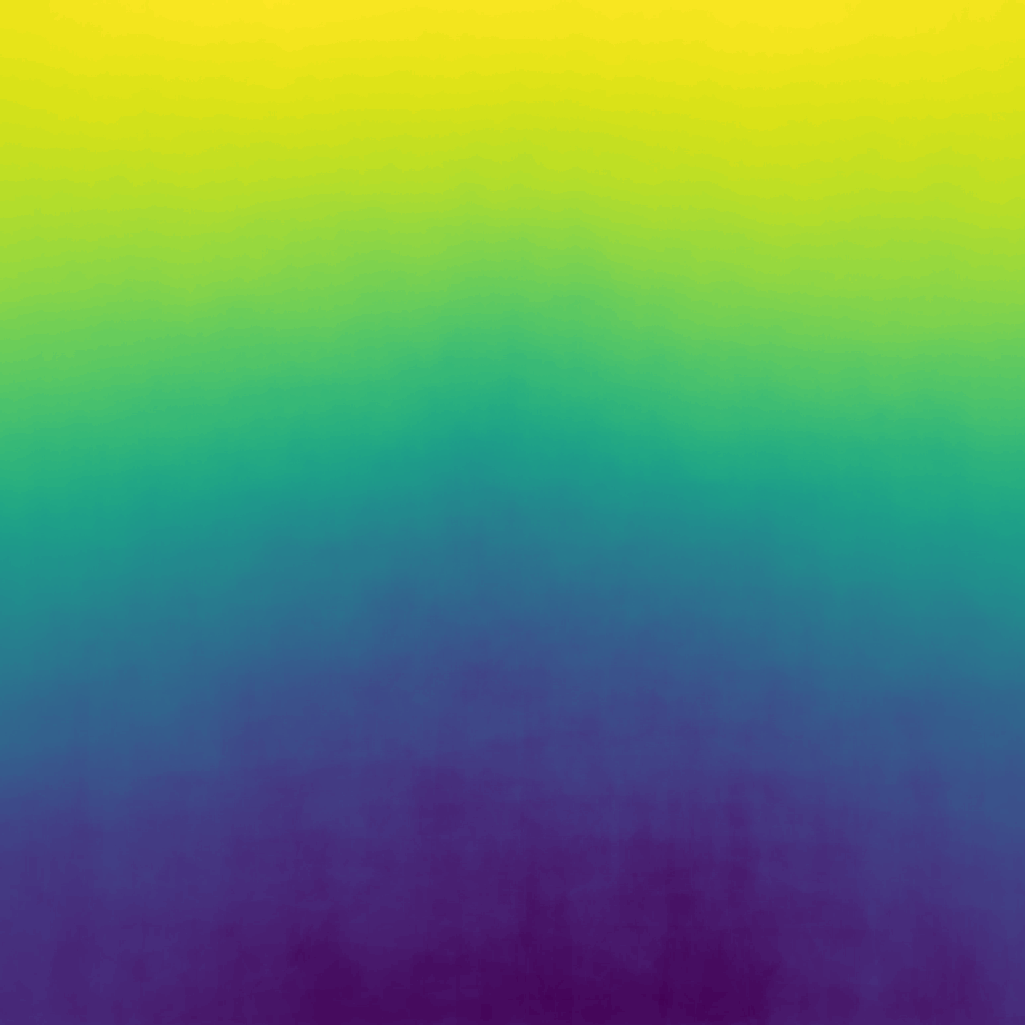}} &
    \raisebox{-.5\height}{\includegraphics[width=0.11\linewidth]{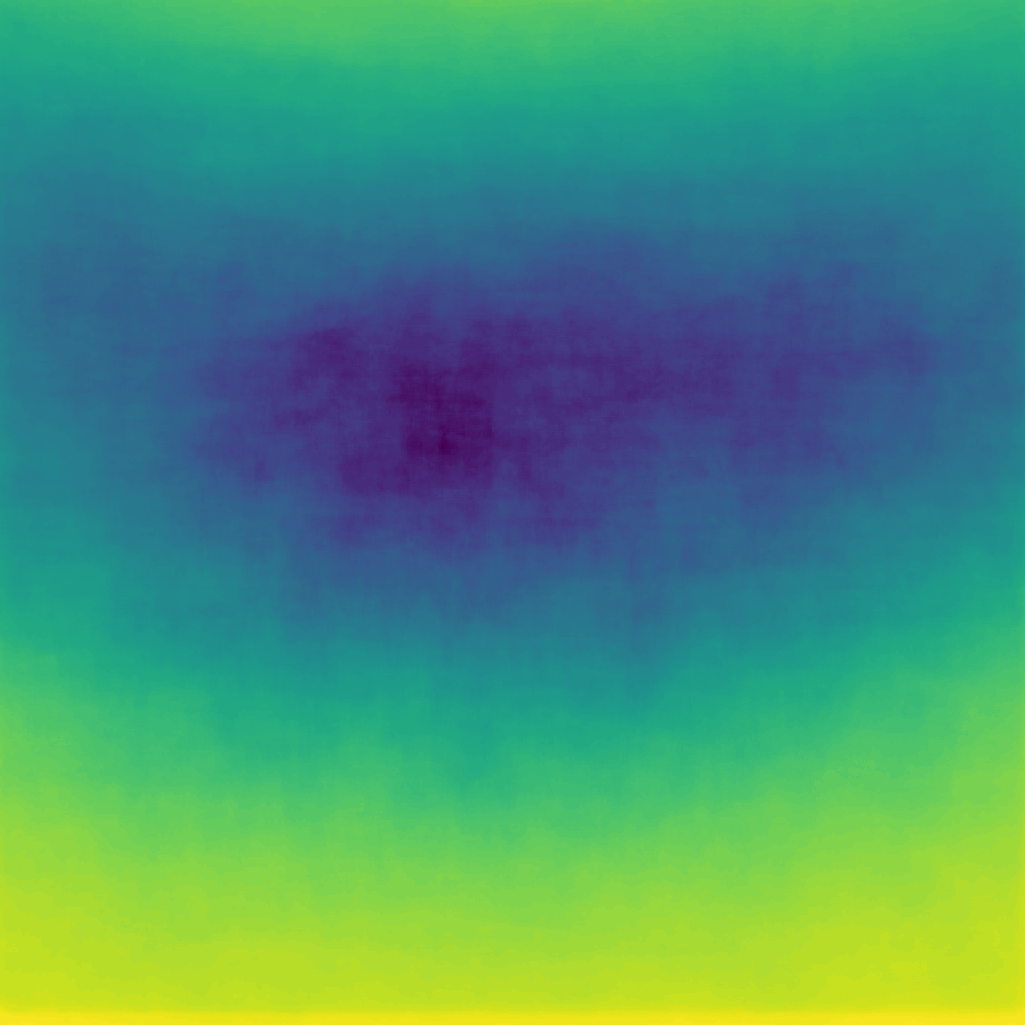}} &
    \raisebox{-.5\height}{\includegraphics[width=0.11\linewidth]{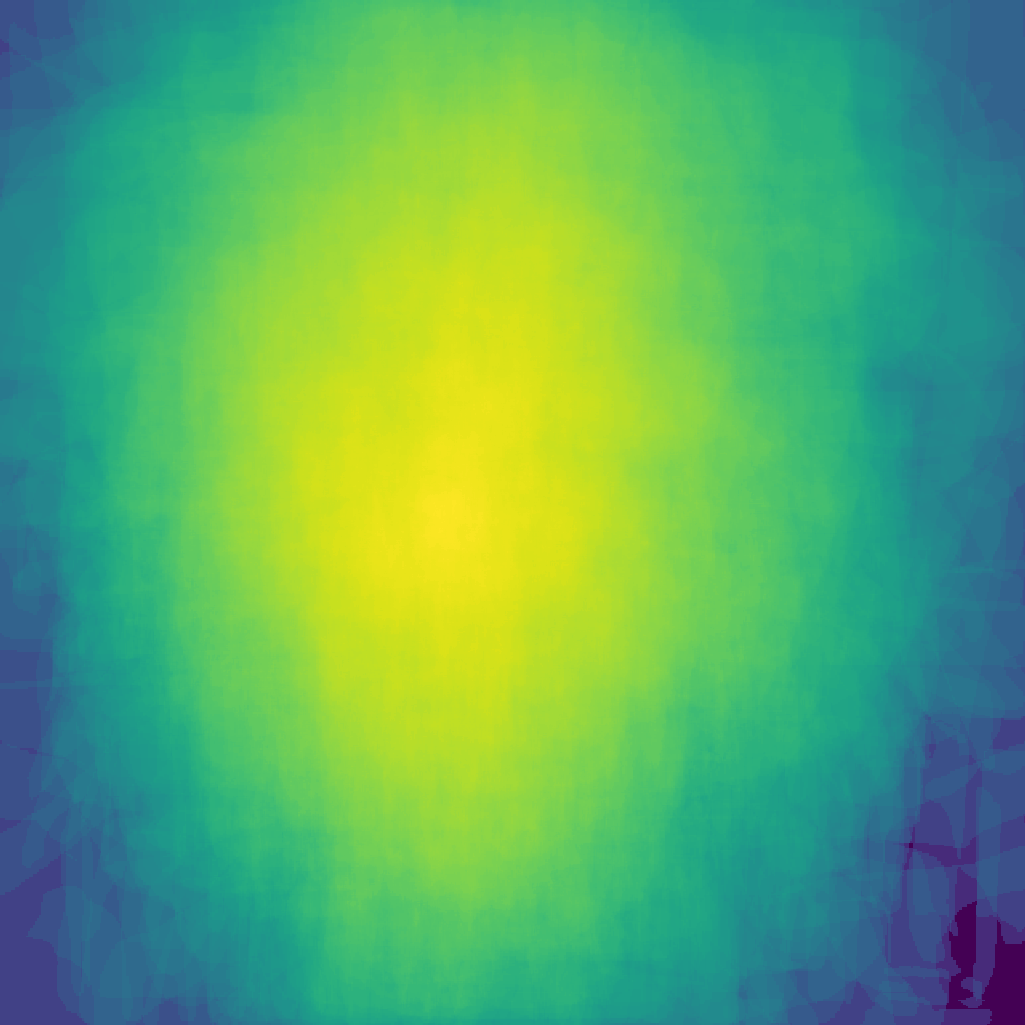}} \\
    \midrule
    & \small Mask Slot 84 & \small Mask Slot 67 & \small Mask Slot 23 & \small Mask Slot 101 & \small Mask Slot 127 & \small Mask Slot 28 & \small Mask Slot 105 & \small Mask Slot 122 \\
    \begin{tabular}{c}
    Medium \\ Firings \\ (sorted)
    \end{tabular} &
    \raisebox{-.5\height}{\includegraphics[width=0.11\linewidth]{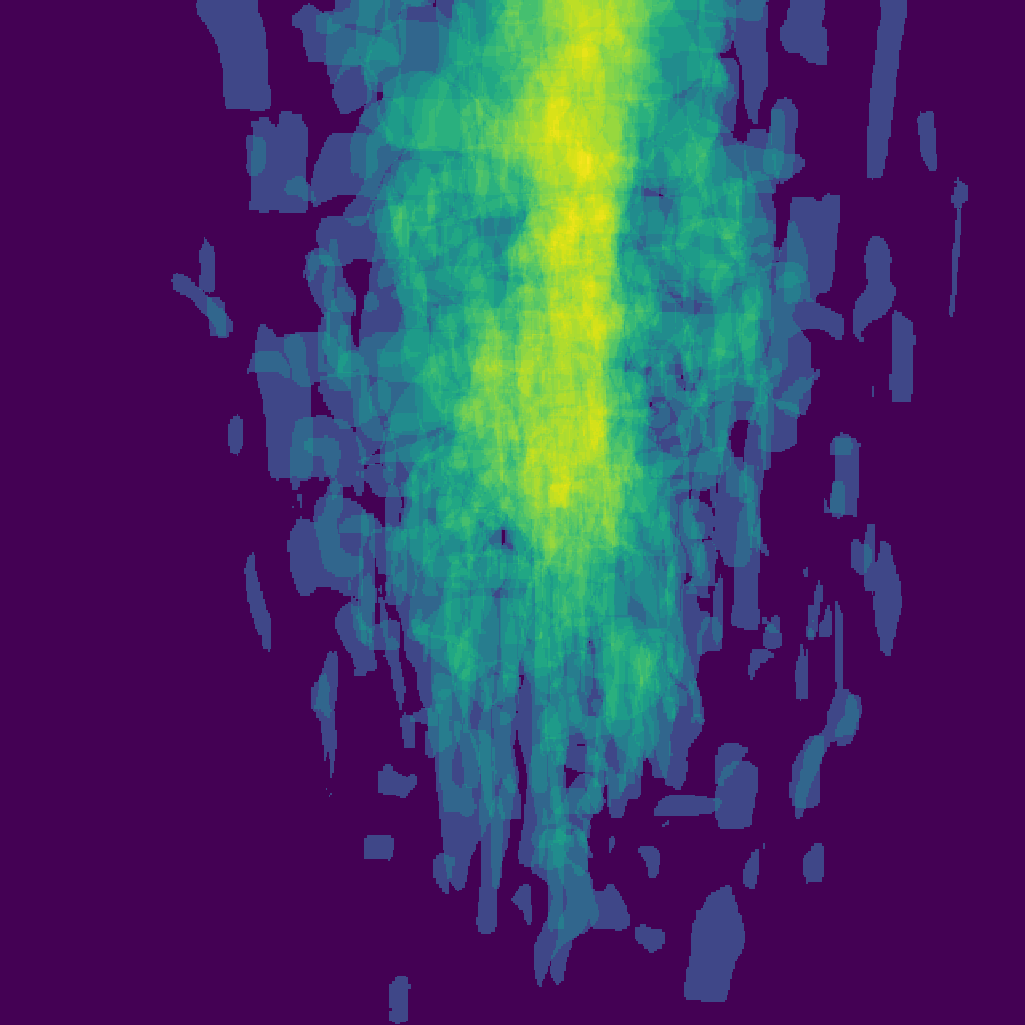}} &
    \raisebox{-.5\height}{\includegraphics[width=0.11\linewidth]{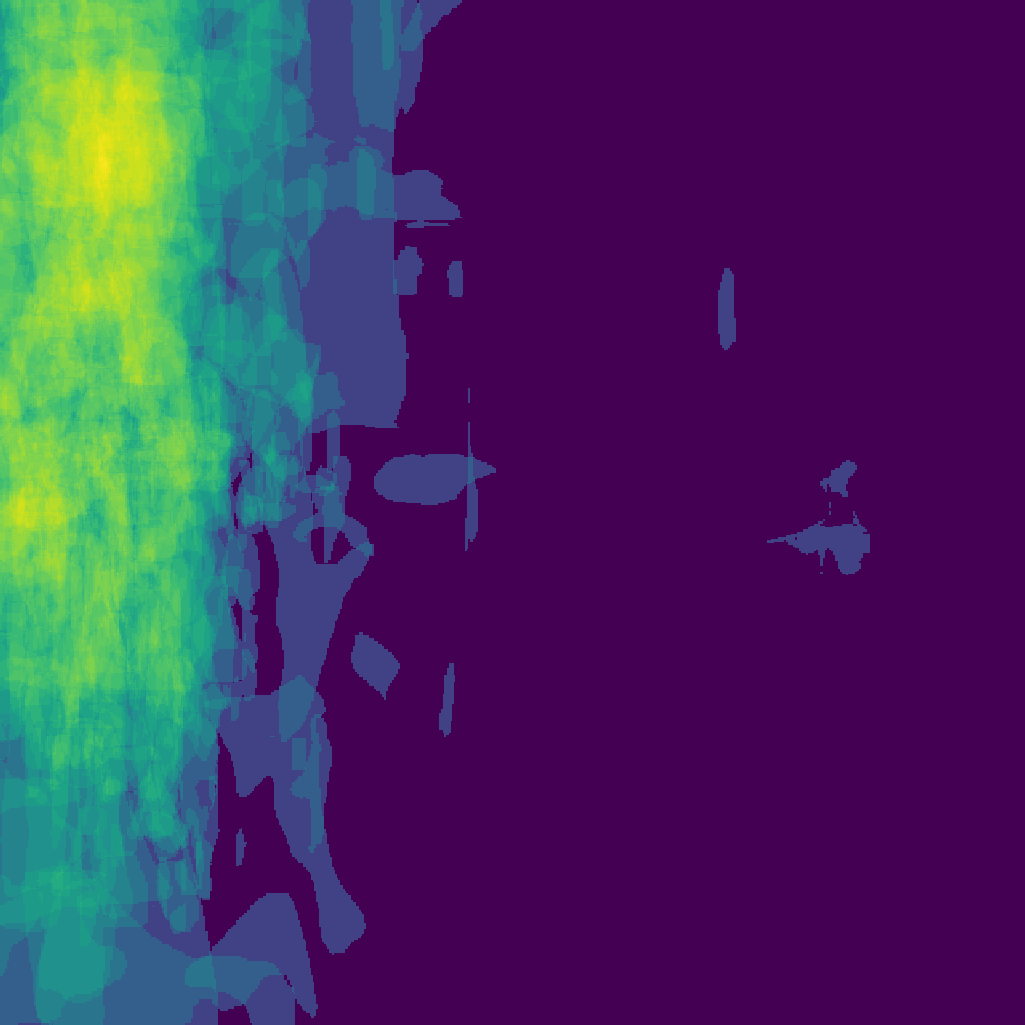}} &
    \raisebox{-.5\height}{\includegraphics[width=0.11\linewidth]{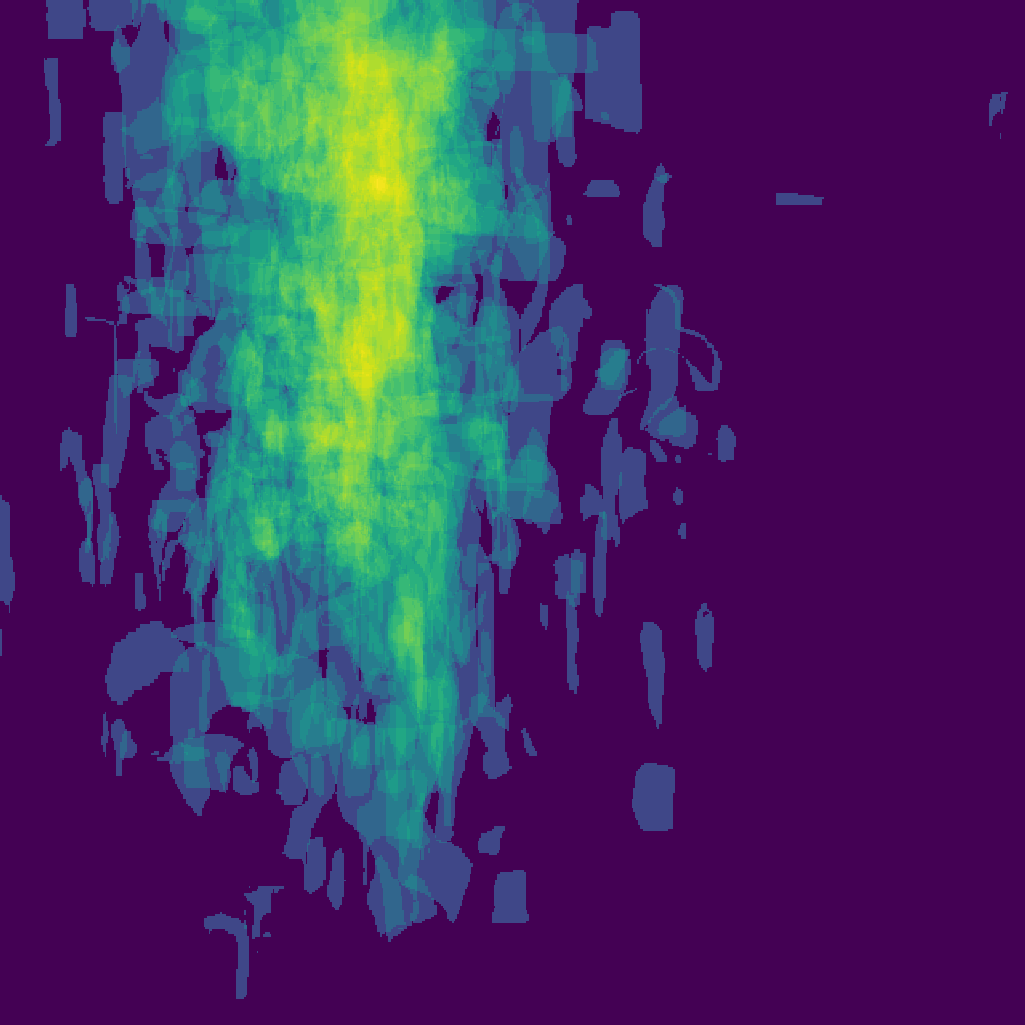}} &
    \raisebox{-.5\height}{\includegraphics[width=0.11\linewidth]{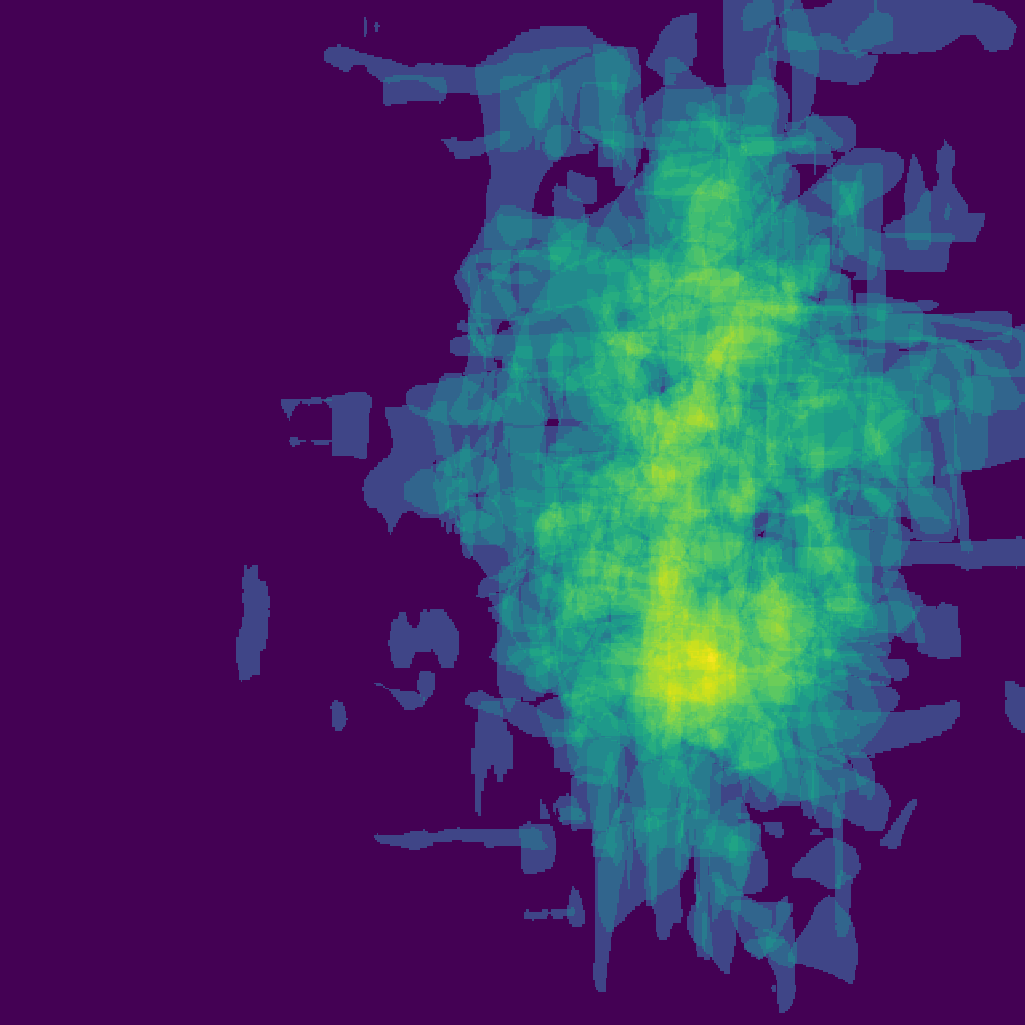}} &
    \raisebox{-.5\height}{\includegraphics[width=0.11\linewidth]{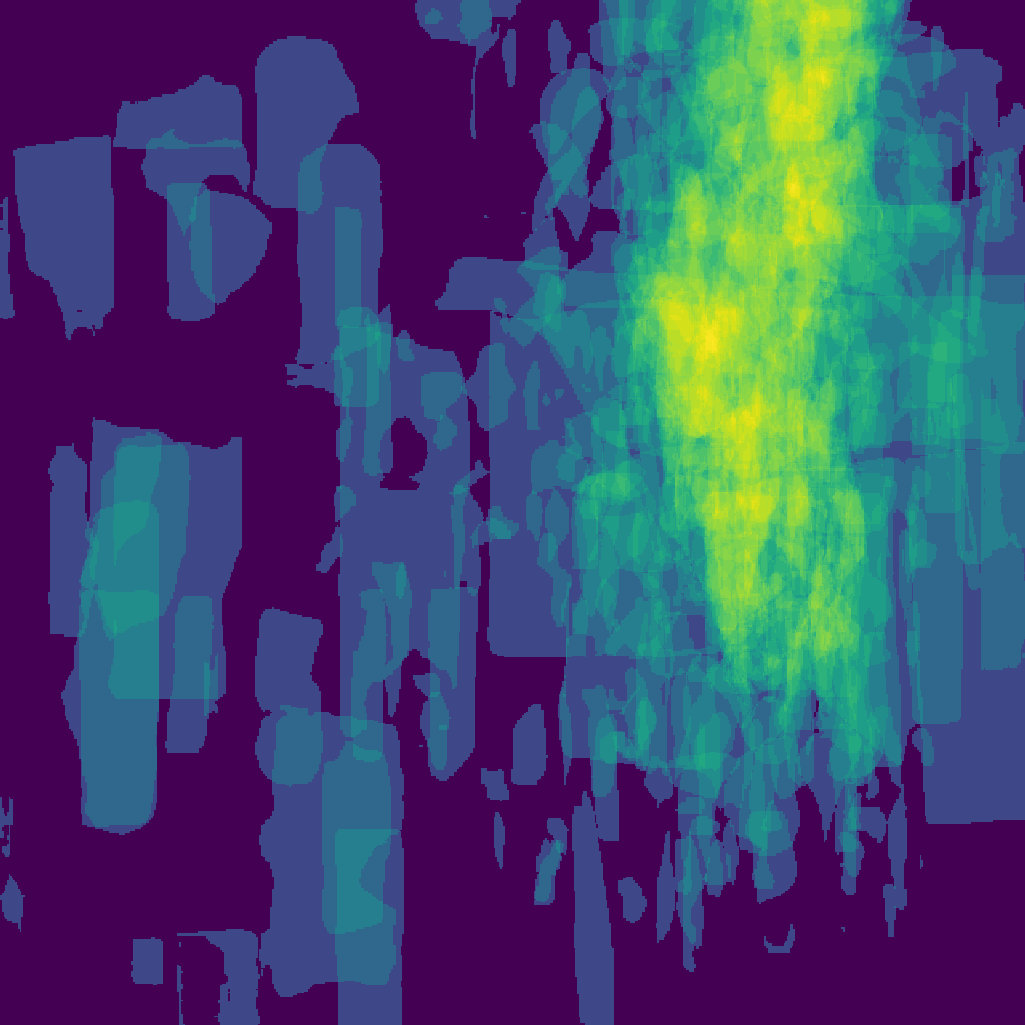}} &
    \raisebox{-.5\height}{\includegraphics[width=0.11\linewidth]{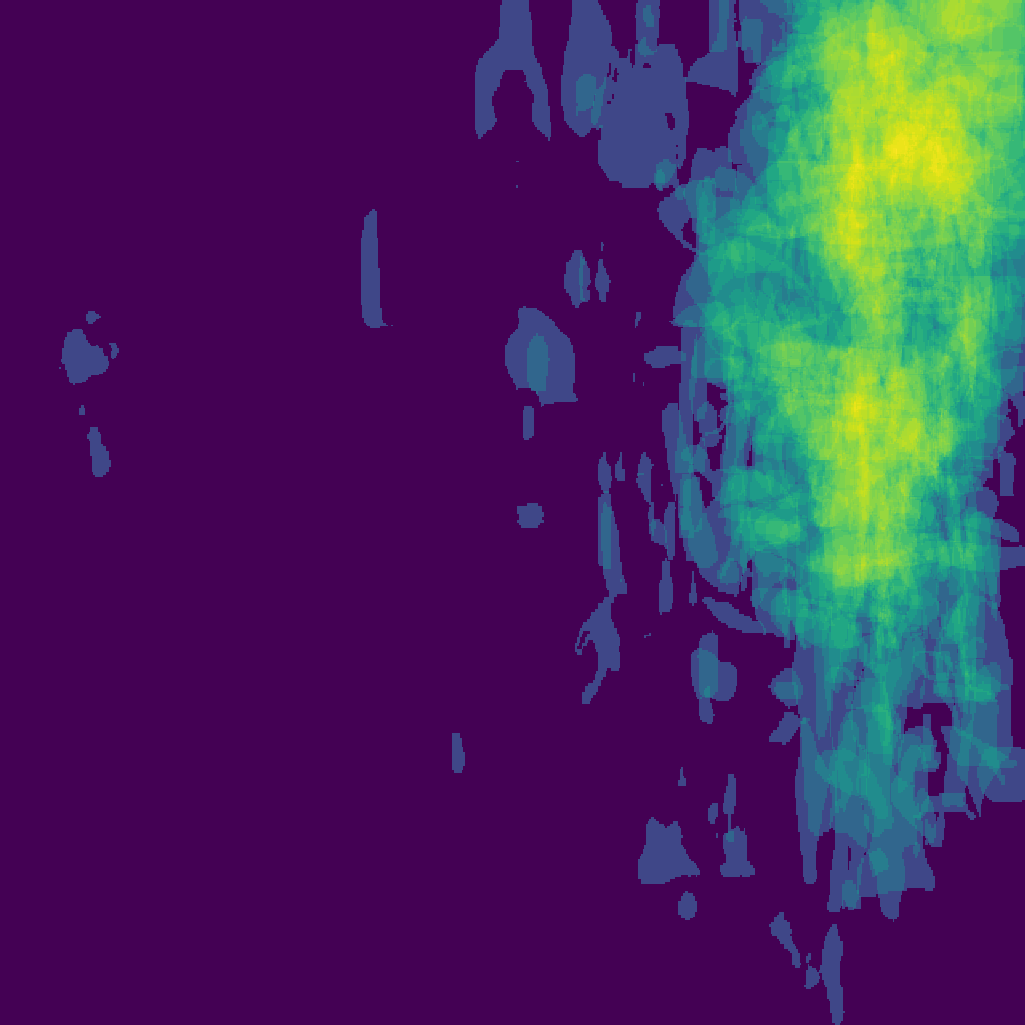}} &
    \raisebox{-.5\height}{\includegraphics[width=0.11\linewidth]{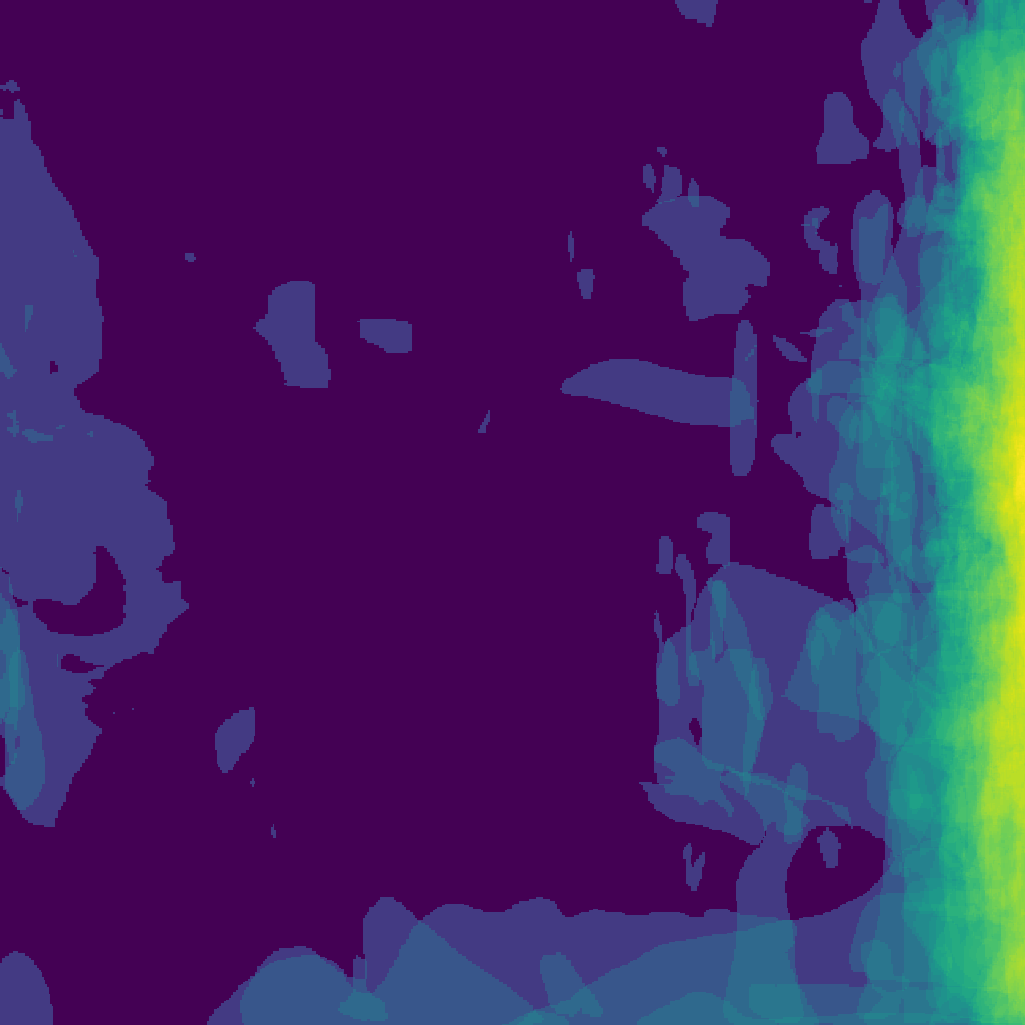}} &
    \raisebox{-.5\height}{\includegraphics[width=0.11\linewidth]{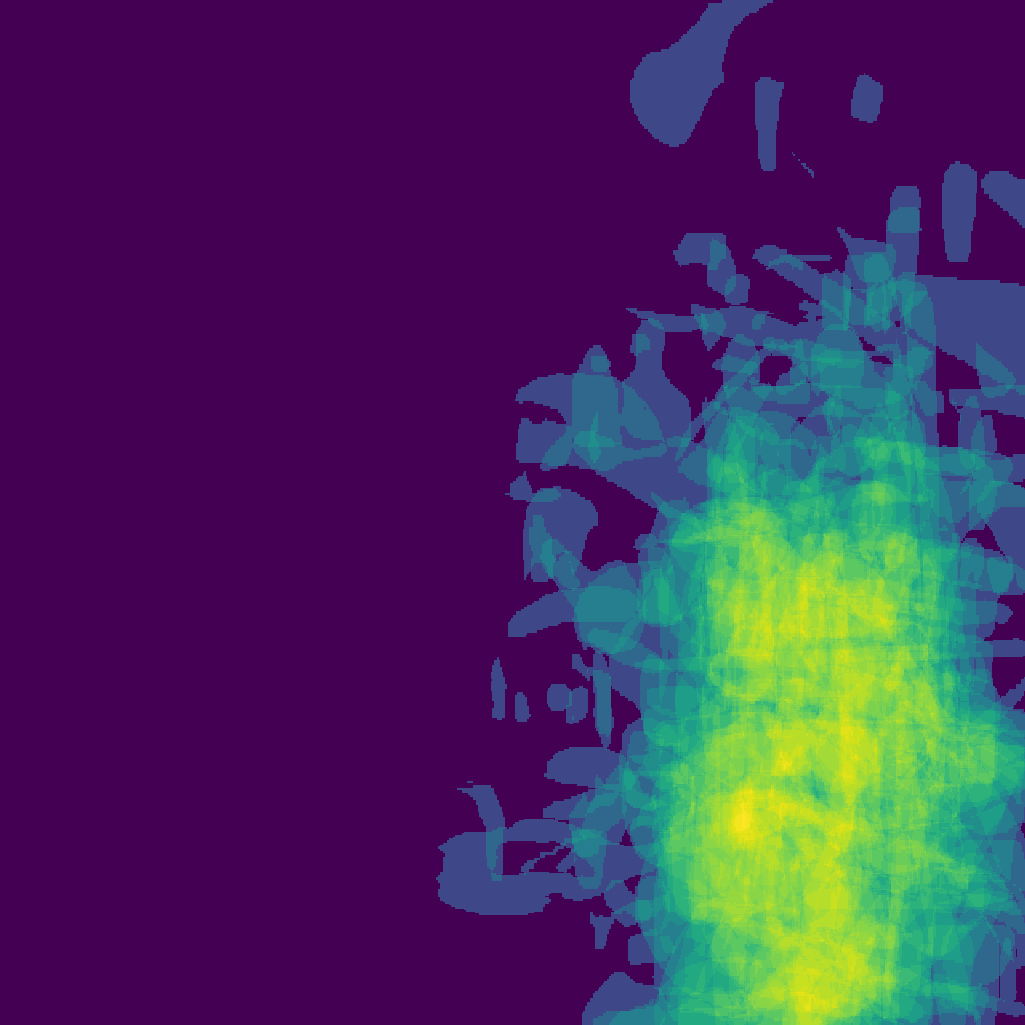}} \\
    \midrule
    & \small Mask Slot 25 & \small Mask Slot 66 & \small Mask Slot 98 & \small Mask Slot 110 & \small Mask Slot 63 & \small Mask Slot 95 & \small Mask Slot 40 & \small Mask Slot 79 \\
    \begin{tabular}{c}
    Few \\ Firings \\ (sorted)
    \end{tabular} &
    \raisebox{-.5\height}{\includegraphics[width=0.11\linewidth]{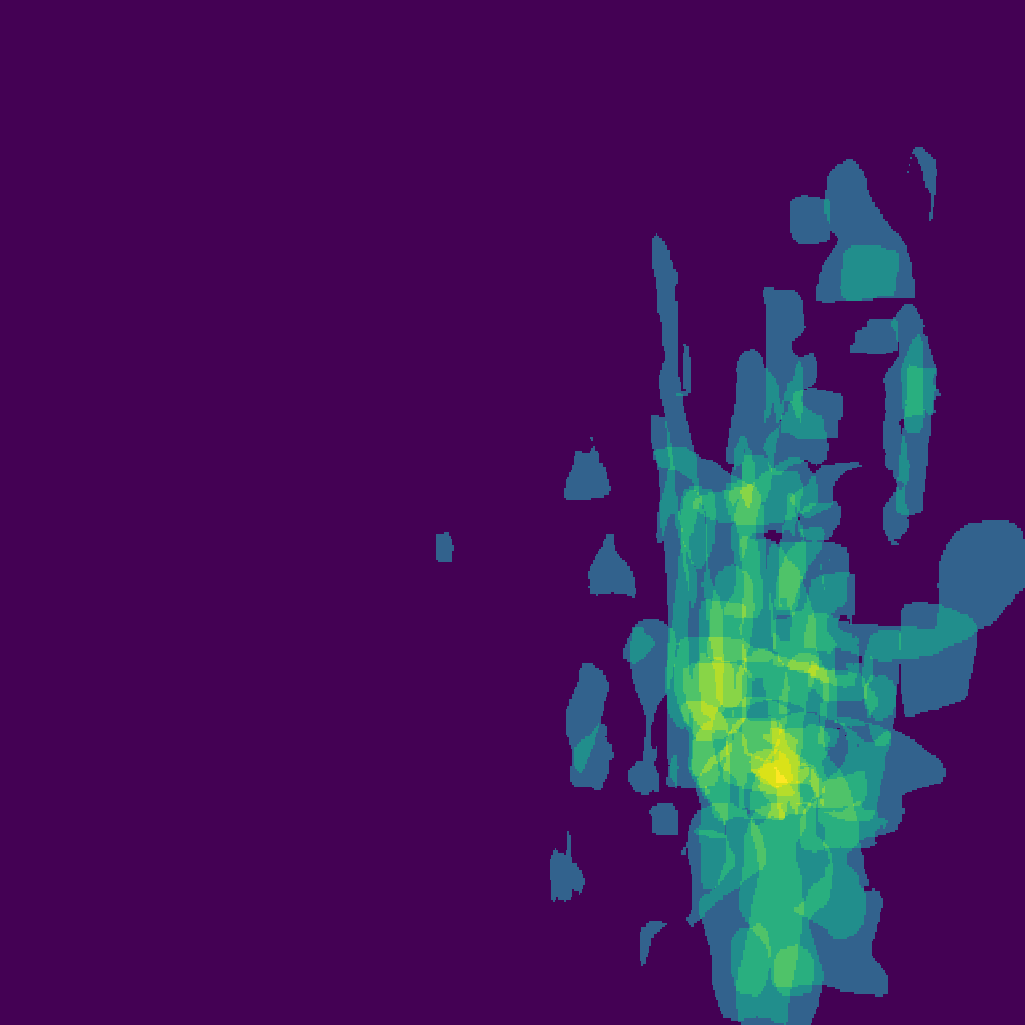}} &
    \raisebox{-.5\height}{\includegraphics[width=0.11\linewidth]{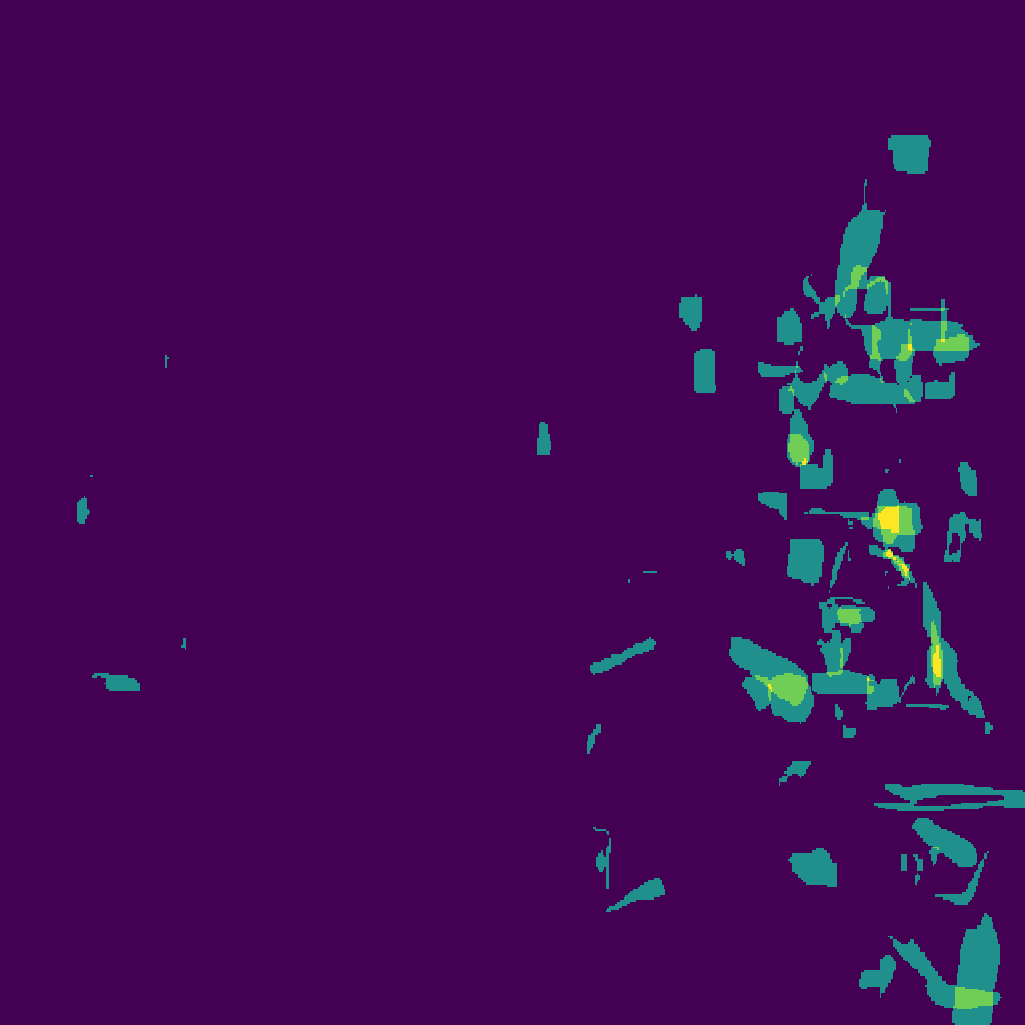}} &
    \raisebox{-.5\height}{\includegraphics[width=0.11\linewidth]{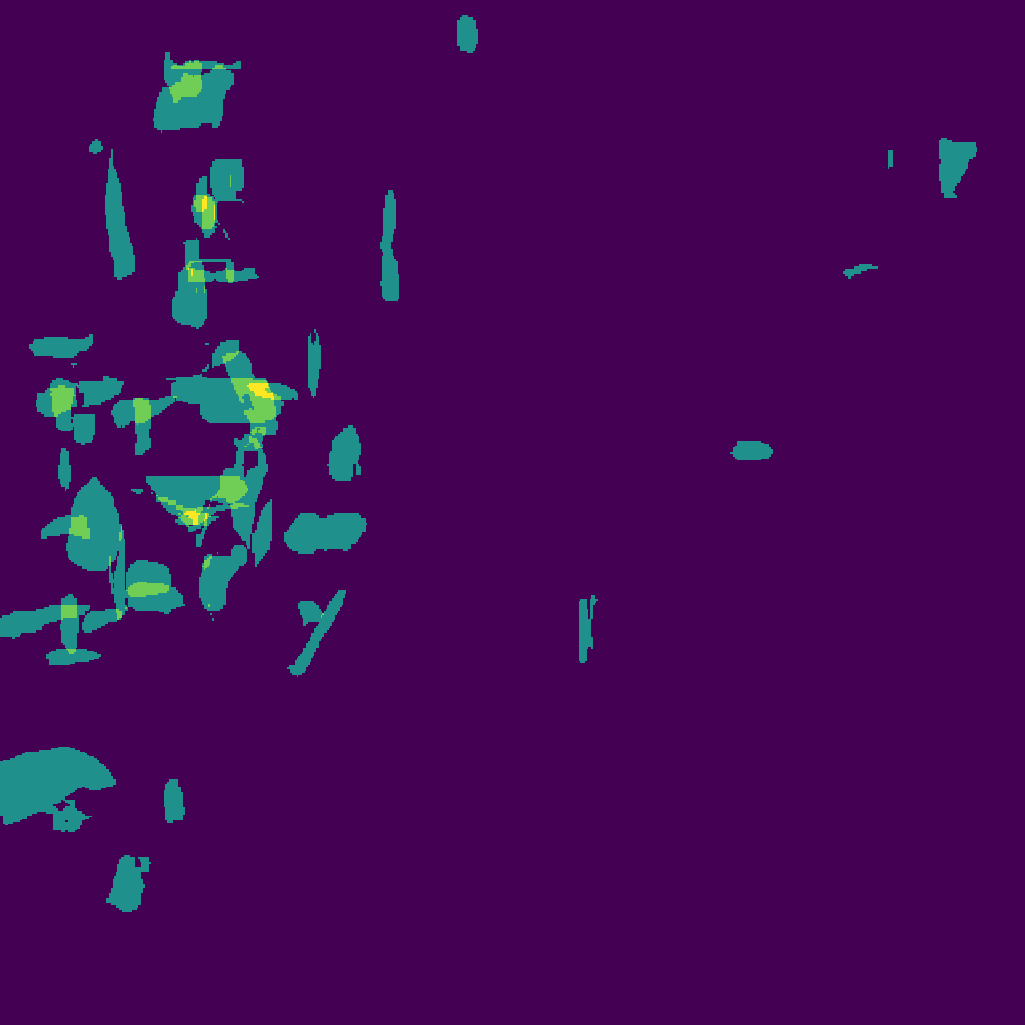}} &
    \raisebox{-.5\height}{\includegraphics[width=0.11\linewidth]{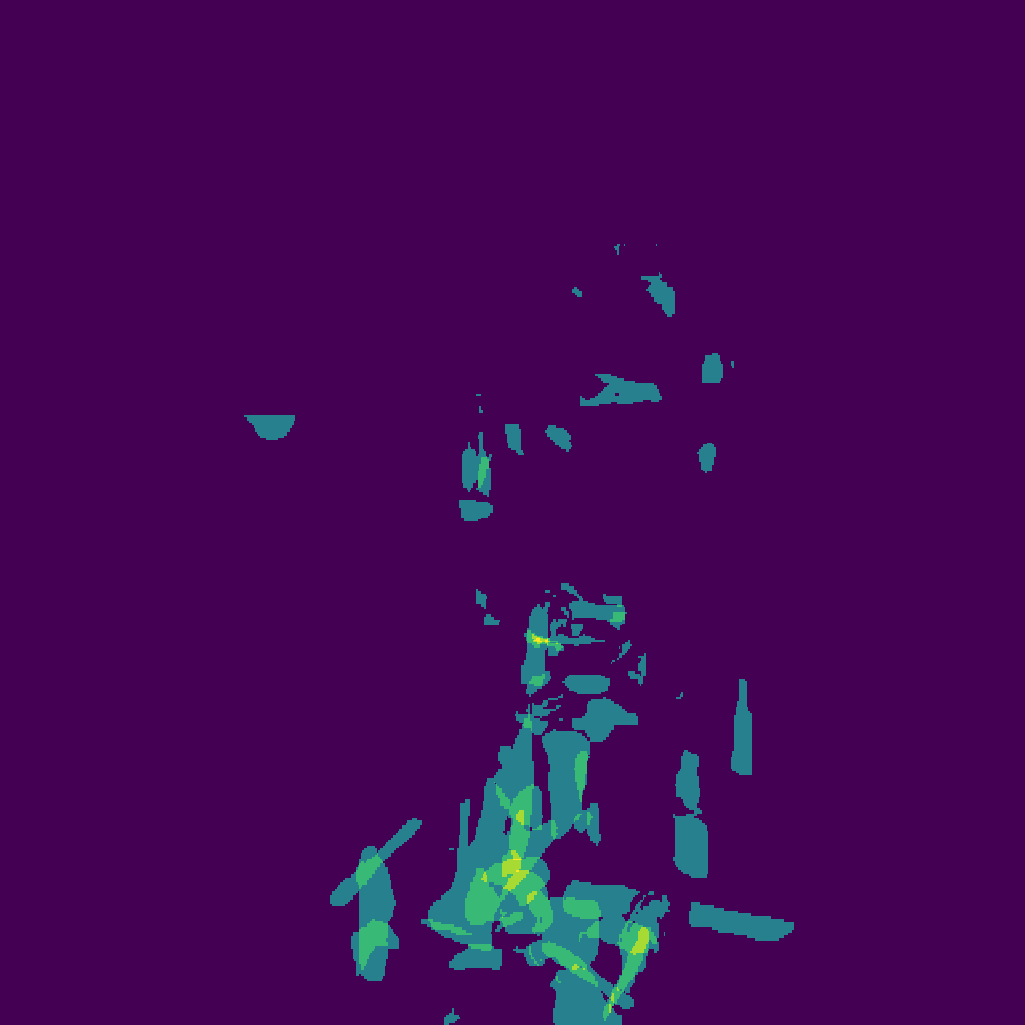}} &
    \raisebox{-.5\height}{\includegraphics[width=0.11\linewidth]{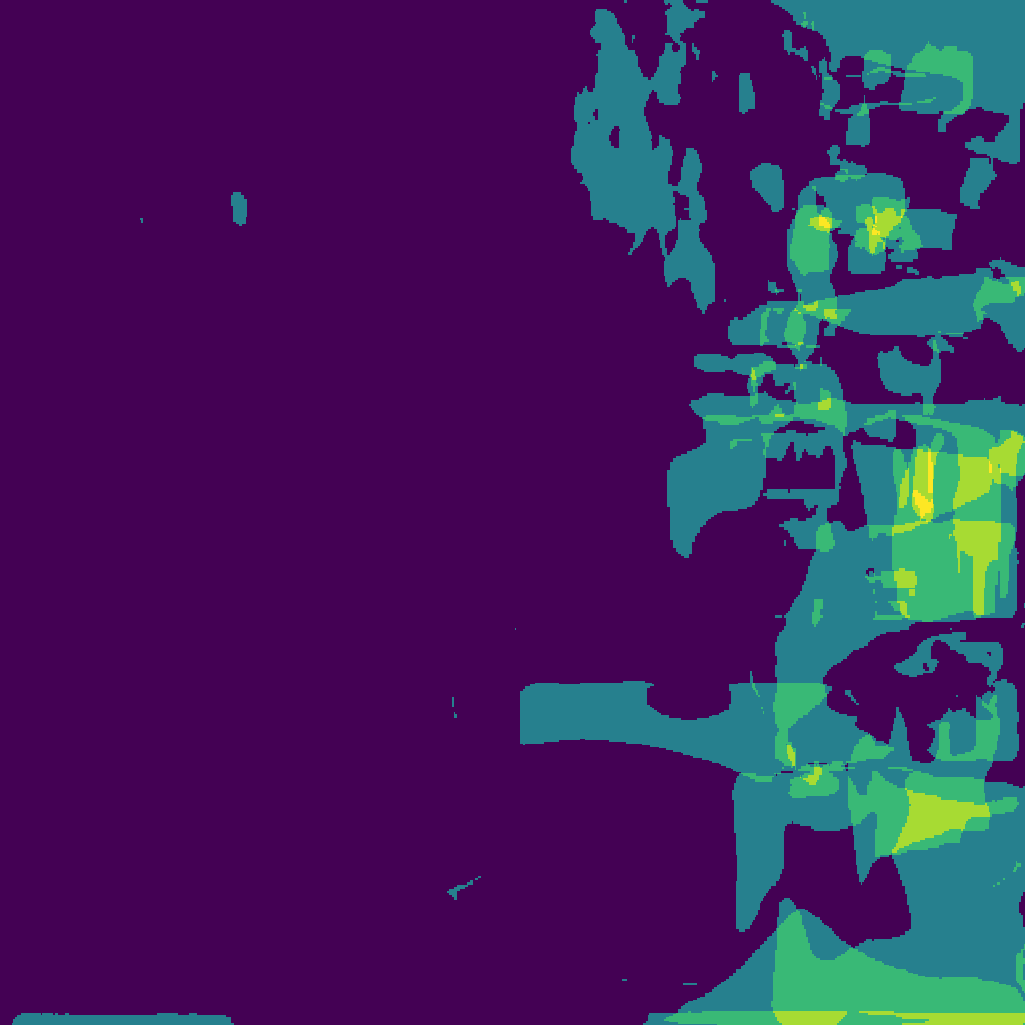}} &
    \raisebox{-.5\height}{\includegraphics[width=0.11\linewidth]{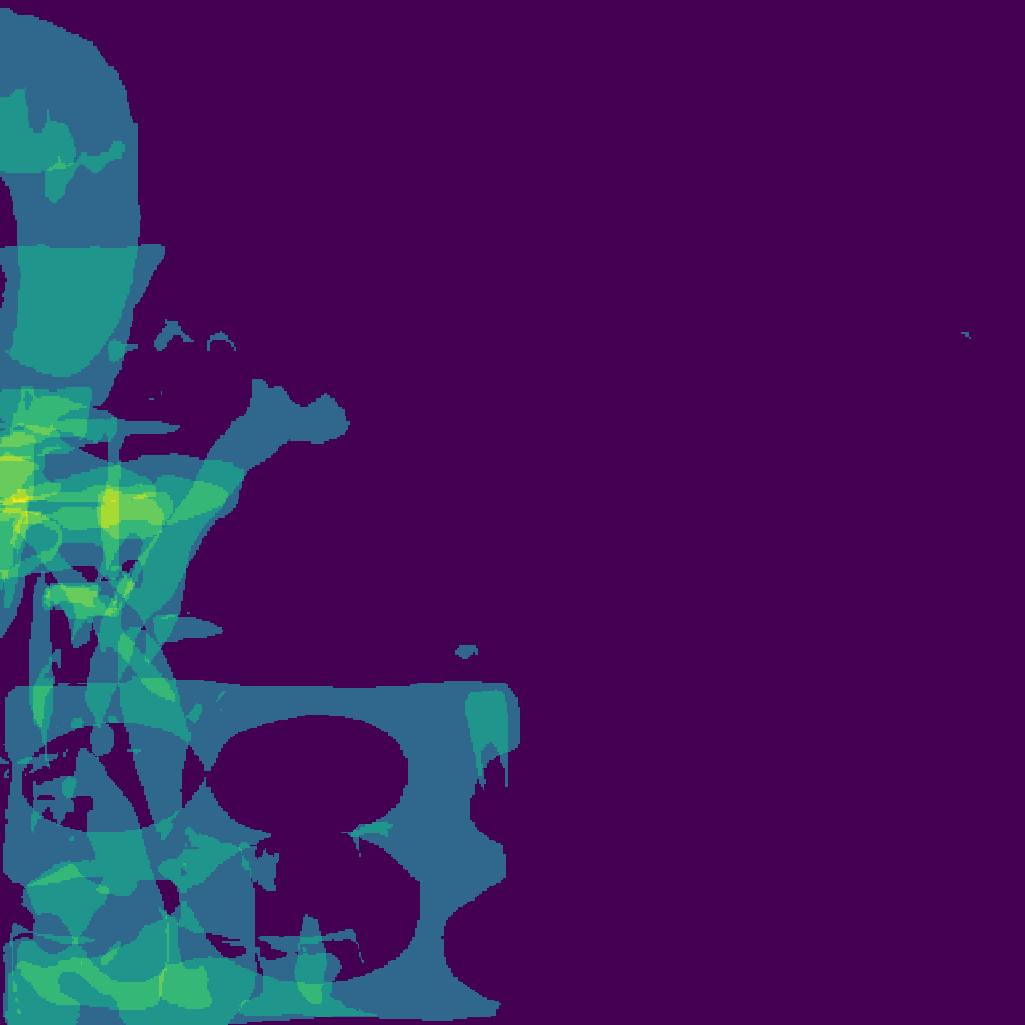}} &
    \raisebox{-.5\height}{\includegraphics[width=0.11\linewidth]{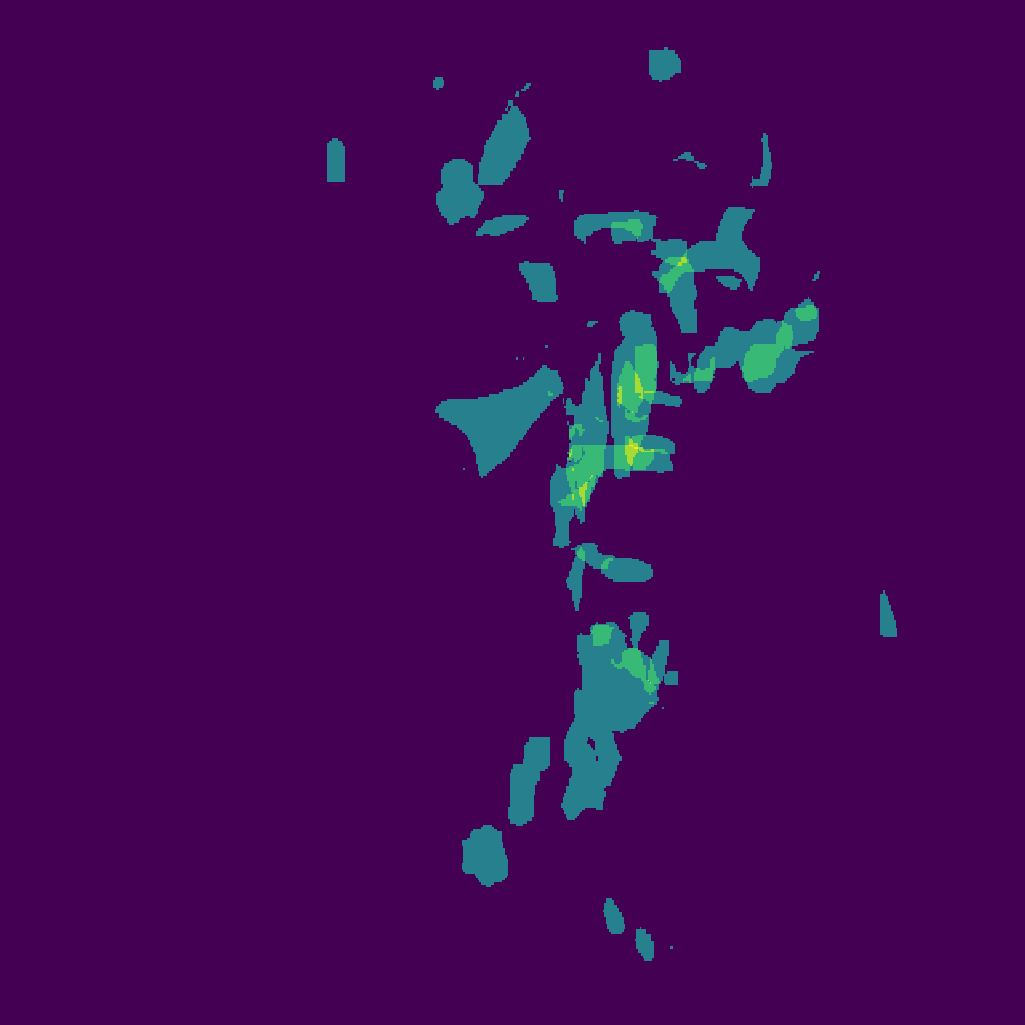}} &
    \raisebox{-.5\height}{\includegraphics[width=0.11\linewidth]{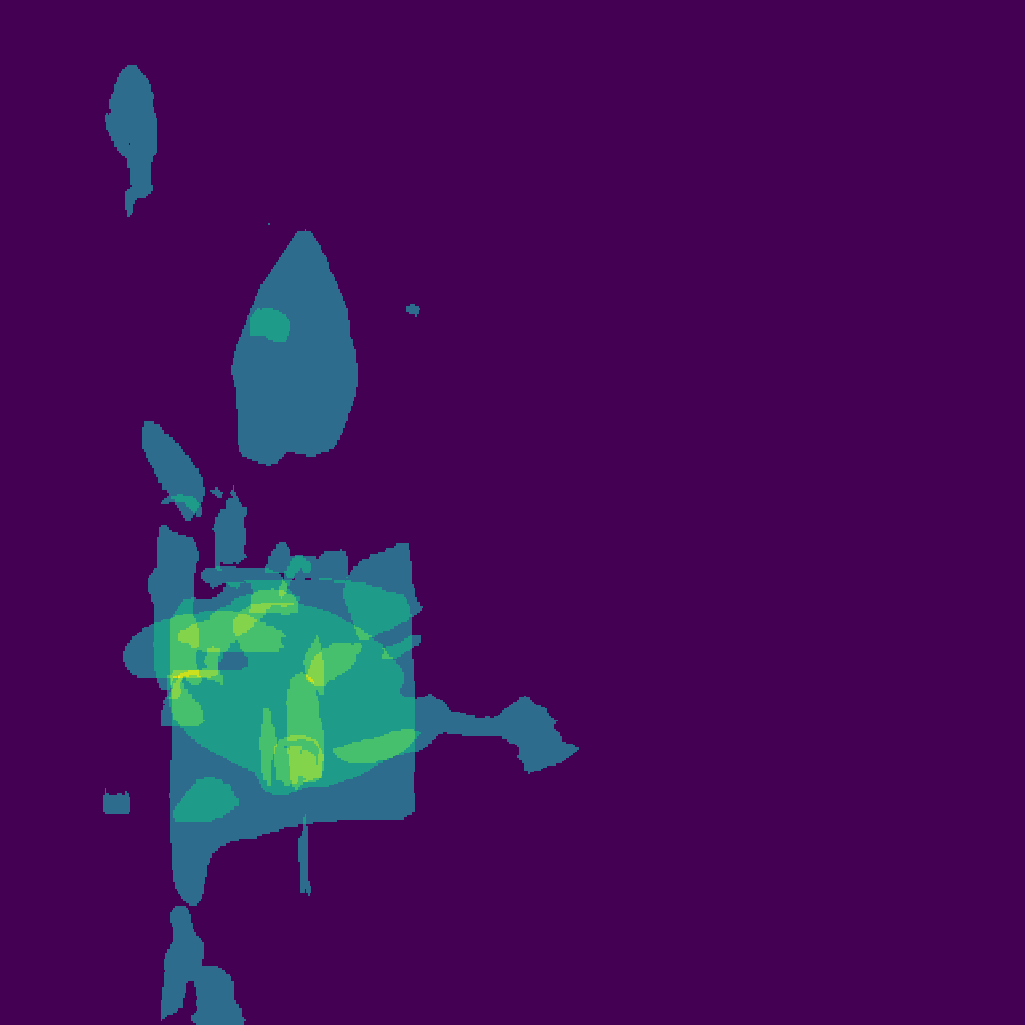}} \\
    \midrule
    \end{tabular}
    \vspace{3pt}
    \caption{The average masks that each mask slot predicts, normalized by image shape. Mask slots are categorized by their total number of firings and sorted from most firings to few firings. We observe spatial clustered patterns, meaning that the mask slots specialize on certain regions of an input image. For example, the most firing mask slot 71, focusing on the center of an image, predicts almost all 80 `thing' classes but ignores `stuff' classes (\figref{fig:distribution}). The top three categories are tennis rackets, cats, and dogs. The second firing mask slot 106 segments 14 classes of masks on the bottom of an image, such as road, floor, or dining-tables. The third firing mask slot 125 concentrates 99.9\% on walls or trees that are usually on the top of an image. The fourth firing mask slot 69 focuses entirely on the person class and predicts 2663 people in the 5000 validation images.
    }
    \label{fig:pixel}
\end{figure*}

\subsection{Panoptic Segmentation Results}
Similar to the case study in \figref{fig:teaser}, we provide more panoptic segmentation results of our MaX-DeepLab-L and compare them to the state-of-the-art {\it box-free} method, Axial-DeepLab~\cite{wang2020axial}, the state-of-the-art {\it box-based} method, DetectoRS~\cite{qiao2020detectors}, and the first Detection Transformer, DETR~\cite{carion2020end} in \figref{fig:comparison} and \figref{fig:failures}. MaX-DeepLab demonstrates robustness to the challenging cases of similar object bounding boxes and nearby objects with close centers, while other methods make systematic mistakes because of their individual surrogate sub-task design. MaX-DeepLab also shows exceptional mask quality, and performs well in the cases of many small objects. Similar to DETR~\cite{carion2020end}, MaX-DeepLab fails typically when there are too many object masks.

\subsection{Runtime}
In \tabref{tab:runtime}, we report the end-to-end runtime (i.e., inference time from an input image to final panoptic segmentation) of MaX-DeepLab on a V100 GPU. All results are obtained by (1) a single-scale input without flipping, and (2) built-in TensorFlow library without extra inference optimization.
In the fast regime, MaX-DeepLab-S takes 67 ms with a typical 641$\times$641 input. This runtime includes 5 ms of postprocessing and 15 ms of batch normalization that can be easily optimized. This fast MaX-DeepLab-S does not only outperform DETR-R101~\cite{carion2020end}, but is also around 2x faster.
In the slow regime, the standard MaX-DeepLab-S takes 131 ms with a 1025$\times$1025 input, similar to Panoptic-DeepLab-X71~\cite{cheng2019panoptic}. This runtime is also similar to our run of the official DETR-R101 which takes 128 ms on a V100, including 63 ms for box detection and 65 ms for the heavy mask decoding.

\subsection{Mask Output Slot Analysis}
In this subsection, we analyze the statistics of all $N=128$ mask prediction slots using MaX-DeepLab-L. In \figref{fig:distribution}, we visualize the joint distribution of mask slot firings and the classes they predict. We observe that the mask slots have imbalanced numbers of predictions and they specialize on `thing' classes and `stuff' classes. Similar to this Mask-Class joint distribution, we visualize the Mask-Pixel joint distribution by extracting an average mask for each mask slot, as shown in \figref{fig:pixel}. Specifically, we resize all COCO~\cite{lin2014microsoft} validation set panoptic segmentation results to a unit square and take an average of masks that are predicted by each mask slot. We split all mask slots into three categories according to their total firings and visualize mask slots in each category. We observe that besides the class-level specialization, our mask slots also specialize on certain regions of an input image. This observation is similar to DETR~\cite{carion2020end}, but we do not see the pattern that almost all slots have a mode of predicting large image-wide masks.

\subsection{Mask Head Visualization}
\begin{figure*}[t]
    \centering
    \setlength\tabcolsep{2.0pt}
    \begin{tabular}{ccc|ccc}
    \includegraphics[width=0.155\linewidth, height=0.11\linewidth]{figures_appendix/000013_image.png} &
    \includegraphics[width=0.155\linewidth, height=0.11\linewidth]{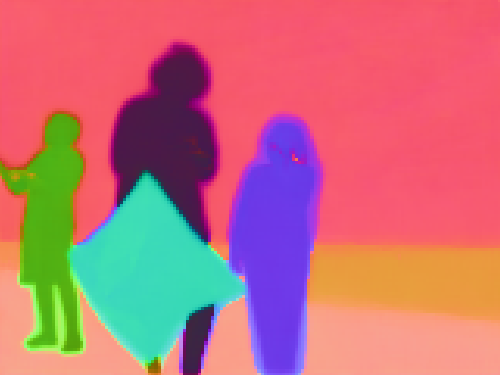} &
    \includegraphics[width=0.155\linewidth, height=0.11\linewidth]{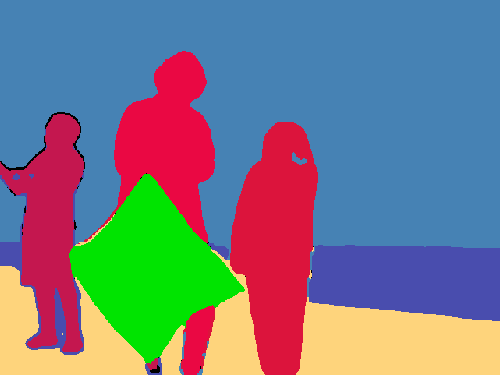}
    \phantom{.} & \phantom{.}
    \includegraphics[width=0.155\linewidth, height=0.11\linewidth]{figures_appendix/000049_image.png} &
    \includegraphics[width=0.155\linewidth, height=0.11\linewidth]{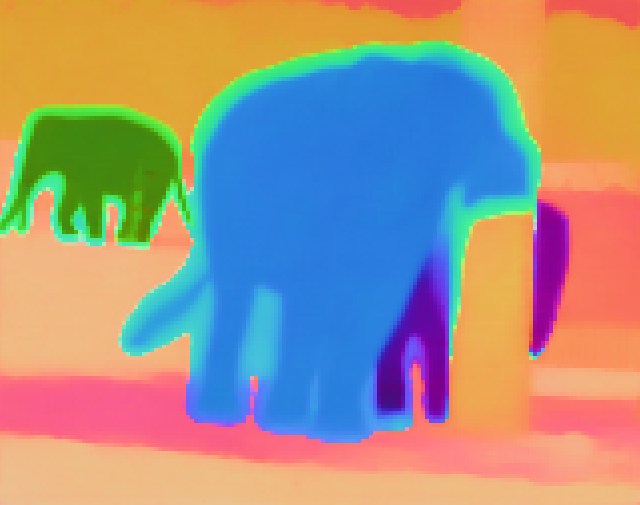} & \includegraphics[width=0.155\linewidth, height=0.11\linewidth]{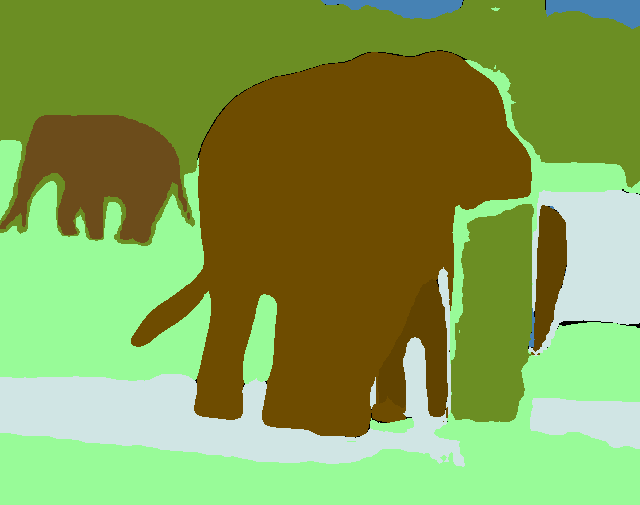} \\
    \includegraphics[width=0.155\linewidth, height=0.11\linewidth]{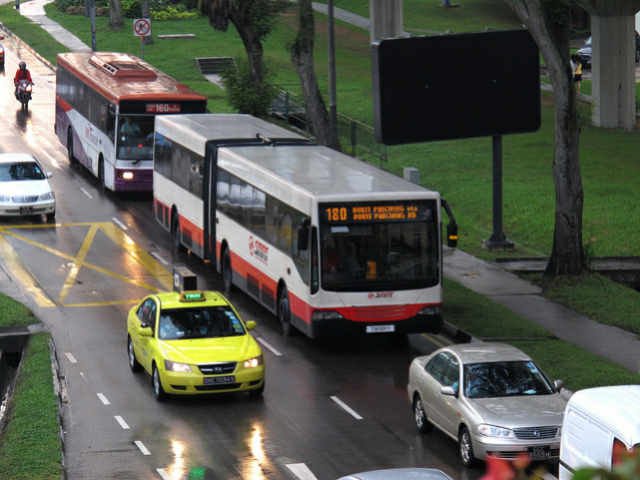} &
    \includegraphics[width=0.155\linewidth, height=0.11\linewidth]{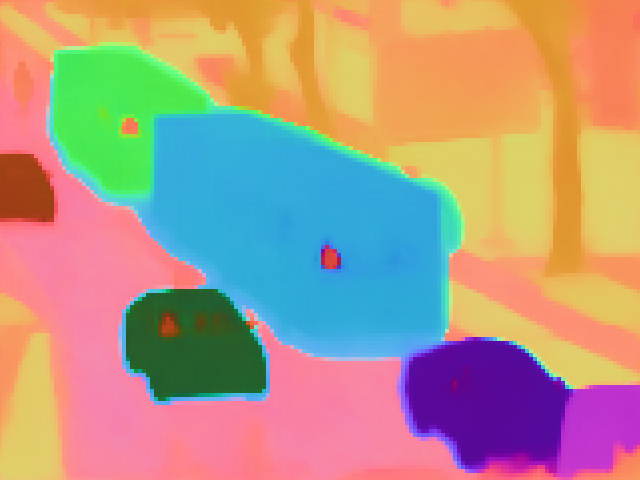} & \includegraphics[width=0.155\linewidth, height=0.11\linewidth]{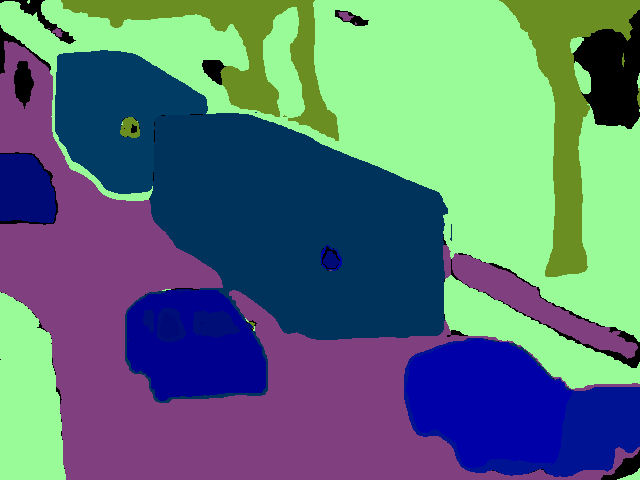}
    \phantom{.} & \phantom{.}
    \includegraphics[width=0.155\linewidth, height=0.11\linewidth]{figures_appendix/000026_image.png} &
    \includegraphics[width=0.155\linewidth, height=0.11\linewidth]{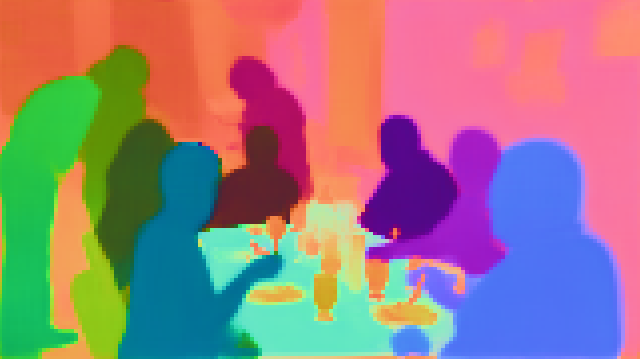} &
    \includegraphics[width=0.155\linewidth, height=0.11\linewidth]{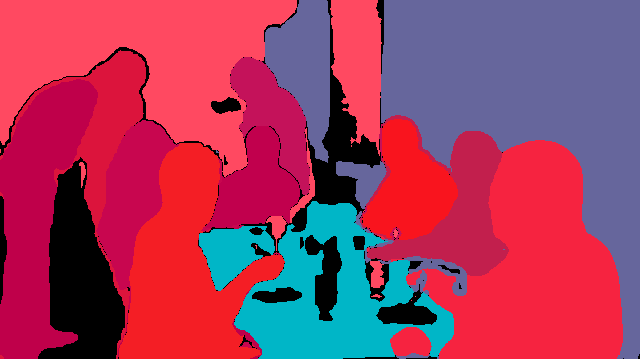} \\
    \small Original Image & \small Decoder Feature $g$ & \small Panoptic Seg. \phantom{.} & \small \phantom{.} Original Image & \small Decoder Feature $g$ & \small Panoptic Seg. \\
    \end{tabular}
    \vspace{3pt}
    \caption{More visualizations of the decoder feature $g$ with $D=3$. Similar to \figref{fig:vis3channels}, we observe a clustering effect of instance colors, \ie, pixels of the same instance have similar colors (features) while pixels of different instances have distinct colors. Note that in this extreme case of $D=3$ (that achieves 37.8\% PQ), there are not enough colors for all masks, which causes missing objects or artifacts at object boundaries, but these artifacts do not present in our normal setting of $D=128$ (that achieves 45.7\%~PQ).}
    \label{fig:maskhead}
\end{figure*}

In \figref{fig:vis3channels}, we visualize how the mask head works by training a MaX-DeepLab with only $D=3$ decoder feature channels (for visualization purpose only). Although this extreme setting degrades the performance from 45.7\% PQ to 37.8\% PQ, it enables us to directly visualize the decoder features as RGB colors. Here in \figref{fig:maskhead} we show more examples using this model, together with the corresponding panoptic sementation results. We see a similar clustering effect of instance colors, which enables our simple mask extraction with just a matrix multiplication (a.k.a. dynamic convolution~\cite{tian2020conditional,wang2020solov2,jia2016dynamic,yang2019condconv}).

\subsection{Transformer Attention Visualization}
\begin{figure*}[!p]
    \centering
    \setlength\tabcolsep{1.12345pt}
    \begin{tabular}{ccccc}
    \midrule
    Original Image & Head 1 & Head 2 & Head 3 & Head 4 \\
    \includegraphics[width=0.19\textwidth]{figures_appendix/486104.jpg} \phantom{.}&\phantom{.} \includegraphics[width=0.19\textwidth]{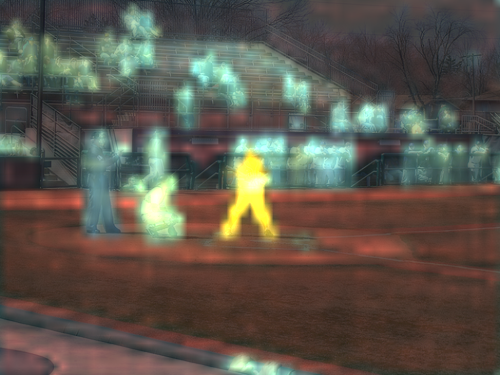} & \includegraphics[width=0.19\textwidth]{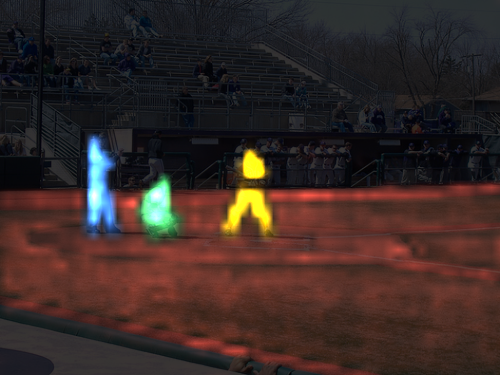} & \includegraphics[width=0.19\textwidth]{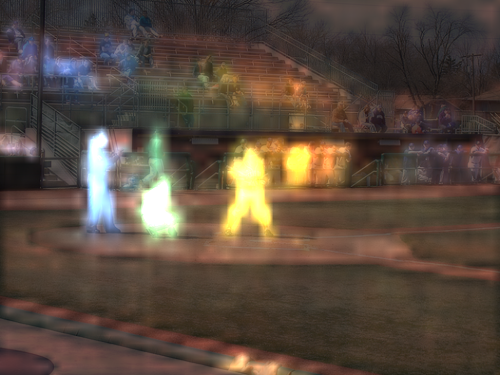} & \includegraphics[width=0.19\textwidth]{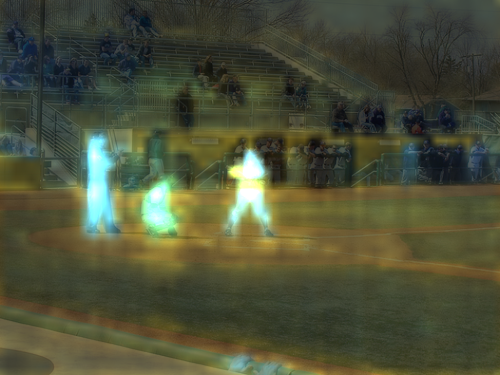} \\
    Panoptic Segmentation \phantom{.} & Head 5 & Head 6 & Head 7 & Head 8 \\
    \includegraphics[width=0.19\textwidth]{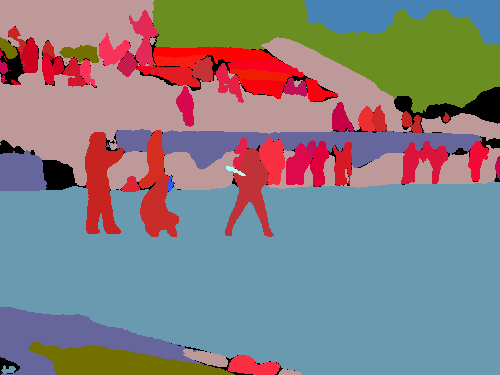} \phantom{.}&\phantom{.} \includegraphics[width=0.19\textwidth]{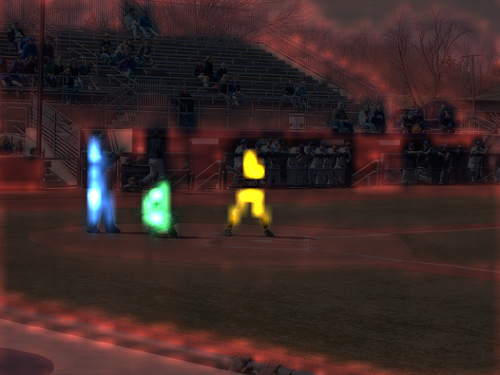} & \includegraphics[width=0.19\textwidth]{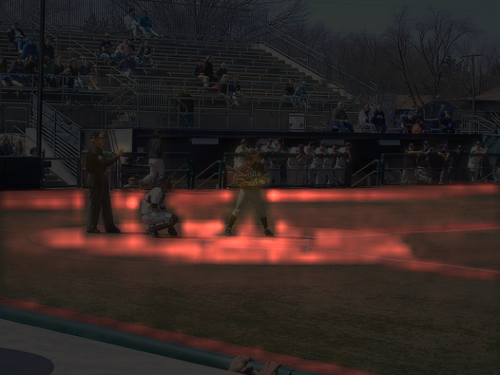} & \includegraphics[width=0.19\textwidth]{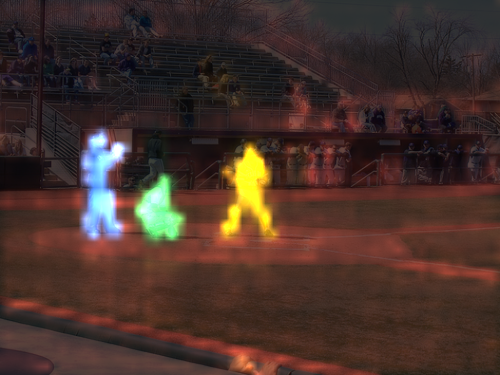} & \includegraphics[width=0.19\textwidth]{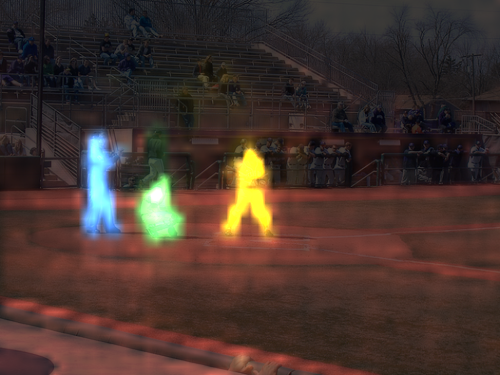} \\
    \multicolumn{5}{p{0.96\textwidth}}{\small Attention maps for three people (\blue{left}, \green{middle}, \yellow{right}) on a \red{playing field}.} \\
    \midrule
    Original Image & Head 1 & Head 2 & Head 3 & Head 4 \\
    \includegraphics[width=0.19\textwidth]{figures_appendix/439525.jpg} \phantom{.}&\phantom{.} \includegraphics[width=0.19\textwidth]{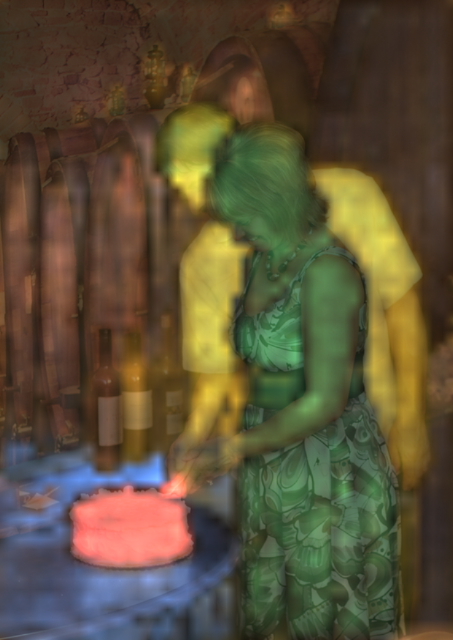} & \includegraphics[width=0.19\textwidth]{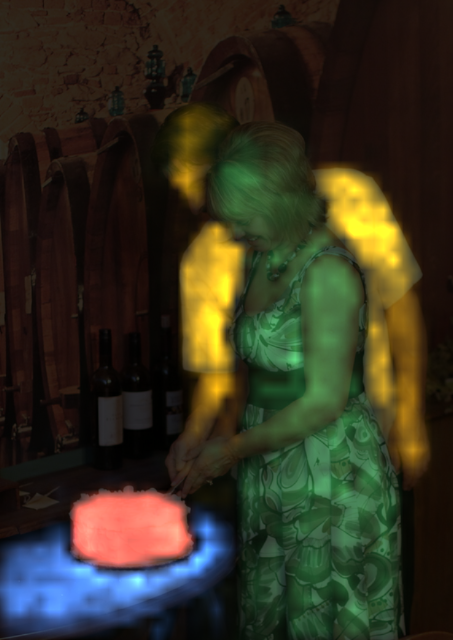} & \includegraphics[width=0.19\textwidth]{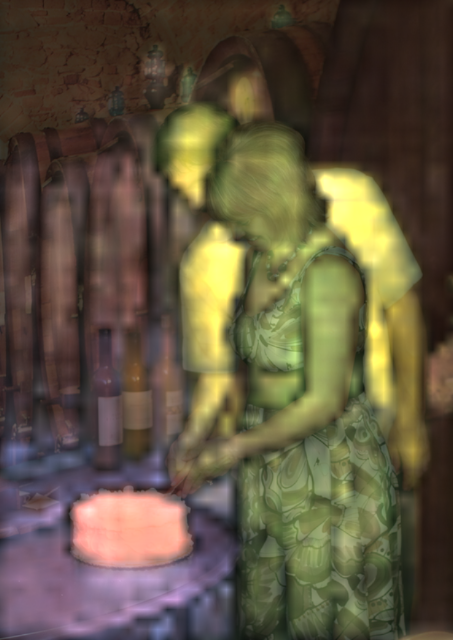} & \includegraphics[width=0.19\textwidth]{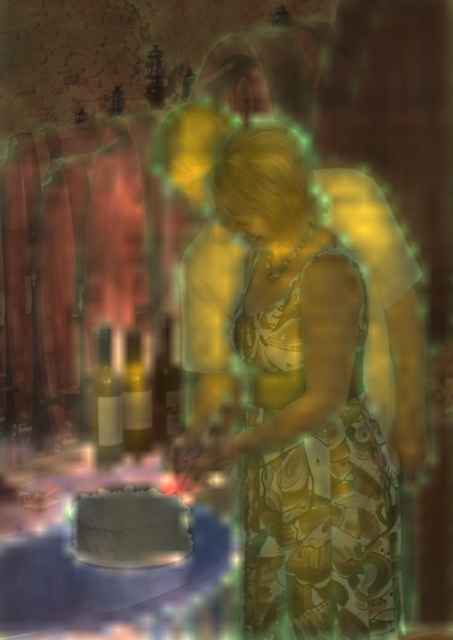} \\
    Panoptic Segmentation \phantom{.} & Head 5 & Head 6 & Head 7 & Head 8 \\
    \includegraphics[width=0.19\textwidth]{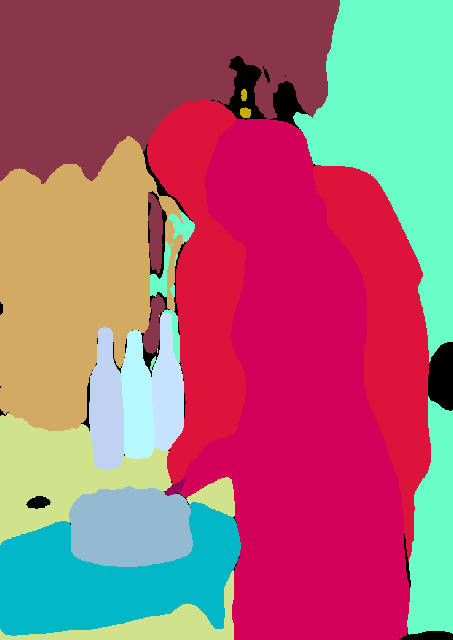} \phantom{.}&\phantom{.} \includegraphics[width=0.19\textwidth]{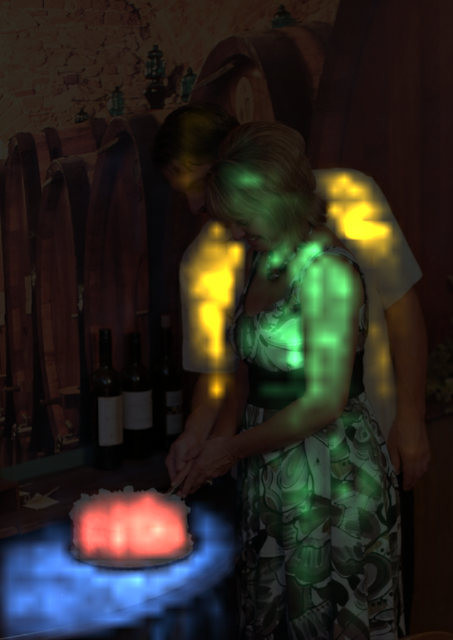} & \includegraphics[width=0.19\textwidth]{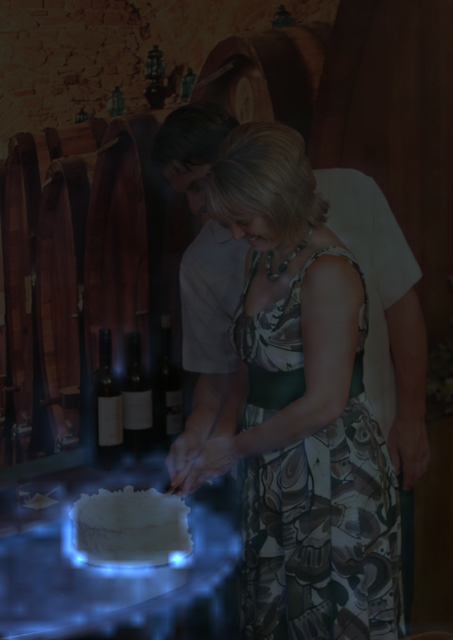} & \includegraphics[width=0.19\textwidth]{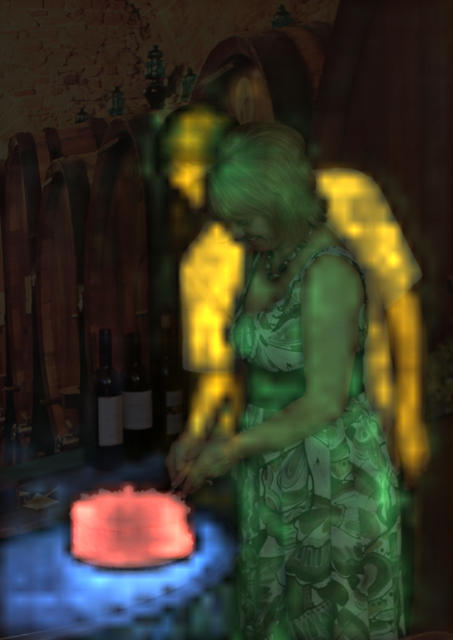} & \includegraphics[width=0.19\textwidth]{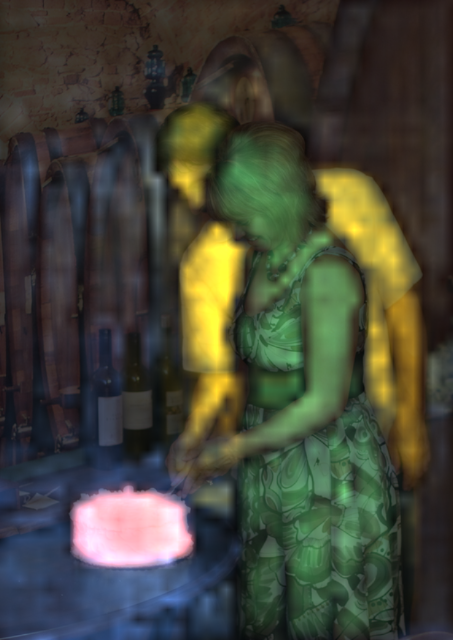} \\
    \multicolumn{5}{p{0.96\textwidth}}{\small Attention maps for two people (\green{woman}, \yellow{man}) cutting a \red{cake} on a \blue{table}.} \\
    \midrule
    \end{tabular}
    \vspace{3pt}
    \caption{Visualizing the transformer {\it M2P} attention maps for selected predicted masks. We observe that head 2, together with head 5, 7, and 8, mainly attends to the output mask regions. Head 1, 3, and 4 gather more context from broader regions, such as semantically-similar instances (scene 1 head 1) or mask boundaries (scene 2 head 4). In addition, we see that head 6 does not pay much attention to the pixel-path, except for some minor firings on the playing field and on the table. Instead, it focuses more on {\it M2M} self-attention which shares the same $\text{softmax}$ with {\it M2P} attention (\equref{eq:memory}).}
    \label{fig:attention}
\end{figure*}

We also visualize the {\it M2P} attention that connects the transformer to the CNN. Specifically, given an input image from COCO validation set, we first select four output masks of interest from the MaX-DeepLab-L panoptic prediction. Then, we probe the attention weights between the four masks and all the pixels, in the last dual-path transformer block. Finally, we colorize the four attention maps with four colors and visualize them in one figure. This process is repeated for two images and all eight attention heads as shown in \figref{fig:attention}. We omit our results for the first transformer block since it is mostly flat. This is expected because the memory feature in the first transformer block is unaware of the pixel-path input image at all. Unlike DETR~\cite{carion2020end} which focuses on object extreme points for detecting bounding boxes, our MaX-DeepLab attends to individual object (or stuff) masks. This mask-attending property makes MaX-DeepLab relatively robust to nearby objects with similar bounding boxes or close mass centers.

\subsection{More Technical Details}
\label{sec:appendix_details}
\begin{figure*}[!t]
    \centering
    \includegraphics[width=0.9\textwidth]{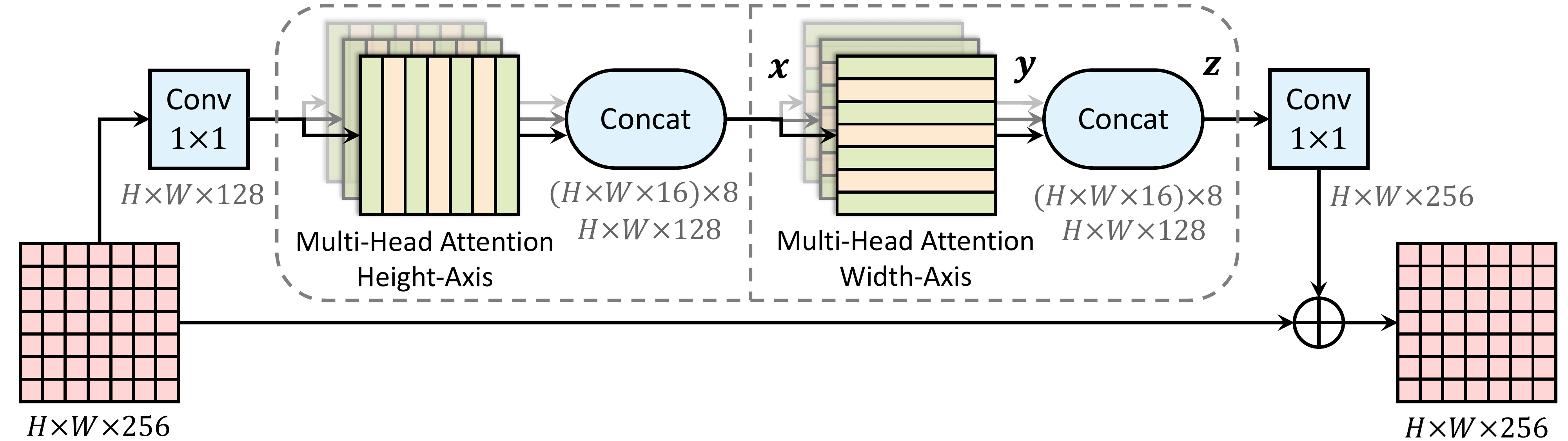}
    \caption{An example Axial-Block from Axial-DeepLab~\cite{wang2020axial}. This axial-attention bottleneck block consists of two axial-attention layers operating along height- and width-axis sequentially.}
    \label{fig:axial_block}
\end{figure*}

\begin{figure*}[t]
    \centering
    \setlength{\tabcolsep}{0.0em}
    \subfigure[Inception Stem~\cite{szegedy2016rethinking, szegedy2017inception}]{
        \includegraphics[width=0.17\linewidth]{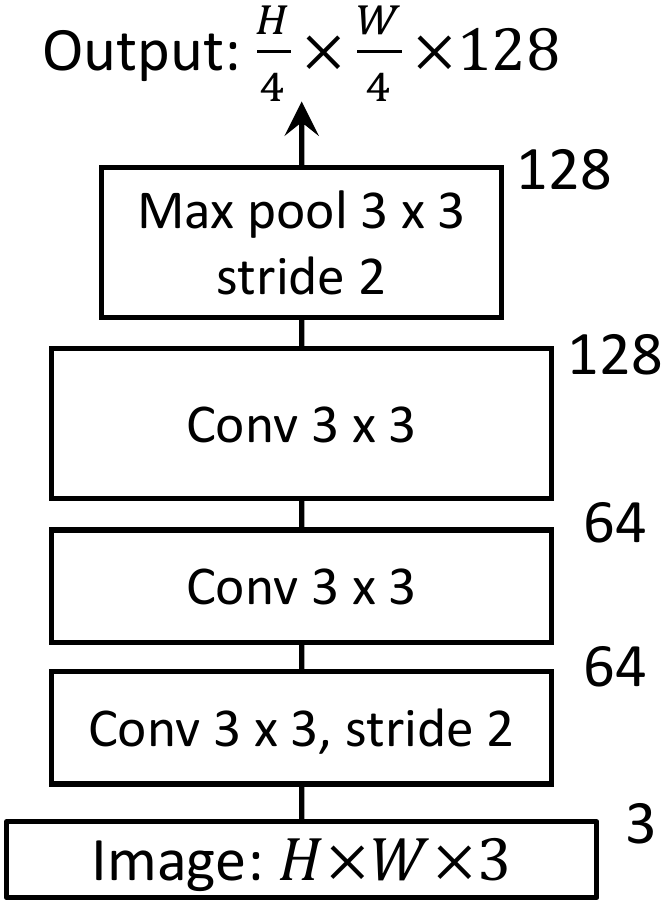}
        \label{fig:inception_stem}}
    \subfigure[Bottleneck block]{
        \includegraphics[width=0.17\linewidth]{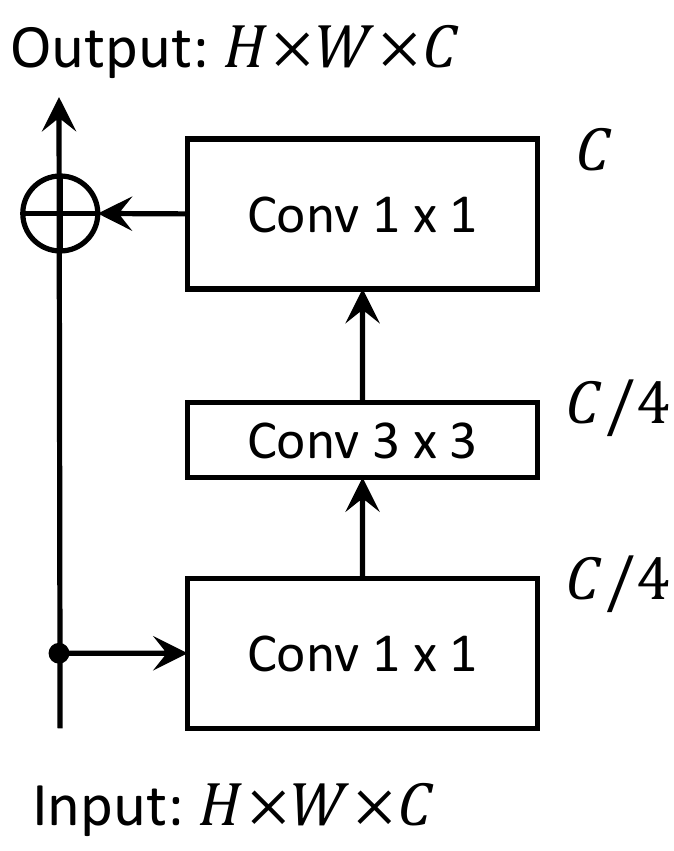}
        \label{fig:bottleneck}}
    \subfigure[Wide-Basic block]{
        \includegraphics[width=0.17\linewidth]{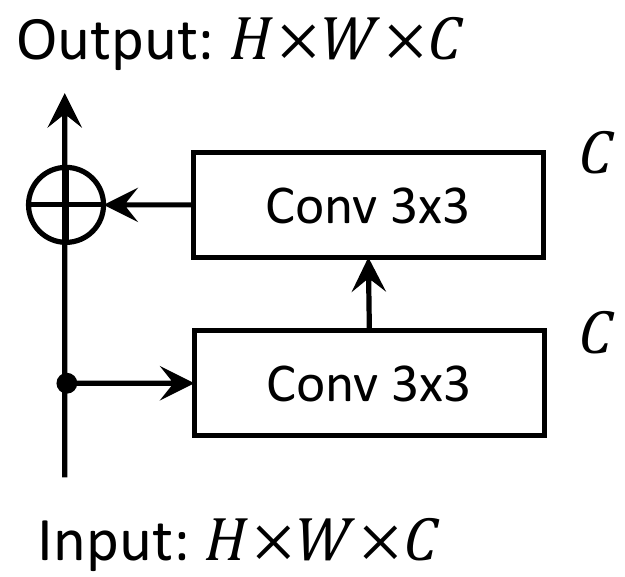}
        \label{fig:basic_block}}
    \subfigure[Wide-Bottle block]{
        \includegraphics[width=0.17\linewidth]{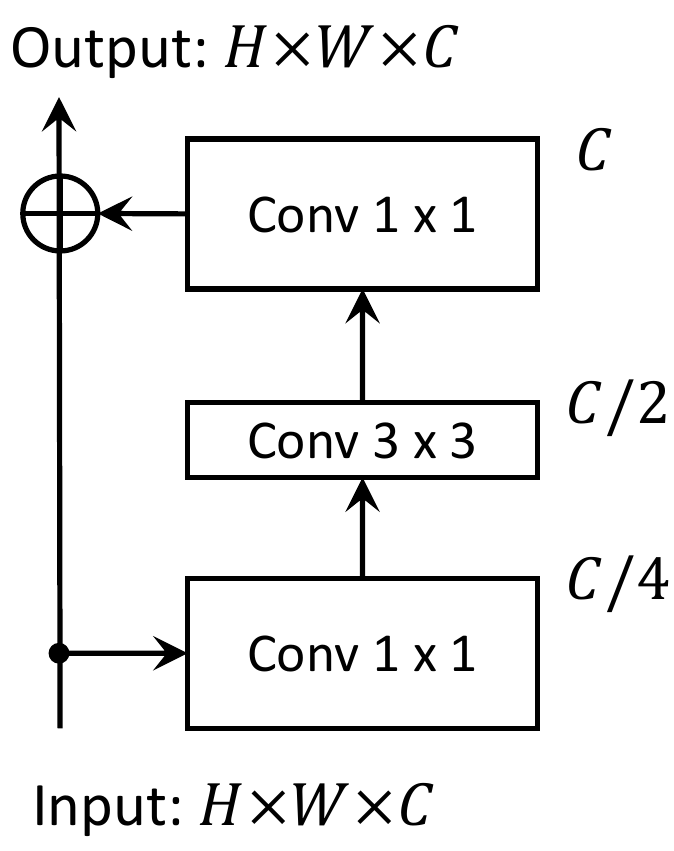}
        \label{fig:wide_bottle}} 
    \caption{Building blocks for our MaX-DeepLab architectures.}
    \label{fig:building_blocks}
\end{figure*}

\begin{figure*}[t]
    \centering
    \setlength{\tabcolsep}{0.0em}
    \subfigure[Dual-Path Transformer Block]{
        \includegraphics[width=0.23\linewidth]{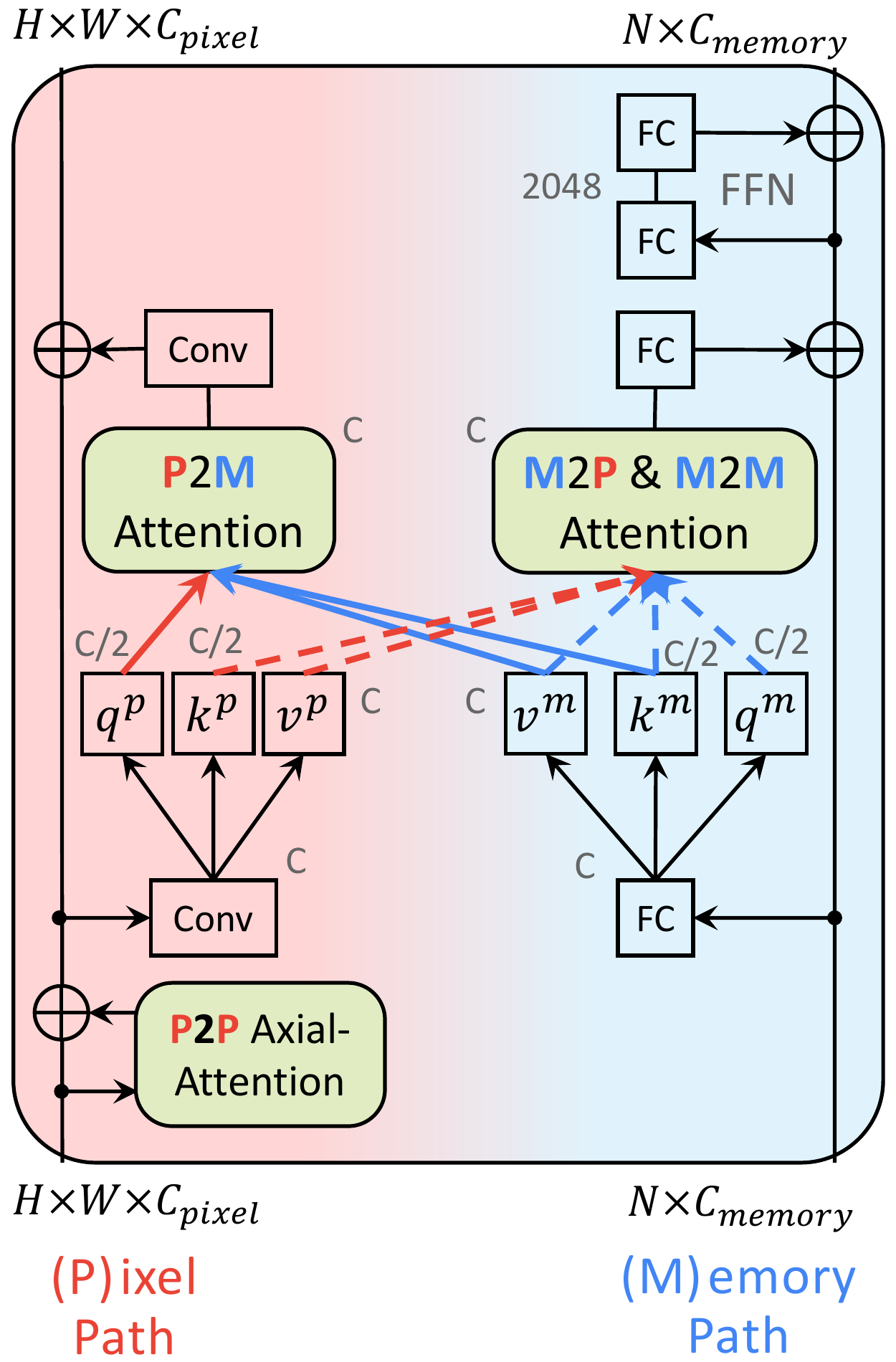}
        \label{fig:max_t}}
    \subfigure[MaX-DeepLab-Ablation]{
        \includegraphics[width=0.23\linewidth]{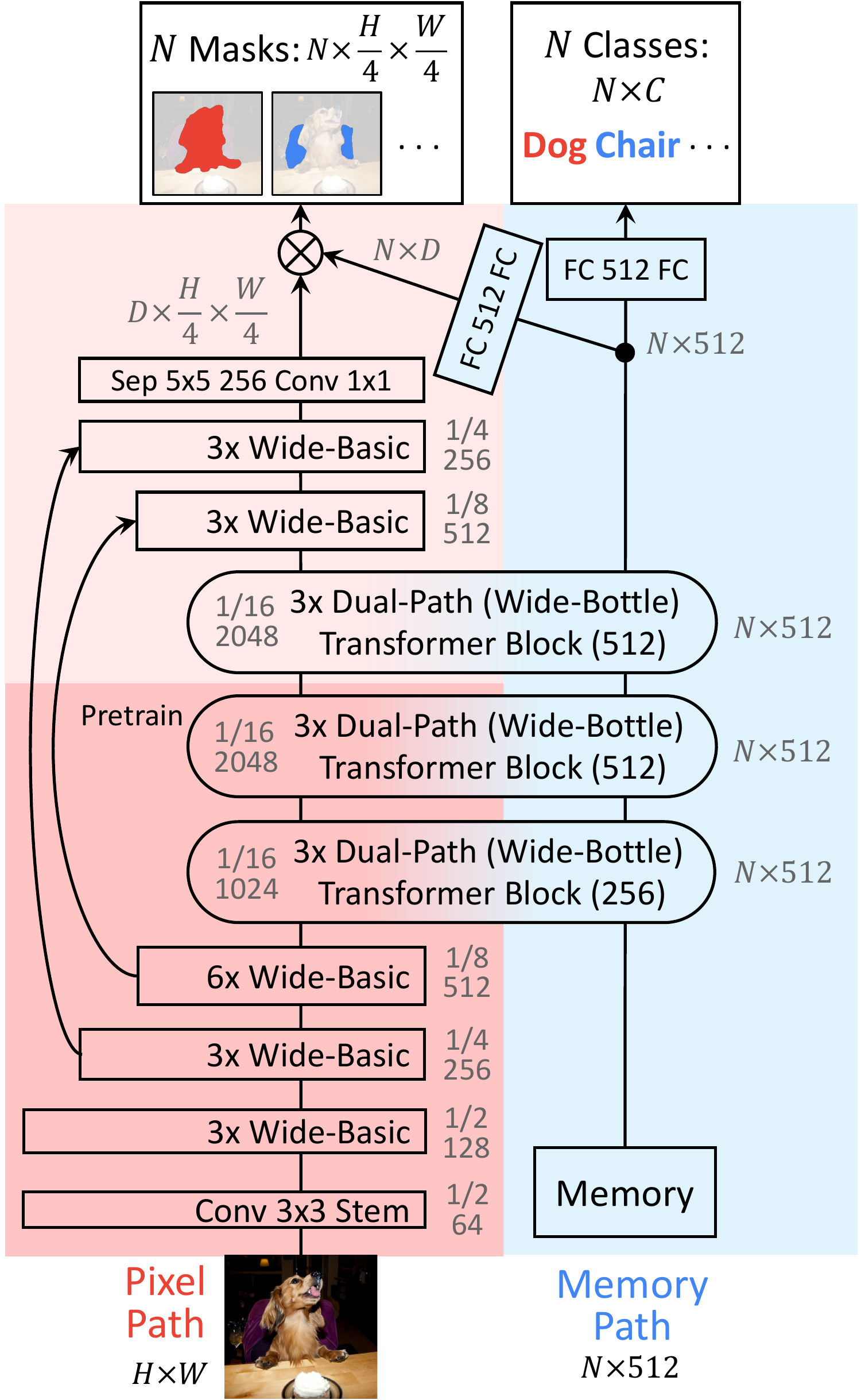}
        \label{fig:max_a}}
    \subfigure[MaX-DeepLab-S]{
        \includegraphics[width=0.23\linewidth]{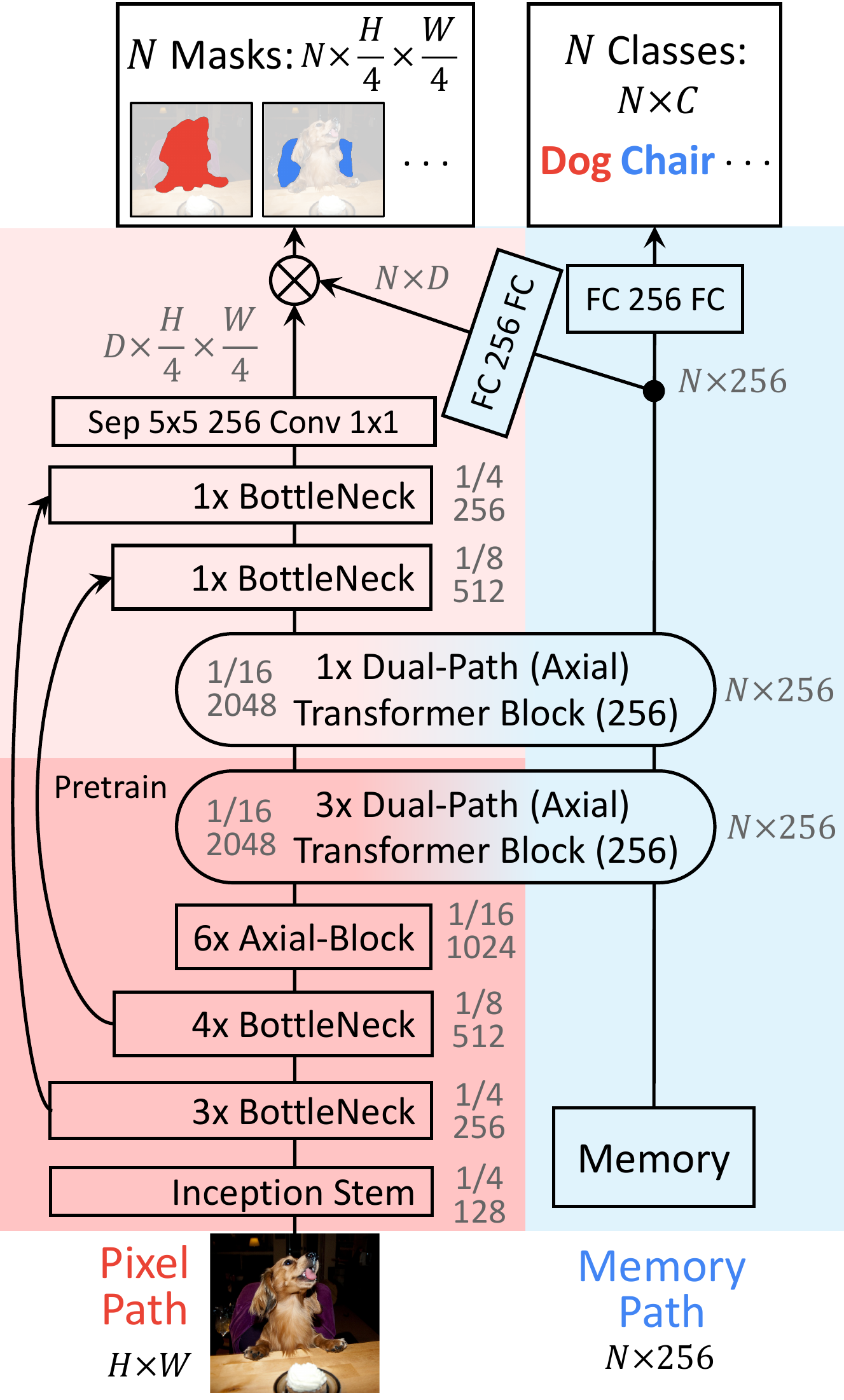}
        \label{fig:max_s}}
    \subfigure[MaX-DeepLab-L]{
        \includegraphics[width=0.23\linewidth]{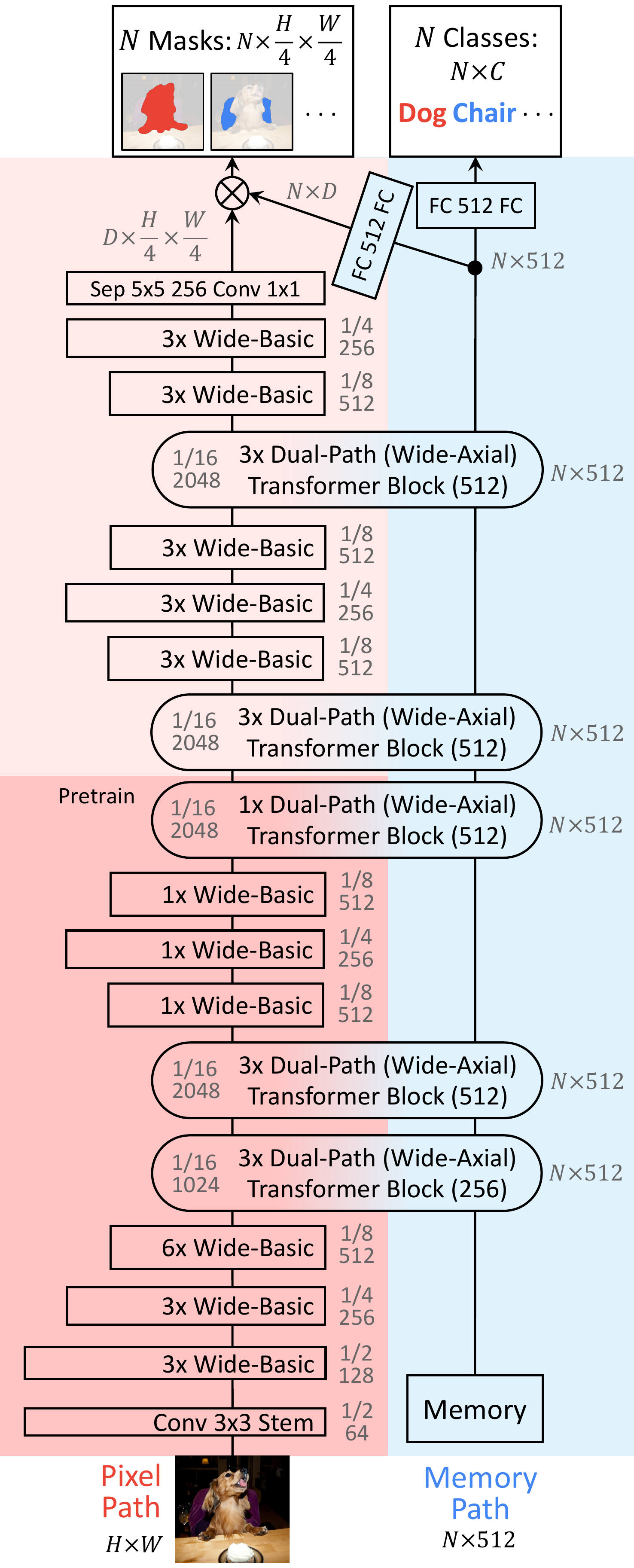}
        \label{fig:max_l}}
    \caption{More detailed MaX-DeepLab architectures. {\bf Pretrain} labels where we use a classification head to pretrain our models on ImageNet~\cite{krizhevsky2012imagenet}. \subref{fig:max_t} A dual-path transformer block with $C$ intermediate bottleneck channels. \subref{fig:max_a} The baseline architecture for our ablation studies in \secref{sec:ablation}. \subref{fig:max_s} MaX-DeepLab-S that matches the number of parameters and M-Adds of DETR-R101-Panoptic~\cite{carion2020end}. {\bf Axial-Block} (\figref{fig:axial_block}) is an axial-attention bottleneck block borrowed from Axial-DeepLab-L~\cite{wang2020axial}. \subref{fig:max_l} MaX-DeepLab-L that achieves the state-of-the-art performance on COCO~\cite{lin2014microsoft}. {\bf Wide-Axial} is a wide version of Axial-Block with doubled intermediate bottleneck channels, similar to the one used in Axial-DeepLab-XL~\cite{wang2020axial}. (The residual connections are dropped for neatness).
    }
    \label{fig:archs}
\end{figure*}

In \figref{fig:axial_block}, \figref{fig:building_blocks}, and \figref{fig:archs}, we include more details of our MaX-DeepLab architectures. As marked in the figure, we pretrain our model on ImageNet~\cite{krizhevsky2012imagenet}. The pretraining model uses only {\it P2P} attention (could be a convolutional residual block or an axial-attention block), without the other three types of attention, the feed-forward network (FFN), or the memory. We directly pretrain with an average pooling followed by a linear layer. This pretrained model is used as a backbone for panoptic segmentation, and it uses the backbone learning rate multiplier we mentioned in \secref{sec:exp}. After pretraining the CNN path, we apply (with random initialization) our proposed memory path, including the memory, the three types of attention, the FFNs, the decoding layers, and the output heads for panoptic segmentation. In addition, we employ multi-head attention with 8 heads for all attention operations. In MaX-DeepLab-L, we use shortcuts in the stacked decoder. Specifically, each decoding stage (resolution) is connected to the nearest two previous decoding stage outputs of the same resolution.

\FloatBarrier

\clearpage
{\small
\bibliographystyle{ieee_fullname}
\bibliography{9_references}

\begin{thebibliography}{100}\itemsep=-1pt

\bibitem{ainslie2020etc}
Joshua Ainslie, Santiago Ontanon, Chris Alberti, Philip Pham, Anirudh Ravula,
  and Sumit Sanghai.
\newblock Etc: Encoding long and structured data in transformers.
\newblock In {\em EMNLP}, 2020.

\bibitem{bahdanau2014neural}
Dzmitry Bahdanau, Kyunghyun Cho, and Yoshua Bengio.
\newblock Neural machine translation by jointly learning to align and
  translate.
\newblock In {\em ICLR}, 2015.

\bibitem{bai2017deep}
Min Bai and Raquel Urtasun.
\newblock Deep watershed transform for instance segmentation.
\newblock In {\em CVPR}, 2017.

\bibitem{ballard1981generalizing}
Dana~H Ballard.
\newblock Generalizing the hough transform to detect arbitrary shapes.
\newblock {\em Pattern Recognition}, 1981.

\bibitem{bello2019attention}
Irwan Bello, Barret Zoph, Ashish Vaswani, Jonathon Shlens, and Quoc~V Le.
\newblock Attention augmented convolutional networks.
\newblock In {\em ICCV}, 2019.

\bibitem{beltagy2020longformer}
Iz Beltagy, Matthew~E Peters, and Arman Cohan.
\newblock Longformer: The long-document transformer.
\newblock {\em arXiv preprint arXiv:2004.05150}, 2020.

\bibitem{bodla2017soft}
Navaneeth Bodla, Bharat Singh, Rama Chellappa, and Larry~S Davis.
\newblock Soft-nms--improving object detection with one line of code.
\newblock In {\em ICCV}, 2017.

\bibitem{bonde2020towards}
Ujwal Bonde, Pablo~F Alcantarilla, and Stefan Leutenegger.
\newblock Towards bounding-box free panoptic segmentation.
\newblock {\em arXiv:2002.07705}, 2020.

\bibitem{buades2005non}
Antoni Buades, Bartomeu Coll, and J-M Morel.
\newblock A non-local algorithm for image denoising.
\newblock In {\em CVPR}, 2005.

\bibitem{carion2020end}
Nicolas Carion, Francisco Massa, Gabriel Synnaeve, Nicolas Usunier, Alexander
  Kirillov, and Sergey Zagoruyko.
\newblock End-to-end object detection with transformers.
\newblock In {\em ECCV}, 2020.

\bibitem{chen2020naive}
Liang-Chieh Chen, Raphael~Gontijo Lopes, Bowen Cheng, Maxwell~D Collins, Ekin~D
  Cubuk, Barret Zoph, Hartwig Adam, and Jonathon Shlens.
\newblock {Naive-Student: Leveraging Semi-Supervised Learning in Video
  Sequences for Urban Scene Segmentation}.
\newblock In {\em ECCV}, 2020.

\bibitem{deeplabv12015}
Liang-Chieh Chen, George Papandreou, Iasonas Kokkinos, Kevin Murphy, and Alan~L
  Yuille.
\newblock Semantic image segmentation with deep convolutional nets and fully
  connected crfs.
\newblock In {\em ICLR}, 2015.

\bibitem{chen2018deeplabv2}
Liang-Chieh Chen, George Papandreou, Iasonas Kokkinos, Kevin Murphy, and Alan~L
  Yuille.
\newblock Deeplab: Semantic image segmentation with deep convolutional nets,
  atrous convolution, and fully connected crfs.
\newblock {\em IEEE TPAMI}, 2017.

\bibitem{chen2017deeplabv3}
Liang-Chieh Chen, George Papandreou, Florian Schroff, and Hartwig Adam.
\newblock Rethinking atrous convolution for semantic image segmentation.
\newblock {\em arXiv:1706.05587}, 2017.

\bibitem{chen2016attention}
Liang-Chieh Chen, Yi Yang, Jiang Wang, Wei Xu, and Alan~L Yuille.
\newblock Attention to scale: Scale-aware semantic image segmentation.
\newblock In {\em CVPR}, 2016.

\bibitem{deeplabv3plus2018}
Liang-Chieh Chen, Yukun Zhu, George Papandreou, Florian Schroff, and Hartwig
  Adam.
\newblock Encoder-decoder with atrous separable convolution for semantic image
  segmentation.
\newblock In {\em ECCV}, 2018.

\bibitem{chen2020simple}
Ting Chen, Simon Kornblith, Mohammad Norouzi, and Geoffrey Hinton.
\newblock A simple framework for contrastive learning of visual
  representations.
\newblock In {\em {ICML}}, 2020.

\bibitem{chen20182}
Yunpeng Chen, Yannis Kalantidis, Jianshu Li, Shuicheng Yan, and Jiashi Feng.
\newblock A\^{} 2-nets: Double attention networks.
\newblock In {\em NeurIPS}, 2018.

\bibitem{cheng2019spgnet}
Bowen Cheng, Liang-Chieh Chen, Yunchao Wei, Yukun Zhu, Zilong Huang, Jinjun
  Xiong, Thomas~S Huang, Wen-Mei Hwu, and Honghui Shi.
\newblock Spgnet: Semantic prediction guidance for scene parsing.
\newblock In {\em ICCV}, 2019.

\bibitem{cheng2019panopticworkshop}
Bowen Cheng, Maxwell~D Collins, Yukun Zhu, Ting Liu, Thomas~S Huang, Hartwig
  Adam, and Liang-Chieh Chen.
\newblock Panoptic-deeplab.
\newblock In {\em ICCV COCO + Mapillary Joint Recognition Challenge Workshop},
  2019.

\bibitem{cheng2019panoptic}
Bowen Cheng, Maxwell~D Collins, Yukun Zhu, Ting Liu, Thomas~S Huang, Hartwig
  Adam, and Liang-Chieh Chen.
\newblock {Panoptic-DeepLab: A Simple, Strong, and Fast Baseline for Bottom-Up
  Panoptic Segmentation}.
\newblock In {\em CVPR}, 2020.

\bibitem{cheng2016long}
Jianpeng Cheng, Li Dong, and Mirella Lapata.
\newblock Long short-term memory-networks for machine reading.
\newblock In {\em EMNLP}, 2016.

\bibitem{child2019generating}
Rewon Child, Scott Gray, Alec Radford, and Ilya Sutskever.
\newblock Generating long sequences with sparse transformers.
\newblock {\em arXiv:1904.10509}, 2019.

\bibitem{chollet2016xception}
Fran{\c{c}}ois Chollet.
\newblock Xception: Deep learning with depthwise separable convolutions.
\newblock In {\em CVPR}, 2017.

\bibitem{dai2017deformable}
Jifeng Dai, Haozhi Qi, Yuwen Xiong, Yi Li, Guodong Zhang, Han Hu, and Yichen
  Wei.
\newblock Deformable convolutional networks.
\newblock In {\em ICCV}, 2017.

\bibitem{dai2019transformer}
Zihang Dai, Zhilin Yang, Yiming Yang, Jaime~G Carbonell, Quoc Le, and Ruslan
  Salakhutdinov.
\newblock Transformer-xl: Attentive language models beyond a fixed-length
  context.
\newblock In {\em {ACL}}, 2019.

\bibitem{devlin2019bert}
Jacob Devlin, Ming-Wei Chang, Kenton Lee, and Kristina Toutanova.
\newblock {BERT}: Pre-training of deep bidirectional transformers for language
  understanding.
\newblock In {\em NAACL}, 2019.

\bibitem{dosovitskiy2020image}
Alexey Dosovitskiy, Lucas Beyer, Alexander Kolesnikov, Dirk Weissenborn,
  Xiaohua Zhai, Thomas Unterthiner, Mostafa Dehghani, Matthias Minderer, Georg
  Heigold, Sylvain Gelly, et~al.
\newblock An image is worth 16x16 words: Transformers for image recognition at
  scale.
\newblock {\em arXiv:2010.11929}, 2020.

\bibitem{fu2019dual}
Jun Fu, Jing Liu, Haijie Tian, Yong Li, Yongjun Bao, Zhiwei Fang, and Hanqing
  Lu.
\newblock Dual attention network for scene segmentation.
\newblock In {\em CVPR}, 2019.

\bibitem{gao2019ssap}
Naiyu Gao, Yanhu Shan, Yupei Wang, Xin Zhao, Yinan Yu, Ming Yang, and Kaiqi
  Huang.
\newblock Ssap: Single-shot instance segmentation with affinity pyramid.
\newblock In {\em ICCV}, 2019.

\bibitem{gupta2020gmat}
Ankit Gupta and Jonathan Berant.
\newblock Gmat: Global memory augmentation for transformers.
\newblock {\em arXiv:2006.03274}, 2020.

\bibitem{he2020momentum}
Kaiming He, Haoqi Fan, Yuxin Wu, Saining Xie, and Ross Girshick.
\newblock Momentum contrast for unsupervised visual representation learning.
\newblock In {\em CVPR}, 2020.

\bibitem{he2017mask}
Kaiming He, Georgia Gkioxari, Piotr Doll{\'a}r, and Ross Girshick.
\newblock Mask r-cnn.
\newblock In {\em ICCV}, 2017.

\bibitem{he2016deep}
Kaiming He, Xiangyu Zhang, Shaoqing Ren, and Jian Sun.
\newblock Deep residual learning for image recognition.
\newblock In {\em CVPR}, 2016.

\bibitem{ho2019axial}
Jonathan Ho, Nal Kalchbrenner, Dirk Weissenborn, and Tim Salimans.
\newblock Axial attention in multidimensional transformers.
\newblock {\em arXiv:1912.12180}, 2019.

\bibitem{hu2018relation}
Han Hu, Jiayuan Gu, Zheng Zhang, Jifeng Dai, and Yichen Wei.
\newblock Relation networks for object detection.
\newblock In {\em CVPR}, 2018.

\bibitem{hu2019local}
Han Hu, Zheng Zhang, Zhenda Xie, and Stephen Lin.
\newblock Local relation networks for image recognition.
\newblock In {\em ICCV}, 2019.

\bibitem{huang2016deep}
Gao Huang, Yu Sun, Zhuang Liu, Daniel Sedra, and Kilian~Q Weinberger.
\newblock Deep networks with stochastic depth.
\newblock In {\em ECCV}, 2016.

\bibitem{huang2019ccnet}
Zilong Huang, Xinggang Wang, Lichao Huang, Chang Huang, Yunchao Wei, and Wenyu
  Liu.
\newblock Ccnet: Criss-cross attention for semantic segmentation.
\newblock In {\em ICCV}, 2019.

\bibitem{hwang2019segsort}
Jyh-Jing Hwang, Stella~X Yu, Jianbo Shi, Maxwell~D Collins, Tien-Ju Yang, Xiao
  Zhang, and Liang-Chieh Chen.
\newblock {SegSort}: Segmentation by discriminative sorting of segments.
\newblock In {\em ICCV}, 2019.

\bibitem{ioffe2015batch}
Sergey Ioffe and Christian Szegedy.
\newblock Batch normalization: accelerating deep network training by reducing
  internal covariate shift.
\newblock In {\em {ICML}}, 2015.

\bibitem{isensee2018nnu}
Fabian Isensee, Jens Petersen, Andre Klein, David Zimmerer, Paul~F Jaeger,
  Simon Kohl, Jakob Wasserthal, Gregor Koehler, Tobias Norajitra, Sebastian
  Wirkert, et~al.
\newblock nnu-net: Self-adapting framework for u-net-based medical image
  segmentation.
\newblock {\em arXiv:1809.10486}, 2018.

\bibitem{jia2016dynamic}
Xu Jia, Bert De~Brabandere, Tinne Tuytelaars, and Luc~V Gool.
\newblock Dynamic filter networks.
\newblock In {\em NeurIPS}, 2016.

\bibitem{kendall2018multi}
Alex Kendall, Yarin Gal, and Roberto Cipolla.
\newblock Multi-task learning using uncertainty to weigh losses for scene
  geometry and semantics.
\newblock In {\em CVPR}, 2018.

\bibitem{keuper2015efficient}
Margret Keuper, Evgeny Levinkov, Nicolas Bonneel, Guillaume Lavou{\'e}, Thomas
  Brox, and Bjorn Andres.
\newblock Efficient decomposition of image and mesh graphs by lifted multicuts.
\newblock In {\em ICCV}, 2015.

\bibitem{khosla2020supervised}
Prannay Khosla, Piotr Teterwak, Chen Wang, Aaron Sarna, Yonglong Tian, Phillip
  Isola, Aaron Maschinot, Ce Liu, and Dilip Krishnan.
\newblock Supervised contrastive learning.
\newblock In {\em NeurIPS}, 2020.

\bibitem{kirillov2019panoptic}
Alexander Kirillov, Ross Girshick, Kaiming He, and Piotr Doll{\'a}r.
\newblock Panoptic feature pyramid networks.
\newblock In {\em CVPR}, 2019.

\bibitem{kirillov2018panoptic}
Alexander Kirillov, Kaiming He, Ross Girshick, Carsten Rother, and Piotr
  Doll{\'a}r.
\newblock Panoptic segmentation.
\newblock In {\em CVPR}, 2019.

\bibitem{kitaev2020reformer}
Nikita Kitaev, {\L}ukasz Kaiser, and Anselm Levskaya.
\newblock Reformer: The efficient transformer.
\newblock In {\em ICLR}, 2020.

\bibitem{krizhevsky2012imagenet}
Alex Krizhevsky, Ilya Sutskever, and Geoffrey~E Hinton.
\newblock Imagenet classification with deep convolutional neural networks.
\newblock In {\em NeurIPS}, 2012.

\bibitem{Kuhn1955NAVAL}
Harold~W Kuhn.
\newblock The hungarian method for the assignment problem.
\newblock {\em Naval research logistics quarterly}, 2(1-2):83--97, 1955.

\bibitem{lecun1998gradient}
Yann LeCun, L{\'e}on Bottou, Yoshua Bengio, and Patrick Haffner.
\newblock Gradient-based learning applied to document recognition.
\newblock {\em Proceedings of the IEEE}, 86(11):2278--2324, 1998.

\bibitem{leibe2004combined}
Bastian Leibe, Ales Leonardis, and Bernt Schiele.
\newblock Combined object categorization and segmentation with an implicit
  shape model.
\newblock In {\em Workshop on statistical learning in computer vision, ECCV},
  2004.

\bibitem{li2018learning}
Jie Li, Allan Raventos, Arjun Bhargava, Takaaki Tagawa, and Adrien Gaidon.
\newblock Learning to fuse things and stuff.
\newblock {\em arXiv:1812.01192}, 2018.

\bibitem{li2020unifying}
Qizhu Li, Xiaojuan Qi, and Philip~HS Torr.
\newblock Unifying training and inference for panoptic segmentation.
\newblock In {\em CVPR}, 2020.

\bibitem{li2018attention}
Yanwei Li, Xinze Chen, Zheng Zhu, Lingxi Xie, Guan Huang, Dalong Du, and
  Xingang Wang.
\newblock Attention-guided unified network for panoptic segmentation.
\newblock In {\em CVPR}, 2019.

\bibitem{li2020neural}
Yingwei Li, Xiaojie Jin, Jieru Mei, Xiaochen Lian, Linjie Yang, Cihang Xie,
  Qihang Yu, Yuyin Zhou, Song Bai, and Alan Yuille.
\newblock Neural architecture search for lightweight non-local networks.
\newblock In {\em CVPR}, 2020.

\bibitem{lin2017feature}
Tsung-Yi Lin, Piotr Doll{\'a}r, Ross Girshick, Kaiming He, Bharath Hariharan,
  and Serge Belongie.
\newblock Feature pyramid networks for object detection.
\newblock In {\em CVPR}, 2017.

\bibitem{lin2017focal}
Tsung-Yi Lin, Priya Goyal, Ross Girshick, Kaiming He, and Piotr Doll{\'a}r.
\newblock Focal loss for dense object detection.
\newblock In {\em ICCV}, 2017.

\bibitem{lin2014microsoft}
Tsung-Yi Lin, Michael Maire, Serge Belongie, James Hays, Pietro Perona, Deva
  Ramanan, Piotr Doll{\'a}r, and C~Lawrence Zitnick.
\newblock Microsoft coco: Common objects in context.
\newblock In {\em ECCV}, 2014.

\bibitem{liu2019auto}
Chenxi Liu, Liang-Chieh Chen, Florian Schroff, Hartwig Adam, Wei Hua, Alan
  Yuille, and Li Fei-Fei.
\newblock Auto-deeplab: Hierarchical neural architecture search for semantic
  image segmentation.
\newblock In {\em CVPR}, 2019.

\bibitem{liu2019variance}
Liyuan Liu, Haoming Jiang, Pengcheng He, Weizhu Chen, Xiaodong Liu, Jianfeng
  Gao, and Jiawei Han.
\newblock On the variance of the adaptive learning rate and beyond.
\newblock In {\em ICLR}, 2020.

\bibitem{liu2018path}
Shu Liu, Lu Qi, Haifang Qin, Jianping Shi, and Jiaya Jia.
\newblock Path aggregation network for instance segmentation.
\newblock In {\em CVPR}, 2018.

\bibitem{liu2016ssd}
Wei Liu, Dragomir Anguelov, Dumitru Erhan, Christian Szegedy, Scott Reed,
  Cheng-Yang Fu, and Alexander~C Berg.
\newblock Ssd: Single shot multibox detector.
\newblock In {\em ECCV}, 2016.

\bibitem{liu2020cbnet}
Yudong Liu, Yongtao Wang, Siwei Wang, TingTing Liang, Qijie Zhao, Zhi Tang, and
  Haibin Ling.
\newblock Cbnet: A novel composite backbone network architecture for object
  detection.
\newblock In {\em AAAI}, 2020.

\bibitem{liu2018affinity}
Yiding Liu, Siyu Yang, Bin Li, Wengang Zhou, Jizheng Xu, Houqiang Li, and Yan
  Lu.
\newblock Affinity derivation and graph merge for instance segmentation.
\newblock In {\em ECCV}, 2018.

\bibitem{liu2019e2e}
Huanyu Liu1, Chao Peng, Changqian Yu, Jingbo Wang, Xu Liu, Gang Yu, and Wei
  Jiang.
\newblock An end-to-end network for panoptic segmentation.
\newblock In {\em CVPR}, 2019.

\bibitem{luong2015effective}
Minh-Thang Luong, Hieu Pham, and Christopher~D Manning.
\newblock Effective approaches to attention-based neural machine translation.
\newblock In {\em EMNLP}, 2015.

\bibitem{milletari2016v}
Fausto Milletari, Nassir Navab, and Seyed-Ahmad Ahmadi.
\newblock V-net: Fully convolutional neural networks for volumetric medical
  image segmentation.
\newblock In {\em 3DV}, 2016.

\bibitem{neven2019instance}
Davy Neven, Bert~De Brabandere, Marc Proesmans, and Luc~Van Gool.
\newblock Instance segmentation by jointly optimizing spatial embeddings and
  clustering bandwidth.
\newblock In {\em CVPR}, 2019.

\bibitem{newell2016stacked}
Alejandro Newell, Kaiyu Yang, and Jia Deng.
\newblock Stacked hourglass networks for human pose estimation.
\newblock In {\em ECCV}, 2016.

\bibitem{parmar2018image}
Niki Parmar, Ashish Vaswani, Jakob Uszkoreit, {\L}ukasz Kaiser, Noam Shazeer,
  Alexander Ku, and Dustin Tran.
\newblock Image transformer.
\newblock In {\em {ICML}}, 2018.

\bibitem{porzi2019seamless}
Lorenzo Porzi, Samuel~Rota Bul{\`o}, Aleksander Colovic, and Peter
  Kontschieder.
\newblock Seamless scene segmentation.
\newblock In {\em CVPR}, 2019.

\bibitem{dai2017coco}
Haozhi Qi, Zheng Zhang, Bin Xiao, Han Hu, Bowen Cheng, Yichen Wei, and Jifeng
  Dai.
\newblock Deformable convolutional networks -- coco detection and segmentation
  challenge 2017 entry.
\newblock {\em ICCV COCO Challenge Workshop}, 2017.

\bibitem{qiao2020detectors}
Siyuan Qiao, Liang-Chieh Chen, and Alan Yuille.
\newblock Detectors: Detecting objects with recursive feature pyramid and
  switchable atrous convolution.
\newblock {\em arXiv:2006.02334}, 2020.

\bibitem{parmar2019stand}
Prajit Ramachandran, Niki Parmar, Ashish Vaswani, Irwan Bello, Anselm Levskaya,
  and Jon Shlens.
\newblock Stand-alone self-attention in vision models.
\newblock In {\em NeurIPS}, 2019.

\bibitem{ren2015faster}
Shaoqing Ren, Kaiming He, Ross Girshick, and Jian Sun.
\newblock Faster r-cnn: Towards real-time object detection with region proposal
  networks.
\newblock In {\em NeurIPS}, 2015.

\bibitem{ronneberger2015u}
Olaf Ronneberger, Philipp Fischer, and Thomas Brox.
\newblock U-net: Convolutional networks for biomedical image segmentation.
\newblock In {\em International Conference on Medical image computing and
  computer-assisted intervention}, 2015.

\bibitem{shaw2018self}
Peter Shaw, Jakob Uszkoreit, and Ashish Vaswani.
\newblock Self-attention with relative position representations.
\newblock In {\em {NAACL}}, 2018.

\bibitem{shen2021efficient}
Zhuoran Shen, Mingyuan Zhang, Haiyu Zhao, Shuai Yi, and Hongsheng Li.
\newblock Efficient attention: Attention with linear complexities.
\newblock In {\em WACV}, 2021.

\bibitem{sofiiuk2019adaptis}
Konstantin Sofiiuk, Olga Barinova, and Anton Konushin.
\newblock Adaptis: Adaptive instance selection network.
\newblock In {\em ICCV}, 2019.

\bibitem{stewart2016end}
Russell Stewart, Mykhaylo Andriluka, and Andrew~Y Ng.
\newblock End-to-end people detection in crowded scenes.
\newblock In {\em CVPR}, 2016.

\bibitem{szegedy2017inception}
Christian Szegedy, Sergey Ioffe, Vincent Vanhoucke, and Alexander~A Alemi.
\newblock Inception-v4, inception-resnet and the impact of residual connections
  on learning.
\newblock In {\em AAAI}, 2017.

\bibitem{szegedy2016rethinking}
Christian Szegedy, Vincent Vanhoucke, Sergey Ioffe, Jon Shlens, and Zbigniew
  Wojna.
\newblock Rethinking the inception architecture for computer vision.
\newblock In {\em CVPR}, 2016.

\bibitem{taghanaki2019combo}
Saeid~Asgari Taghanaki, Yefeng Zheng, S~Kevin Zhou, Bogdan Georgescu, Puneet
  Sharma, Daguang Xu, Dorin Comaniciu, and Ghassan Hamarneh.
\newblock Combo loss: Handling input and output imbalance in multi-organ
  segmentation.
\newblock {\em Computerized Medical Imaging and Graphics}, 75:24--33, 2019.

\bibitem{tan2020efficientdet}
Mingxing Tan, Ruoming Pang, and Quoc~V Le.
\newblock Efficientdet: Scalable and efficient object detection.
\newblock In {\em CVPR}, 2020.

\bibitem{tian2020conditional}
Zhi Tian, Chunhua Shen, and Hao Chen.
\newblock Conditional convolutions for instance segmentation.
\newblock In {\em ECCV}, 2020.

\bibitem{uhrig2018box2pix}
Jonas Uhrig, Eike Rehder, Bj{\"o}rn Fr{\"o}hlich, Uwe Franke, and Thomas Brox.
\newblock Box2pix: Single-shot instance segmentation by assigning pixels to
  object boxes.
\newblock In {\em IEEE Intelligent Vehicles Symposium (IV)}, 2018.

\bibitem{vaswani2017attention}
Ashish Vaswani, Noam Shazeer, Niki Parmar, Jakob Uszkoreit, Llion Jones,
  Aidan~N Gomez, {\L}ukasz Kaiser, and Illia Polosukhin.
\newblock Attention is all you need.
\newblock In {\em NeurIPS}, 2017.

\bibitem{vincent1991watersheds}
Luc Vincent and Pierre Soille.
\newblock Watersheds in digital spaces: an efficient algorithm based on
  immersion simulations.
\newblock {\em IEEE TPAMI}, 1991.

\bibitem{wang2020axial}
Huiyu Wang, Yukun Zhu, Bradley Green, Hartwig Adam, Alan Yuille, and
  Liang-Chieh Chen.
\newblock {Axial-DeepLab: Stand-Alone Axial-Attention for Panoptic
  Segmentation}.
\newblock In {\em ECCV}, 2020.

\bibitem{wang2020linformer}
Sinong Wang, Belinda Li, Madian Khabsa, Han Fang, and Hao Ma.
\newblock Linformer: Self-attention with linear complexity.
\newblock {\em arXiv:2006.04768}, 2020.

\bibitem{wang2018non}
Xiaolong Wang, Ross Girshick, Abhinav Gupta, and Kaiming He.
\newblock Non-local neural networks.
\newblock In {\em CVPR}, 2018.

\bibitem{wang2020solov2}
Xinlong Wang, Rufeng Zhang, Tao Kong, Lei Li, and Chunhua Shen.
\newblock {SOLOv2}: Dynamic and fast instance segmentation.
\newblock In {\em NeurIPS}, 2020.

\bibitem{wu2019detectron2}
Yuxin Wu, Alexander Kirillov, Francisco Massa, Wan-Yen Lo, and Ross Girshick.
\newblock Detectron2.
\newblock \url{https://github.com/facebookresearch/detectron2}, 2019.

\bibitem{wu2018improving}
Zhirong Wu, Alexei~A Efros, and Stella~X Yu.
\newblock Improving generalization via scalable neighborhood component
  analysis.
\newblock In {\em ECCV}, 2018.

\bibitem{wu2019wider}
Zifeng Wu, Chunhua Shen, and Anton Van Den~Hengel.
\newblock Wider or deeper: Revisiting the {ResNet} model for visual
  recognition.
\newblock {\em Pattern Recognition}, 2019.

\bibitem{wu2018unsupervised}
Zhirong Wu, Yuanjun Xiong, Stella~X Yu, and Dahua Lin.
\newblock Unsupervised feature learning via non-parametric instance
  discrimination.
\newblock In {\em CVPR}, 2018.

\bibitem{xie2017aggregated}
Saining Xie, Ross Girshick, Piotr Doll{\'a}r, Zhuowen Tu, and Kaiming He.
\newblock Aggregated residual transformations for deep neural networks.
\newblock In {\em CVPR}, 2017.

\bibitem{xiong2019upsnet}
Yuwen Xiong, Renjie Liao, Hengshuang Zhao, Rui Hu, Min Bai, Ersin Yumer, and
  Raquel Urtasun.
\newblock Upsnet: A unified panoptic segmentation network.
\newblock In {\em CVPR}, 2019.

\bibitem{yang2019condconv}
Brandon Yang, Gabriel Bender, Quoc~V Le, and Jiquan Ngiam.
\newblock Condconv: Conditionally parameterized convolutions for efficient
  inference.
\newblock In {\em NeurIPS}, 2019.

\bibitem{yang2019deeperlab}
Tien-Ju Yang, Maxwell~D Collins, Yukun Zhu, Jyh-Jing Hwang, Ting Liu, Xiao
  Zhang, Vivienne Sze, George Papandreou, and Liang-Chieh Chen.
\newblock Deeperlab: Single-shot image parser.
\newblock {\em arXiv:1902.05093}, 2019.

\bibitem{yang2020sognet}
Yibo Yang, Hongyang Li, Xia Li, Qijie Zhao, Jianlong Wu, and Zhouchen Lin.
\newblock Sognet: Scene overlap graph network for panoptic segmentation.
\newblock In {\em AAAI}, 2020.

\bibitem{wrn2016wide}
Sergey Zagoruyko and Nikos Komodakis.
\newblock Wide residual networks.
\newblock In {\em BMVC}, 2016.

\bibitem{zaheer2020big}
Manzil Zaheer, Guru Guruganesh, Kumar~Avinava Dubey, Joshua Ainslie, Chris
  Alberti, Santiago Ontanon, Philip Pham, Anirudh Ravula, Qifan Wang, Li Yang,
  et~al.
\newblock Big bird: Transformers for longer sequences.
\newblock In {\em NeurIPS}, 2020.

\bibitem{zhang2019lookahead}
Michael Zhang, James Lucas, Jimmy Ba, and Geoffrey~E Hinton.
\newblock Lookahead optimizer: k steps forward, 1 step back.
\newblock In {\em NeurIPS}, 2019.

\bibitem{zhang2020bridging}
Shifeng Zhang, Cheng Chi, Yongqiang Yao, Zhen Lei, and Stan~Z Li.
\newblock Bridging the gap between anchor-based and anchor-free detection via
  adaptive training sample selection.
\newblock In {\em CVPR}, 2020.

\bibitem{zhao2018psanet}
Hengshuang Zhao, Yi Zhang, Shu Liu, Jianping Shi, Chen Change~Loy, Dahua Lin,
  and Jiaya Jia.
\newblock Psanet: Point-wise spatial attention network for scene parsing.
\newblock In {\em ECCV}, 2018.

\bibitem{zhu2019empirical}
Xizhou Zhu, Dazhi Cheng, Zheng Zhang, Stephen Lin, and Jifeng Dai.
\newblock An empirical study of spatial attention mechanisms in deep networks.
\newblock In {\em ICCV}, pages 6688--6697, 2019.

\bibitem{zhu2020deformable}
Xizhou Zhu, Weijie Su, Lewei Lu, Bin Li, Xiaogang Wang, and Jifeng Dai.
\newblock Deformable detr: Deformable transformers for end-to-end object
  detection.
\newblock {\em arXiv:2010.04159}, 2020.

\bibitem{zhu2019asymmetric}
Zhen Zhu, Mengde Xu, Song Bai, Tengteng Huang, and Xiang Bai.
\newblock Asymmetric non-local neural networks for semantic segmentation.
\newblock In {\em CVPR}, 2019.

\end{thebibliography}
}

\end{document}